\documentclass[journal]{IEEEtran}  % Comment this line out
\usepackage{array}
\usepackage{ifpdf}
\usepackage{graphicx}
\usepackage{epsfig} % for postscript graphics files

\usepackage[cmex10]{amsmath}
\usepackage{amsfonts}
\usepackage{amsthm}
\usepackage{algorithmic}
\usepackage{algorithm2e}
\usepackage{mdwmath}
\usepackage{colortbl}
\usepackage{hhline}
\usepackage{multirow}
\usepackage[caption=false,font=footnotesize]{subfig}
\usepackage{url}
\usepackage{float}
\usepackage{balance}
\usepackage[english]{babel}
\usepackage{comment}
\usepackage{adjustbox}

\theoremstyle{definition}

\newcommand{\figref}[1]{Fig.~\ref{fig:#1}}
\newcommand{\tblref}[1]{Table~\ref{tbl:#1}}
\newcommand{\secref}[1]{Section~\ref{sec:#1}}
\newcommand{\algref}[1]{Algorithm.~\ref{alg:#1}}

\newcommand{\figIRS}{
\begin{figure}[b]
 \centering
 \subfloat[]{\label{fig:IRSProc}\includegraphics[width=0.42\linewidth, trim=0in 0in 0in 0in,
  clip=true]{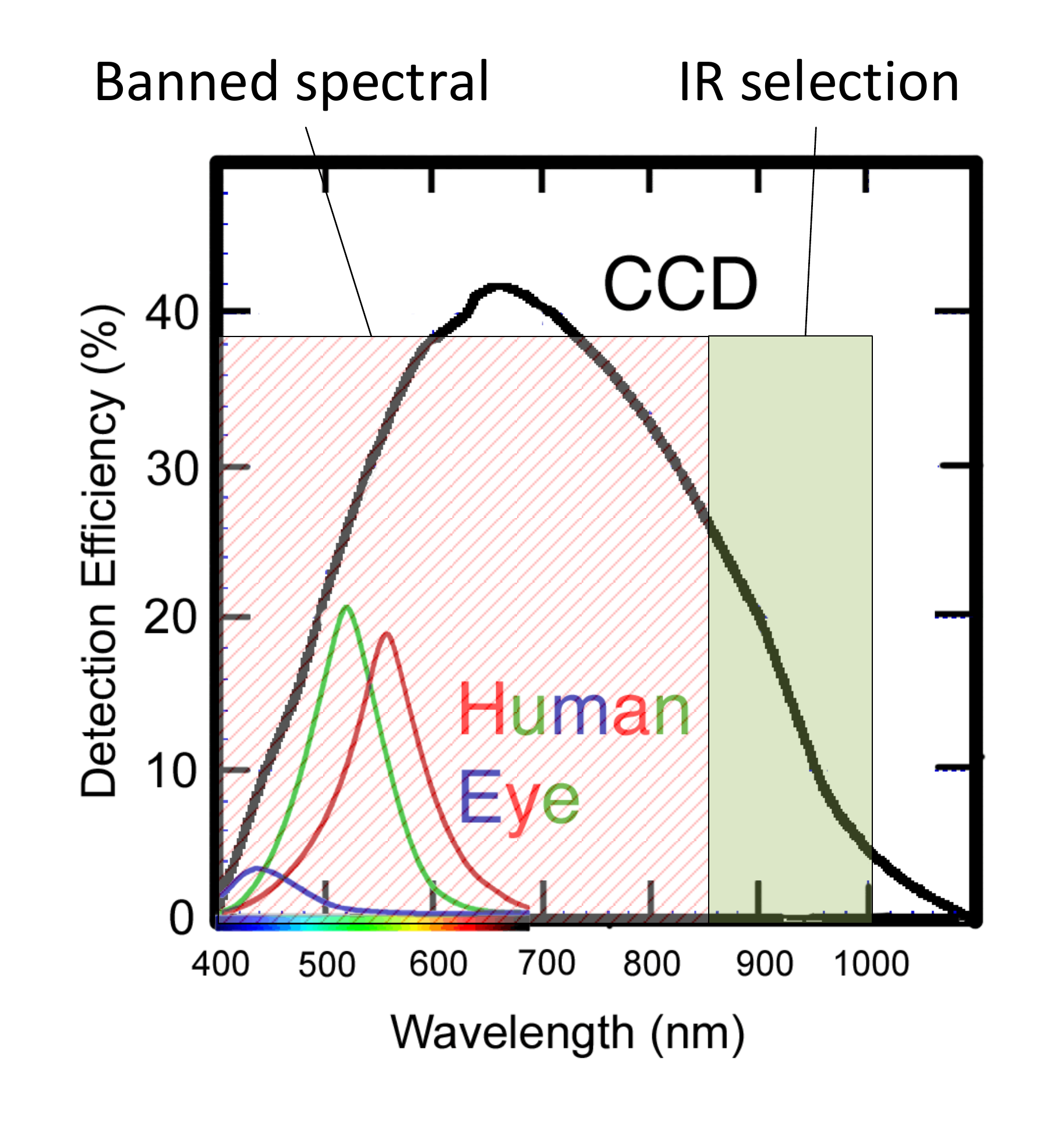}}
 \subfloat[]{\label{fig:IRSdiagram}\includegraphics[width=0.42\linewidth,{trim=0in 0in 0in 0in,
  clip=true}]{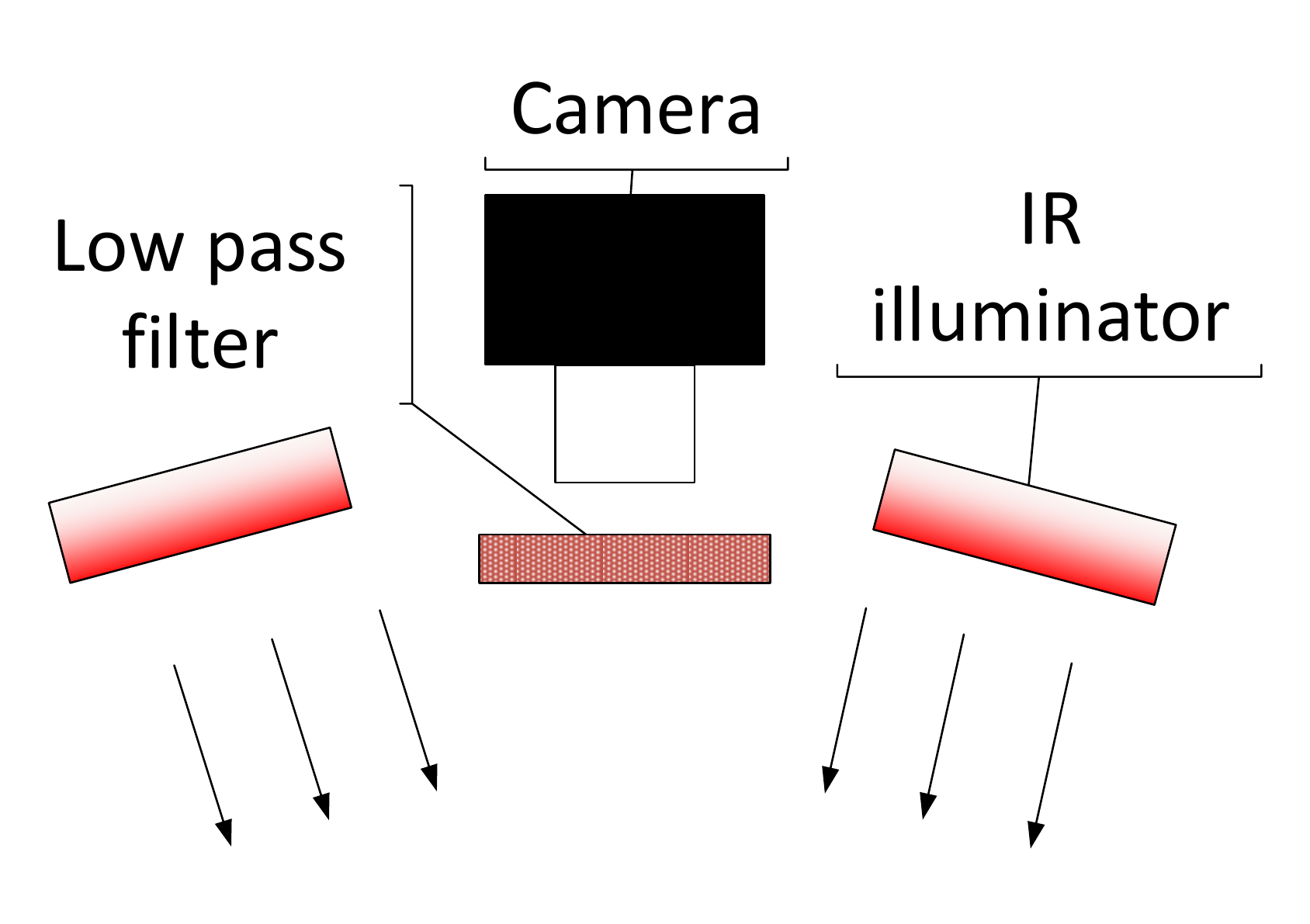}}
  \caption{Infrared selective (IRS) acquisition method, (a) IRS spectrum \cite{CCDSpectrum}, (b) IRS hardware diagram.}
\label{fig:IRS}
\vspace{-.2in}
\end{figure}
}

\newcommand{\figIRSsampling}{
\begin{figure}[t]
 \centering
 \subfloat[]{\label{fig:webLightOn}\includegraphics[width=0.42\linewidth, trim=0in 0in 0in 0in,
  clip=true]{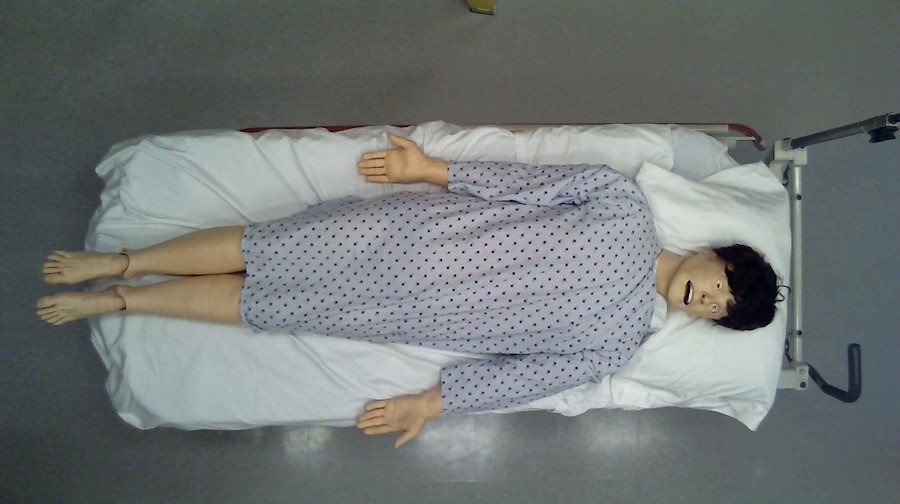}}
 \subfloat[]{\label{fig:webLightOff}\includegraphics[width=0.42\linewidth,{trim=0in 0in 0in 0in,
  clip=true}]{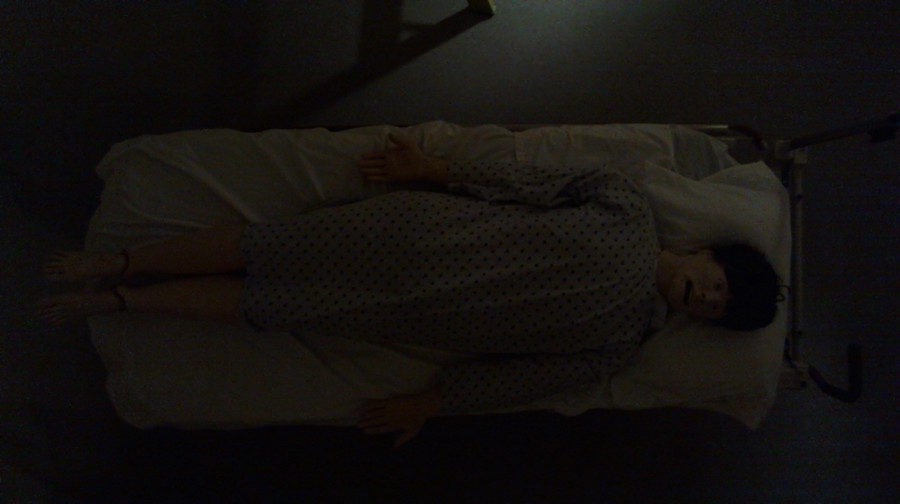}}
  \\
   \subfloat[]{\label{fig:IRSlightOn}\includegraphics[width=0.42\linewidth, trim=0in 0in 0in 0in,
  clip=true]{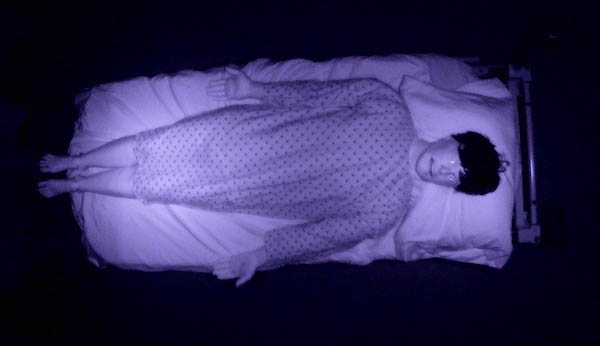}}
 \subfloat[]{\label{fig:IRSlightOff}\includegraphics[width=0.42\linewidth,{trim=0in 0in 0in 0in,
  clip=true}]{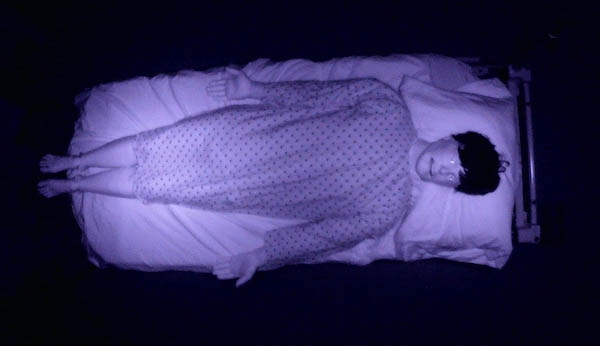}}
    \caption{Image captured by normal webcam (a) with light on and (b) with light off. The same images captured by IRS imaging system (c) with light on and (d) with light off.}
\label{fig:IRSsampling}
\end{figure}
}

\newcommand{\figCPMorient}{
\begin{figure}[t]
 \centering
 \subfloat[]{\label{fig:000001NLimbs}\includegraphics[width=0.25\linewidth,{trim=0in 0in 0in 0in,
  clip=true}]{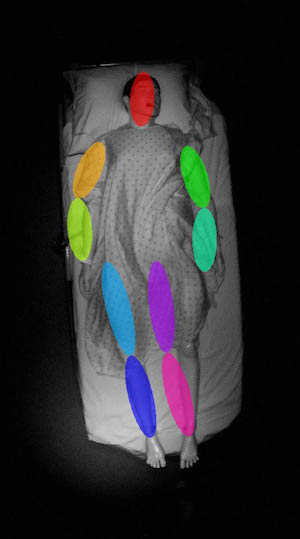}}
   \subfloat[]{\label{fig:000001SLimbs}\includegraphics[width=0.25\linewidth,{trim=0in 0in 0in 0in,
  clip=true}]{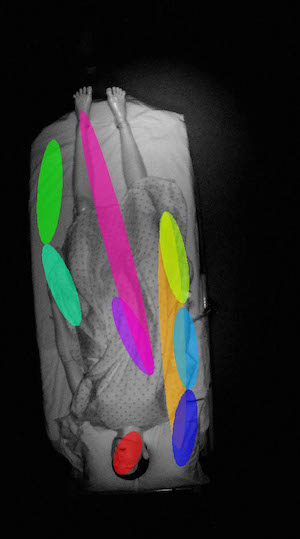}}
   \caption{Convolutional pose machine (CPM) detection result of same image with different orientations, sleeping position with the head (a) in the top of the image, and (b) in the bottom of the image.}
\label{fig:CPMorient}
\end{figure}
}

\newcommand{\figBB}{
\begin{figure}[h]
 \centering
 \subfloat[]{\label{fig:thrImBd}\includegraphics[width=0.25\linewidth,{trim=0in 0in 0in 0in,
  clip=true}]{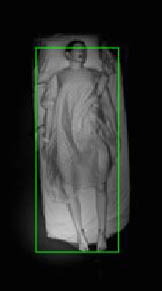}}
   \subfloat[]{\label{fig:thrIm}\includegraphics[width=0.25\linewidth,{trim=0in 0in 0in 0in,
  clip=true}]{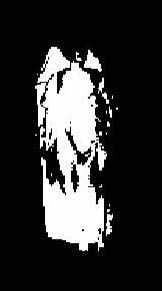}}
   \subfloat[]{\label{fig:edImBd}\includegraphics[width=0.25\linewidth,{trim=0in 0in 0in 0in,
  clip=true}]{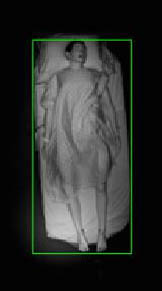}}
   \subfloat[]{\label{fig:edIm}\includegraphics[width=0.25\linewidth,{trim=0in 0in 0in 0in,
  clip=true}]{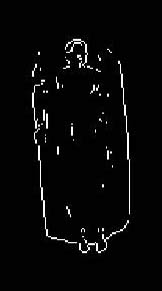}}
   \caption{Bounding box extraction using (a) threshold method, (b) binary image from thresholding, (c) the edge detection method, (d) edge detected with the 'Sobel' operator.}
\label{fig:BB}
\end{figure}
}

\newcommand{\figHogNend}{
\begin{figure}[t]
 \centering
 \subfloat[]{\label{fig:HogCenter}\includegraphics[width=0.25\linewidth,{trim=0in 0in 0in 0in,
  clip=true}]{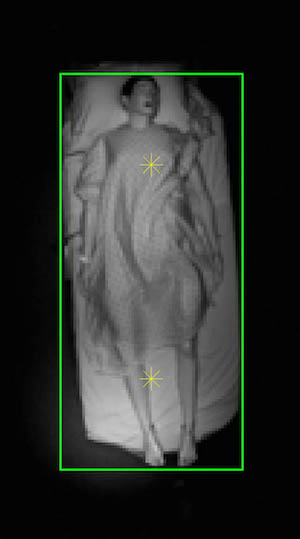}}
   \subfloat[]{\label{fig:Hog2end}\includegraphics[width=0.25\linewidth,{trim=0in 0in 0in 0in,
  clip=true}]{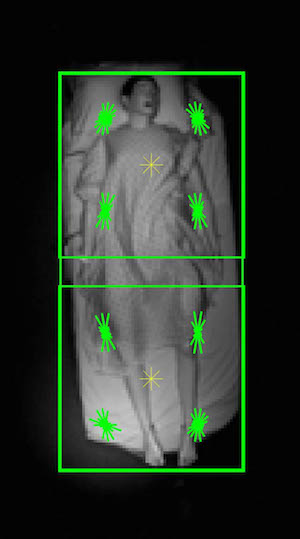}}
   \caption{2-end HOG feature extraction, (a) candidate HOG center locations, (b) HOG features extracted from the candidate locations.}
\label{fig:HogNend}
\end{figure}
}

\newcommand{\figIRSex}{
\begin{figure}[t]
 \centering
 \subfloat[]{\label{fig:IRSdevice}\includegraphics[width=0.42\linewidth,{trim=0in 0in 0in 0in,
  clip=true}]{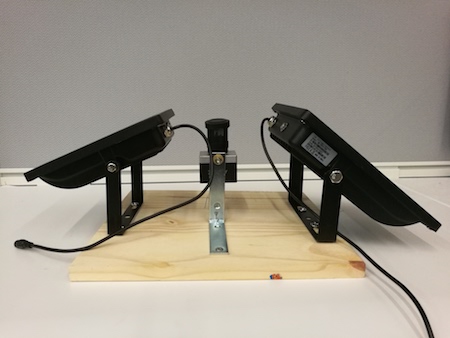}}
   \subfloat[]{\label{fig:exSetup}\includegraphics[width=0.42\linewidth,{trim=0in 0in 0in 0in,
  clip=true}]{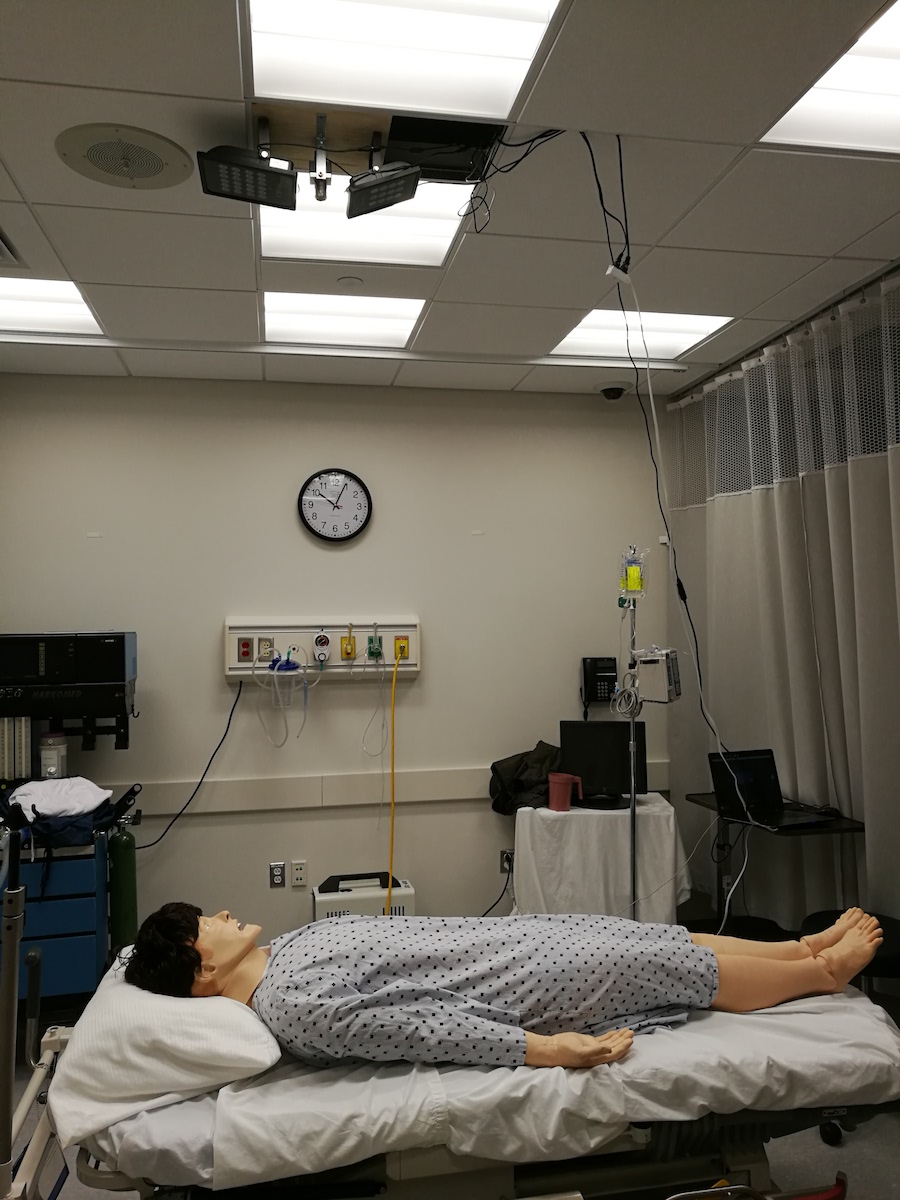}}
   \caption{IRS data collection setup, (a) customized IRS imaging device, (b) experimental setup in a simulated hospital room at Health Science Department of Northeastern University.}
\label{fig:IRSex}
\end{figure}
}

\newcommand{\figSystem}{
\begin{figure*}[h]
 \centering
 \includegraphics[width=0.84\linewidth,{trim=0in 0in 0in 0in,
  clip=true}]{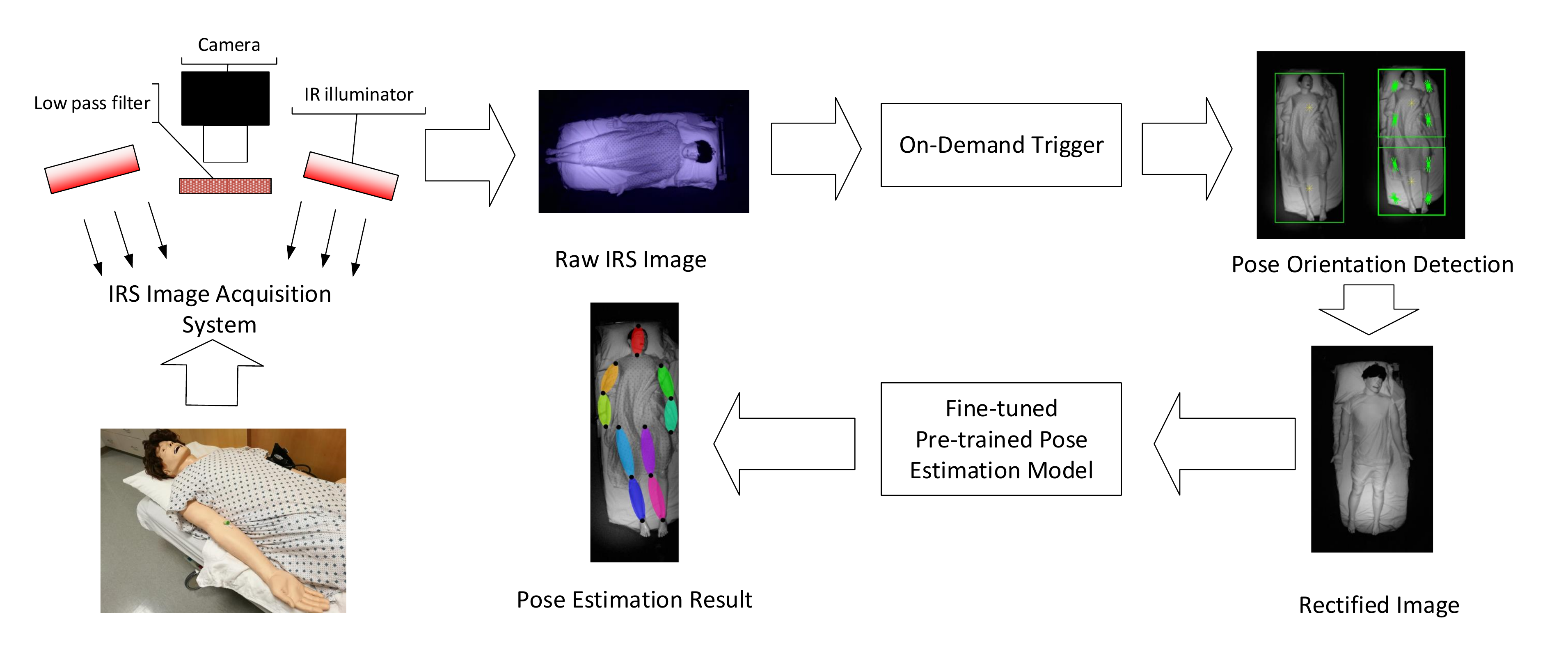}
 \caption{Overview of our in-bed human pose estimation system. In-bed images are collected from the proposed IRS system, then based on the system user's demand, pose estimation routine is triggered. Raw images are first preprocessed by a rectification method to get rectified and then fed into a fine-tuned pre-trained pose estimation model to produce pose estimation results.}
\label{fig:system}
\vspace{-.2in}
\end{figure*}
}

\newcommand{\figSampImg}{
\begin{figure}[t]
 \centering
 \subfloat[]{\label{fig:000001}\includegraphics[trim=0in 0in 0in 0in,clip,width=0.16\linewidth]{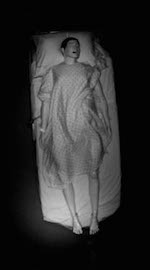}}
 \subfloat[]{\label{fig:000011}\includegraphics[trim=0in 0in 0in 0in,clip,width=0.16\linewidth]{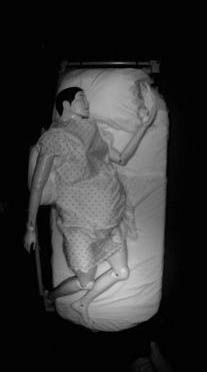}}
  \subfloat[]{\label{fig:000021}\includegraphics[trim=0in 0in 0in 0in,clip,width=0.16\linewidth]{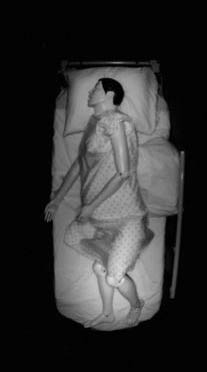}}
  \subfloat[]{\label{fig:000376}\includegraphics[trim=0in 0in 0in 0in,clip,width=0.16\linewidth]{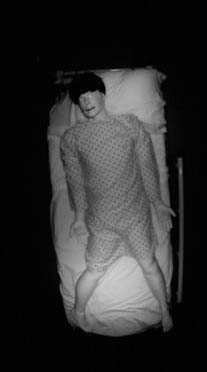}}
  \subfloat[]{\label{fig:000401}\includegraphics[trim=0in 0in 0in 0in,clip,width=0.16\linewidth]{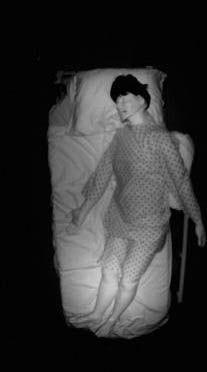}}
  \subfloat[]{\label{fig:000416}\includegraphics[trim=0in 0in 0in 0in,clip,width=0.16\linewidth]{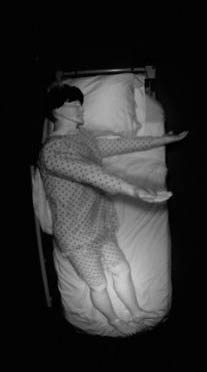}}
  \\
 \subfloat[]{\label{fig:000001LB}\includegraphics[trim=0in 0in 0in 0in,clip,width=0.16\linewidth]{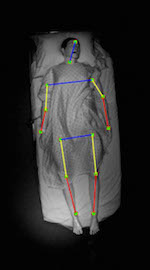}}
 \subfloat[]{\label{fig:000011LB}\includegraphics[trim=0in 0in 0in 0in,clip,width=0.16\linewidth]{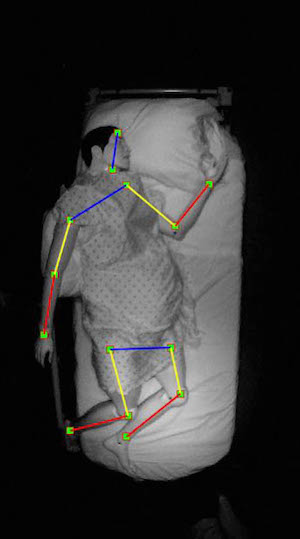}}
  \subfloat[]{\label{fig:000021LB}\includegraphics[trim=0in 0in 0in 0in,clip,width=0.16\linewidth]{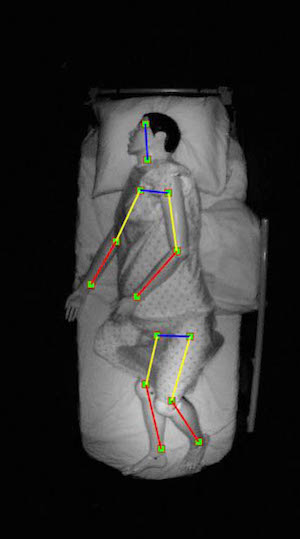}}
  \subfloat[]{\label{fig:000376LB}\includegraphics[trim=0in 0in 0in 0in,clip,width=0.16\linewidth]{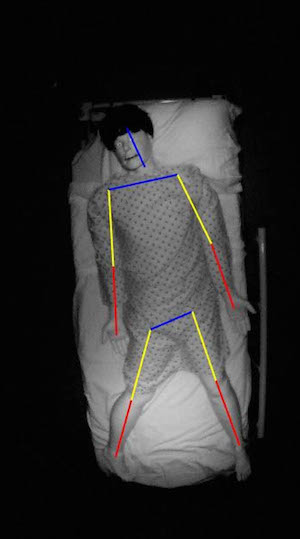}}
  \subfloat[]{\label{fig:000401LB}\includegraphics[trim=0in 0in 0in 0in,clip,width=0.16\linewidth]{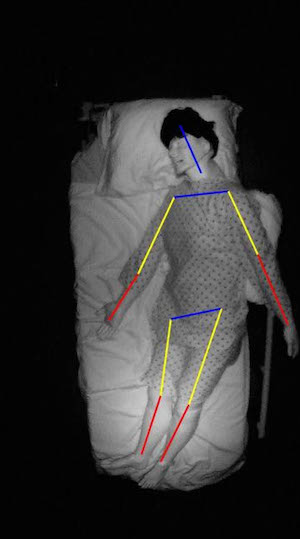}}
  \subfloat[]{\label{fig:000416LB}\includegraphics[trim=0in 0in 0in 0in,clip,width=0.16\linewidth]{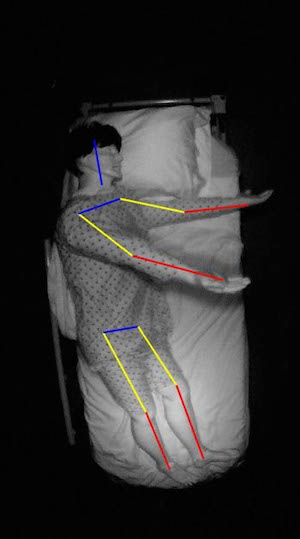}}
  \caption{Mannequin pose dataset collected in a simulated hospital room. First row images show the raw image collected via IRS system. Second row shows manually annotation pose results of first row images.}
\label{fig:smpImg}
\vspace{-.2in}
\end{figure}
}

\newcommand{\figSampImgColor}{
\begin{figure}[t]
 \centering
 \subfloat[]{\label{fig:000137c}\includegraphics[width=0.16\linewidth,{trim=0in 0in 0in 0in,
  clip=true}]{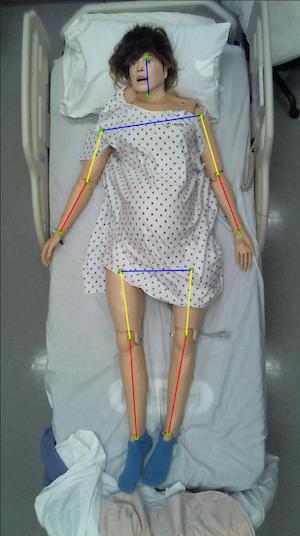}}
 \subfloat[]{\label{fig:000225c}\includegraphics[width=0.16\linewidth,{trim=0in 0in 0in 0in,
  clip=true}]{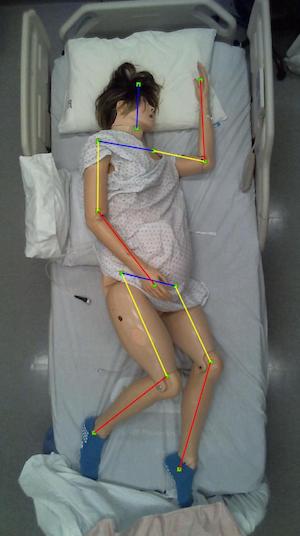}}
   \subfloat[]{\label{fig:000150c}\includegraphics[width=0.16\linewidth,{trim=0in 0in 0in 0in,
  clip=true}]{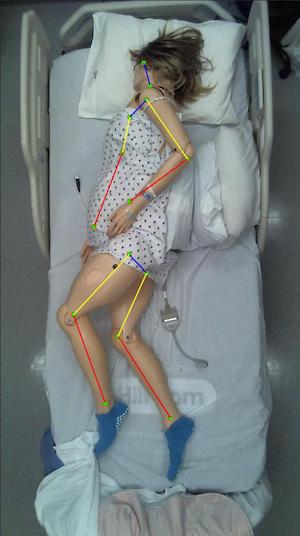}}
   \subfloat[]{\label{fig:000014c}\includegraphics[width=0.16\linewidth,{trim=0in 0in 0in 0in,
  clip=true}]{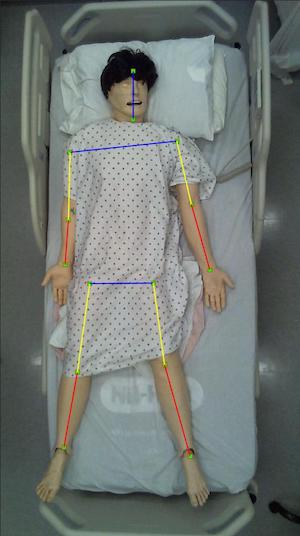}}
   \subfloat[]{\label{fig:000063c}\includegraphics[width=0.16\linewidth,{trim=0in 0in 0in 0in,
  clip=true}]{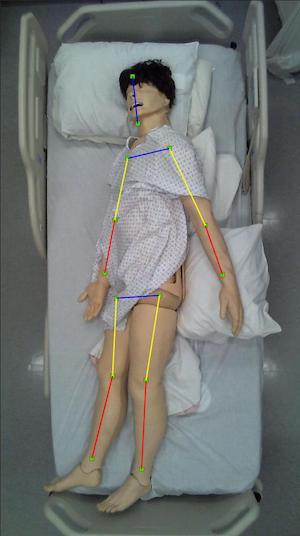}}
   \subfloat[]{\label{fig:000125c}\includegraphics[width=0.16\linewidth,{trim=0in 0in 0in 0in,
  clip=true}]{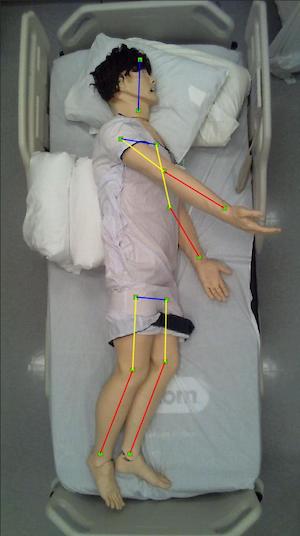}}
   \caption{Annotated mannequin pose samples collect via webcam system in a simulated hospital room.}
\label{fig:sampleImgColor}
\vspace{-.2in}
\end{figure}
}

\newcommand{\figTestBC}{
\begin{figure*}[t]
 \centering
 \subfloat[]{\label{fig:pckTotalBC}\includegraphics[width=0.24\linewidth,{trim=0in 0in 0in 0in,
  clip=true}]{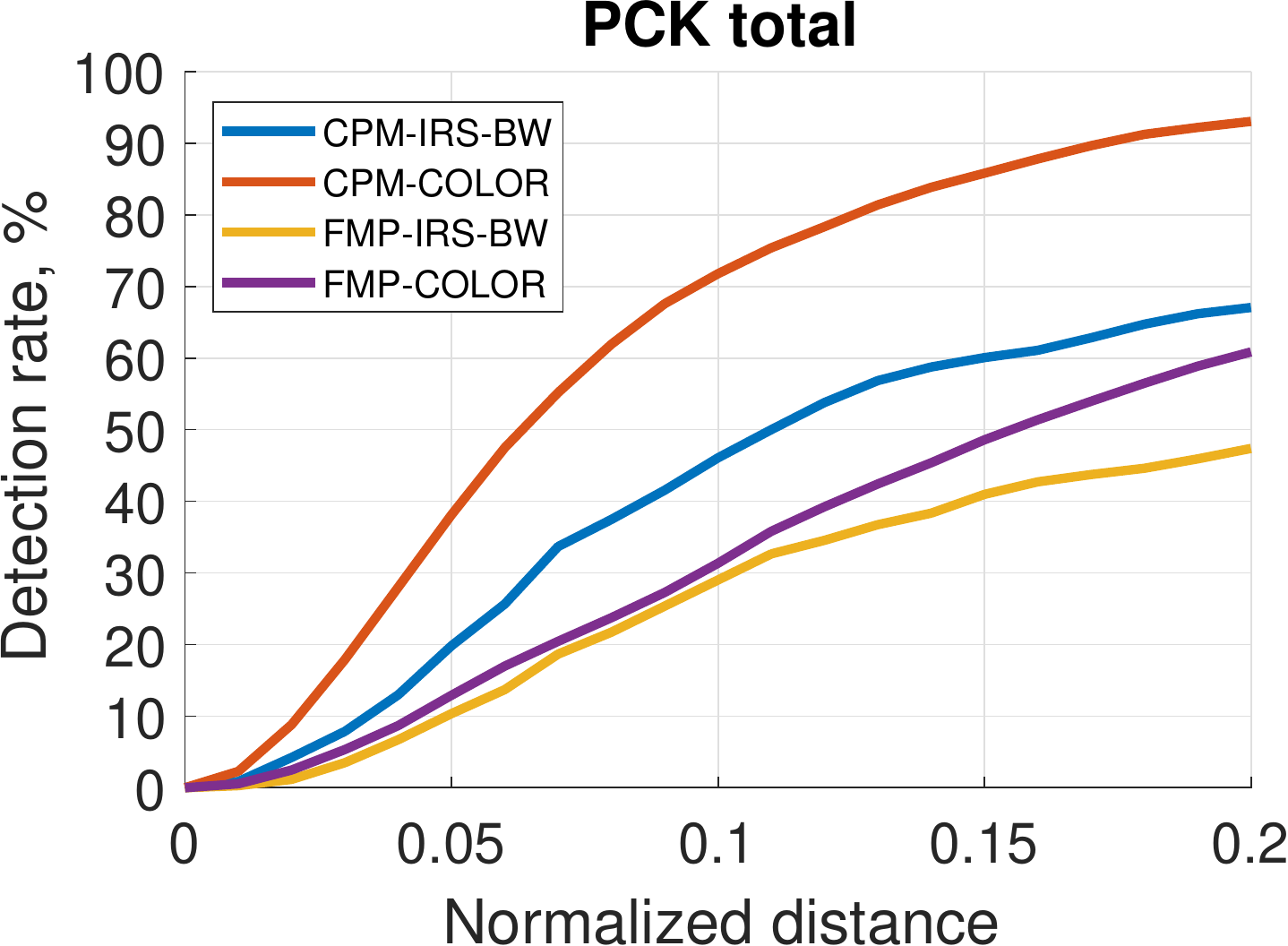}}
   \subfloat[]{\label{fig:pckHipBC}\includegraphics[width=0.24\linewidth,{trim=0in 0in 0in 0in,
  clip=true}]{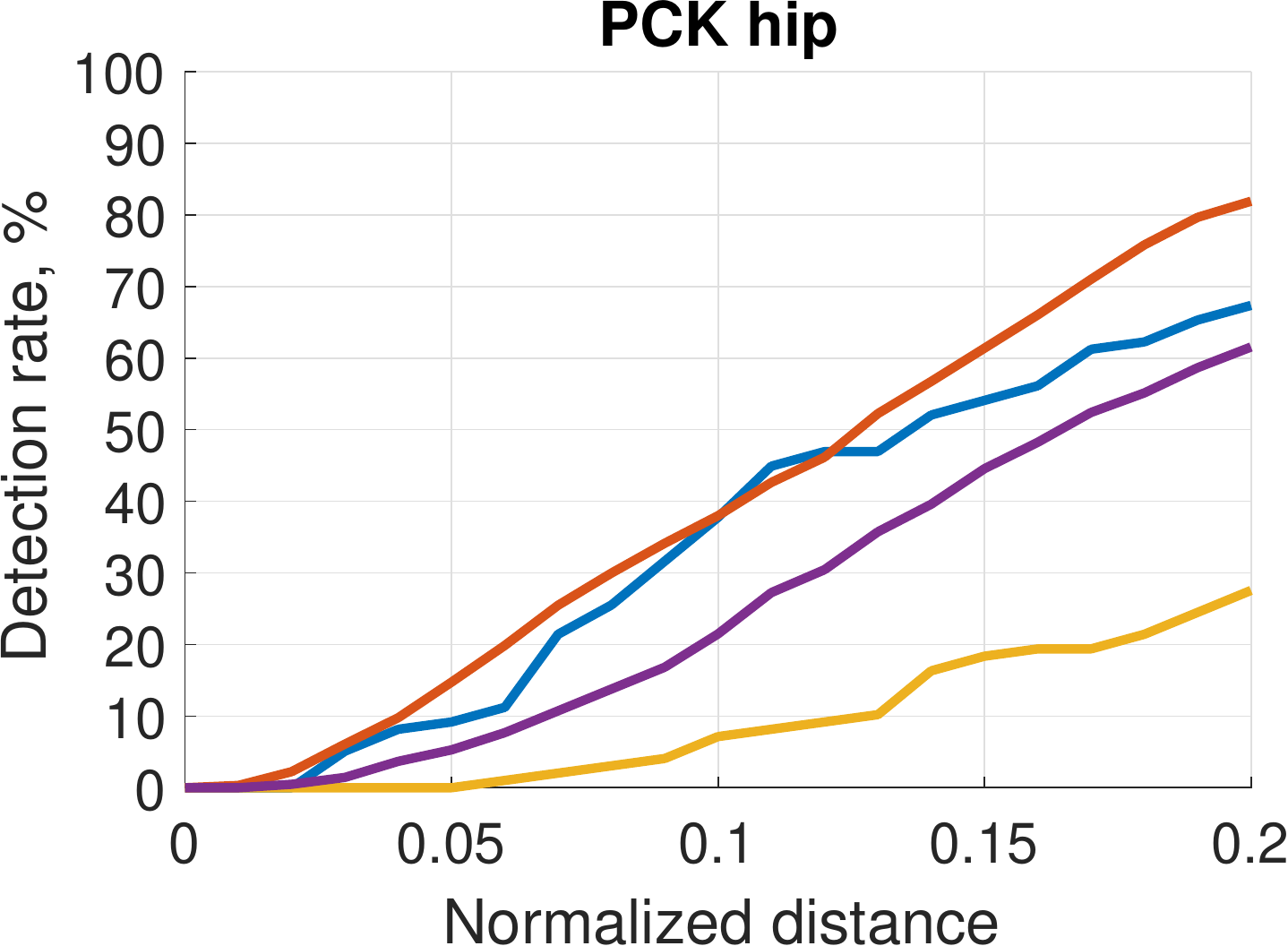}}
   \subfloat[]{\label{fig:pckKneeBC}\includegraphics[width=0.24\linewidth,{trim=0in 0in 0in 0in,
  clip=true}]{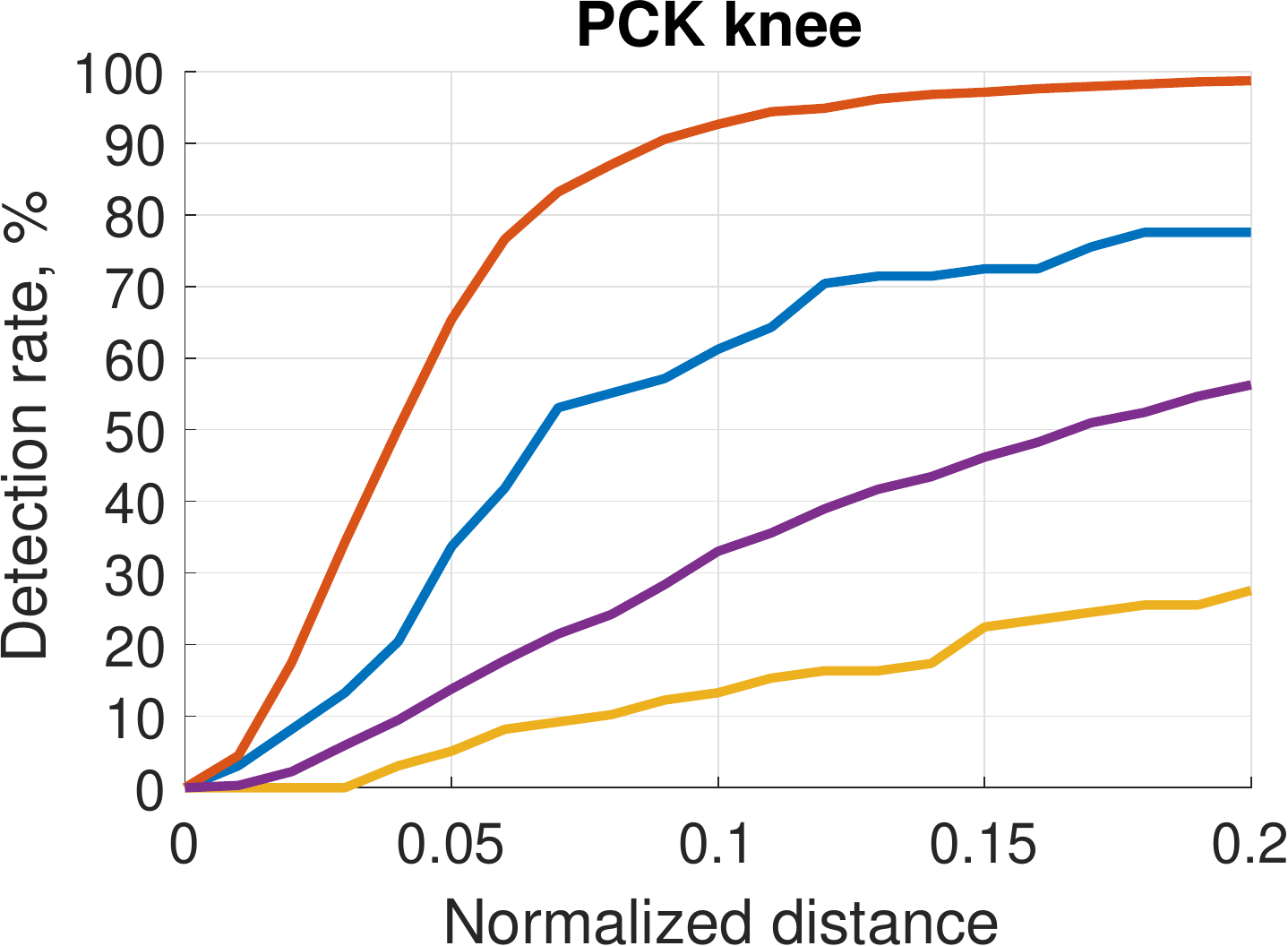}}
   \subfloat[]{\label{fig:pckAnkleBC}\includegraphics[width=0.24\linewidth,{trim=0in 0in 0in 0in,
  clip=true}]{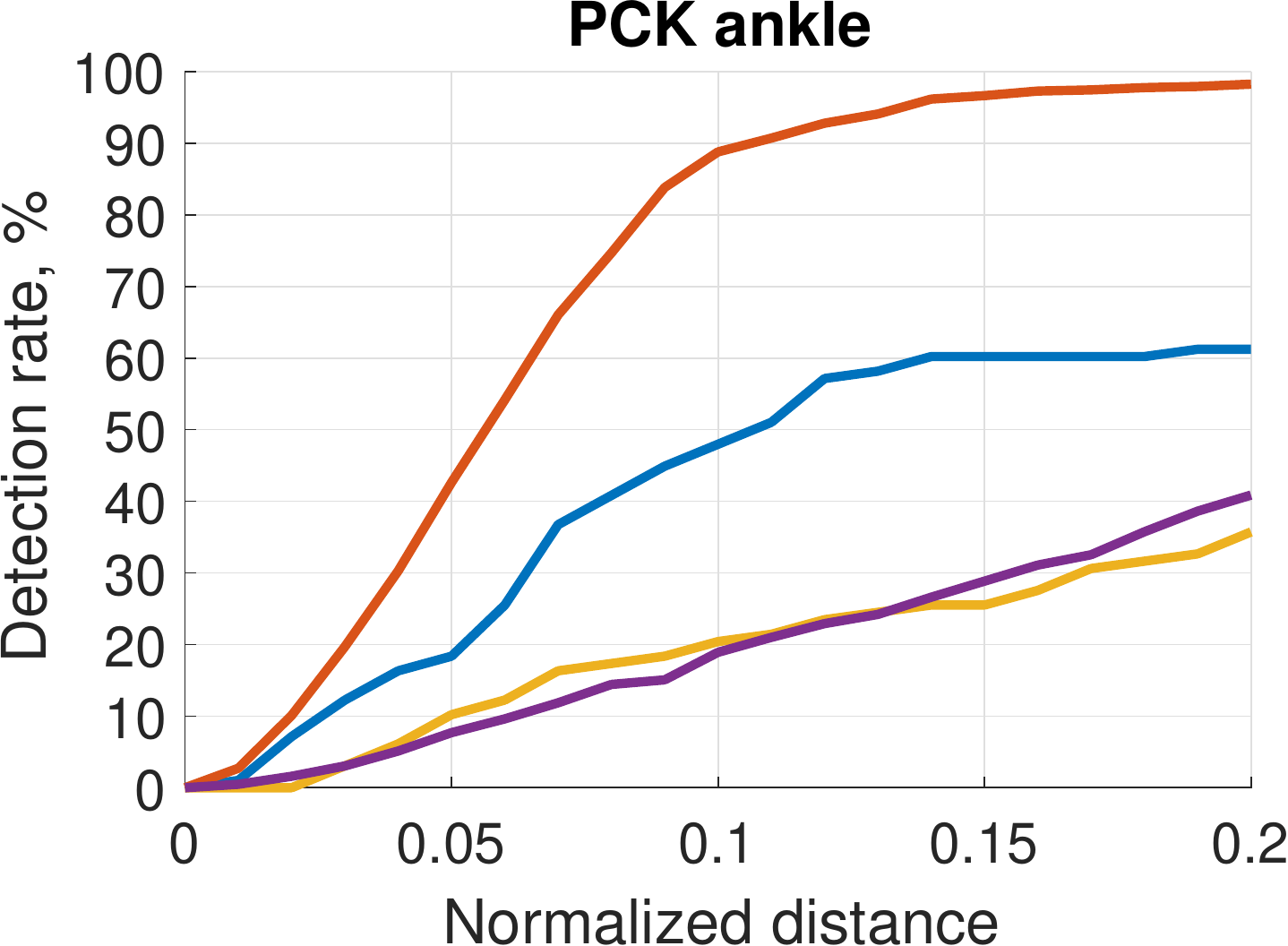}} 
  \\
   \subfloat[]{\label{fig:pckHeadBC}\includegraphics[width=0.24\linewidth,{trim=0in 0in 0in 0in,
  clip=true}]{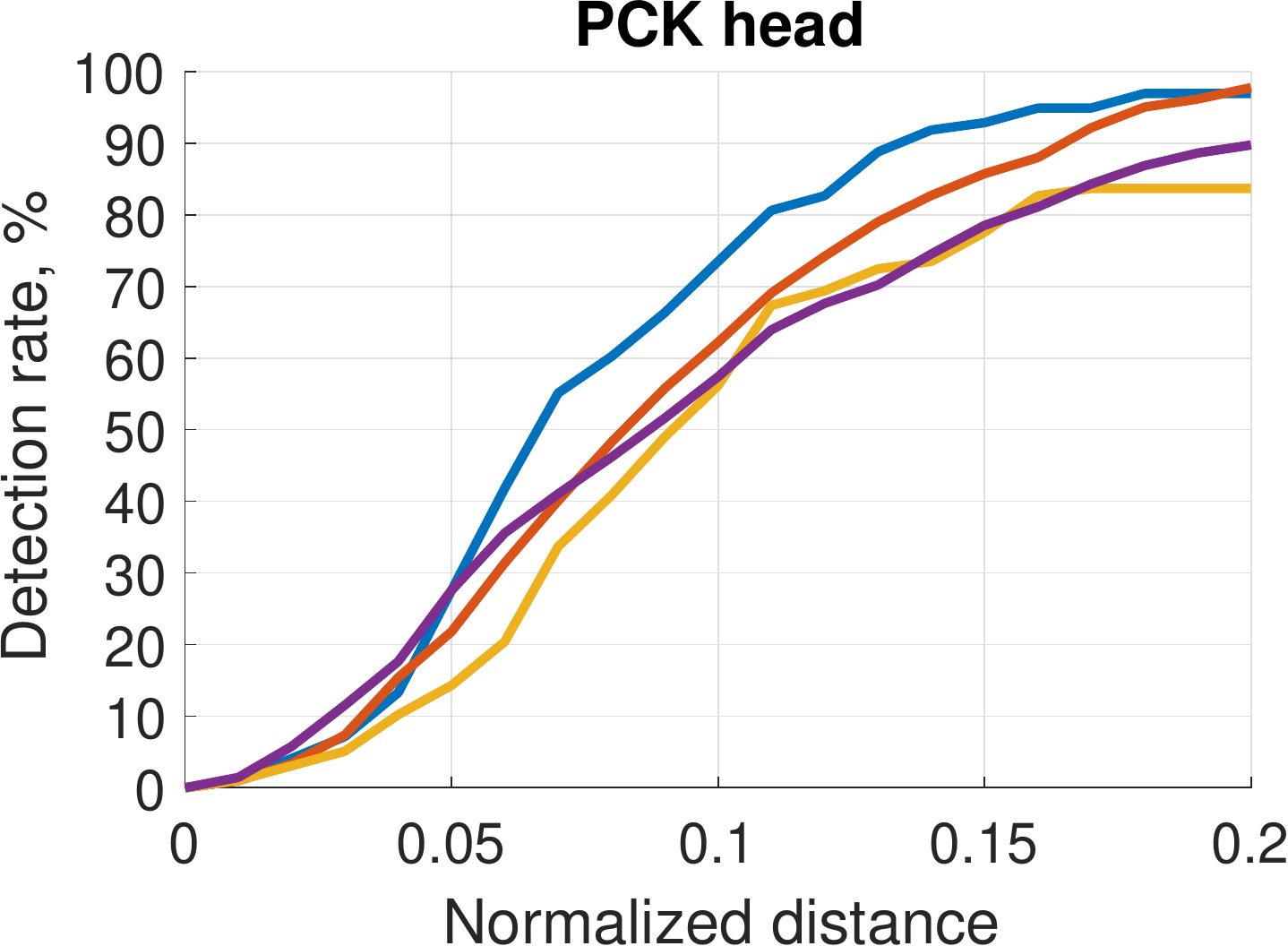}}
   \subfloat[]{\label{fig:pckShoulderBC}\includegraphics[width=0.24\linewidth,{trim=0in 0in 0in 0in,
  clip=true}]{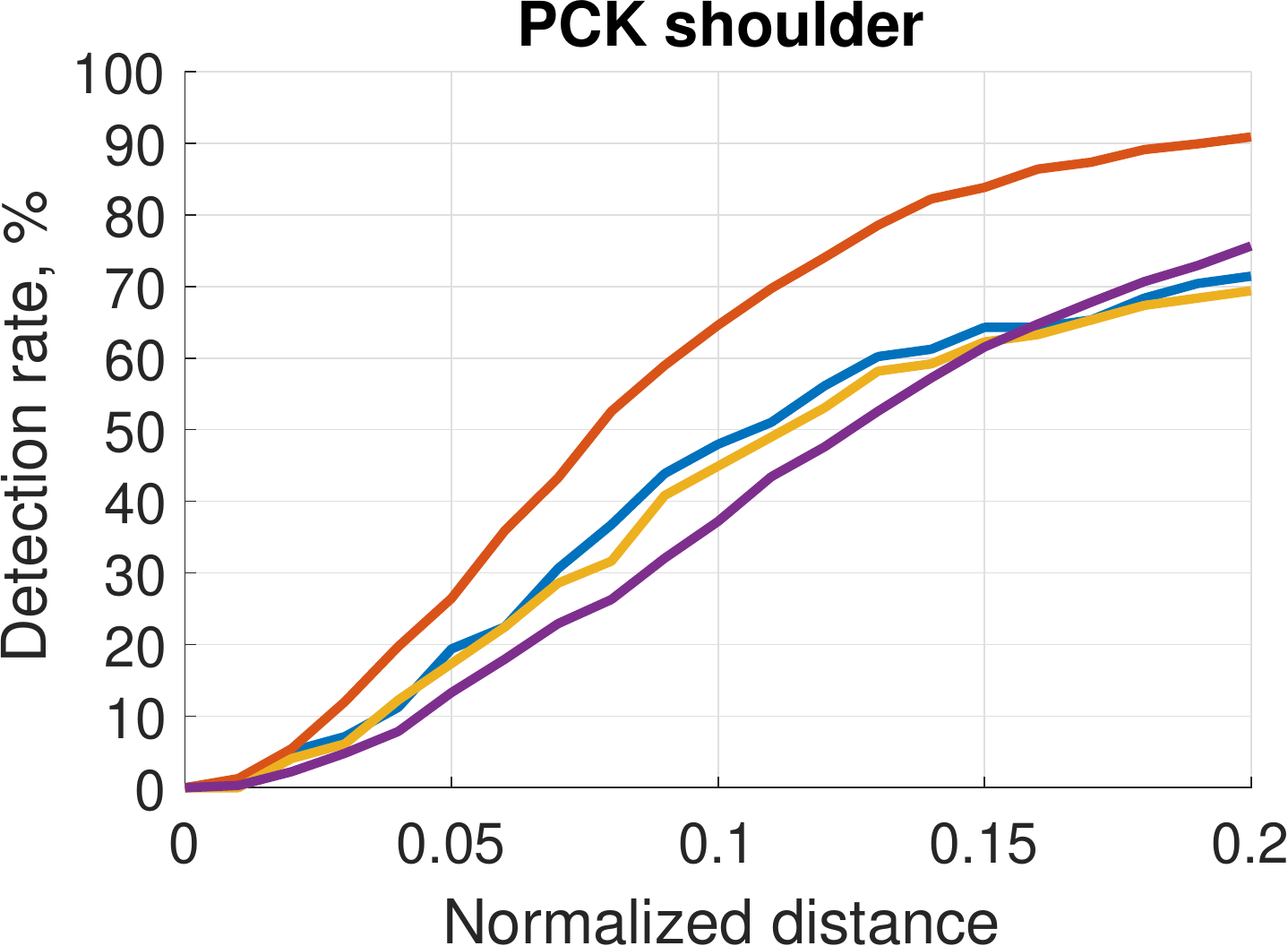}}
   \subfloat[]{\label{fig:pckElbowBC}\includegraphics[width=0.24\linewidth,{trim=0in 0in 0in 0in,
  clip=true}]{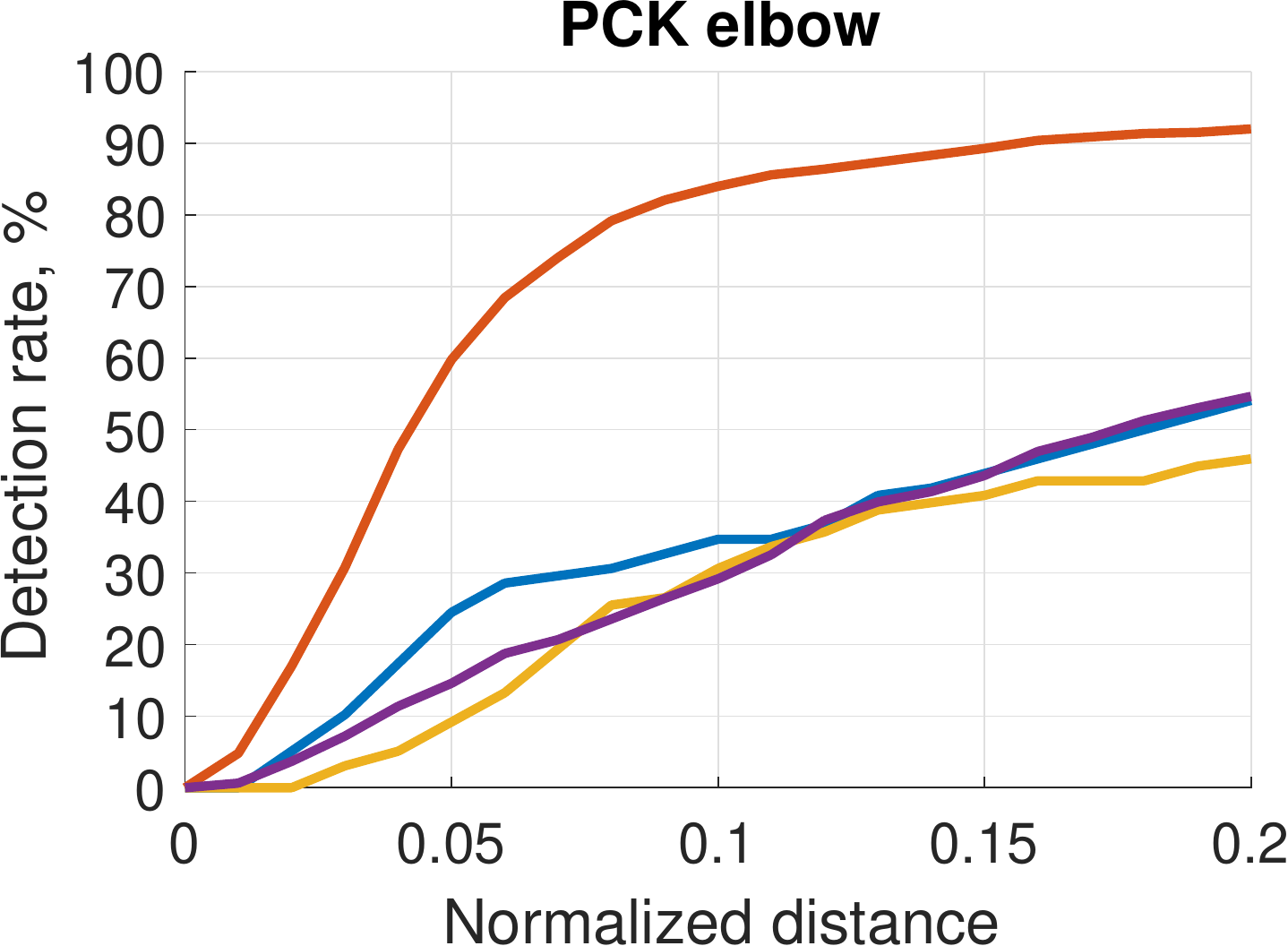}}
   \subfloat[]{\label{fig:pckWristBC}\includegraphics[width=0.24\linewidth,{trim=0in 0in 0in 0in,
  clip=true}]{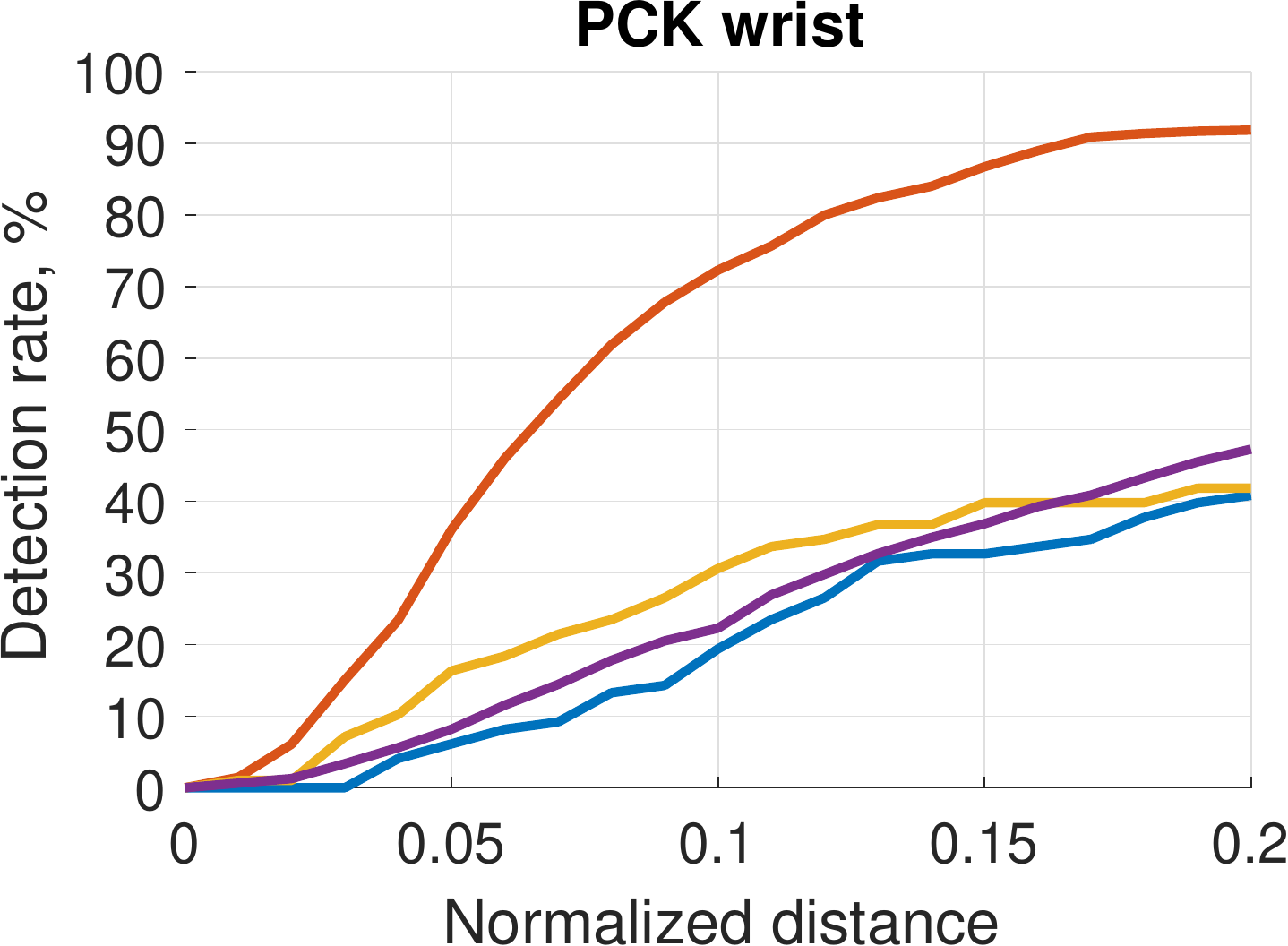}} 
   \caption{Quantitative posture estimation result with different IRS mannequin black and white dataset and webcam mannequin color dataset via MPII-LSP pre-trained CPM model and FMP model. All images are portrait view similar to normal view point to exclude the in-bed orientation factor. }
\label{fig:testBC}
\vspace{-.2in}
\end{figure*}
}

\newcommand{\figMultOrt}{
\begin{figure*}[t]
 \centering
 \subfloat[]{\label{fig:pckTotalOrt}\includegraphics[width=0.24\linewidth,{trim=0in 0in 0in 0in,
  clip=true}]{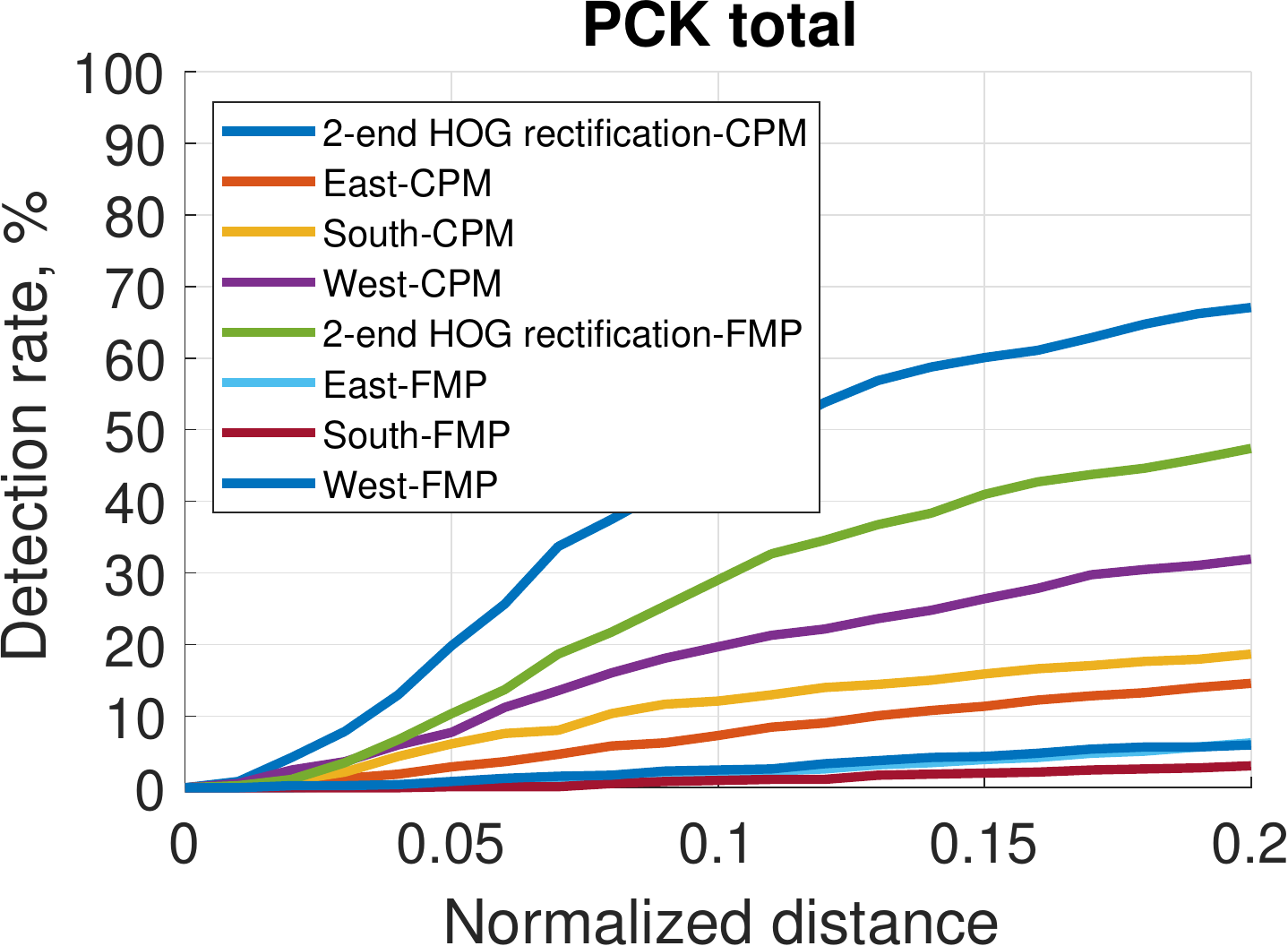}}
   \subfloat[]{\label{fig:pckHipOrt}\includegraphics[width=0.24\linewidth,{trim=0in 0in 0in 0in,
  clip=true}]{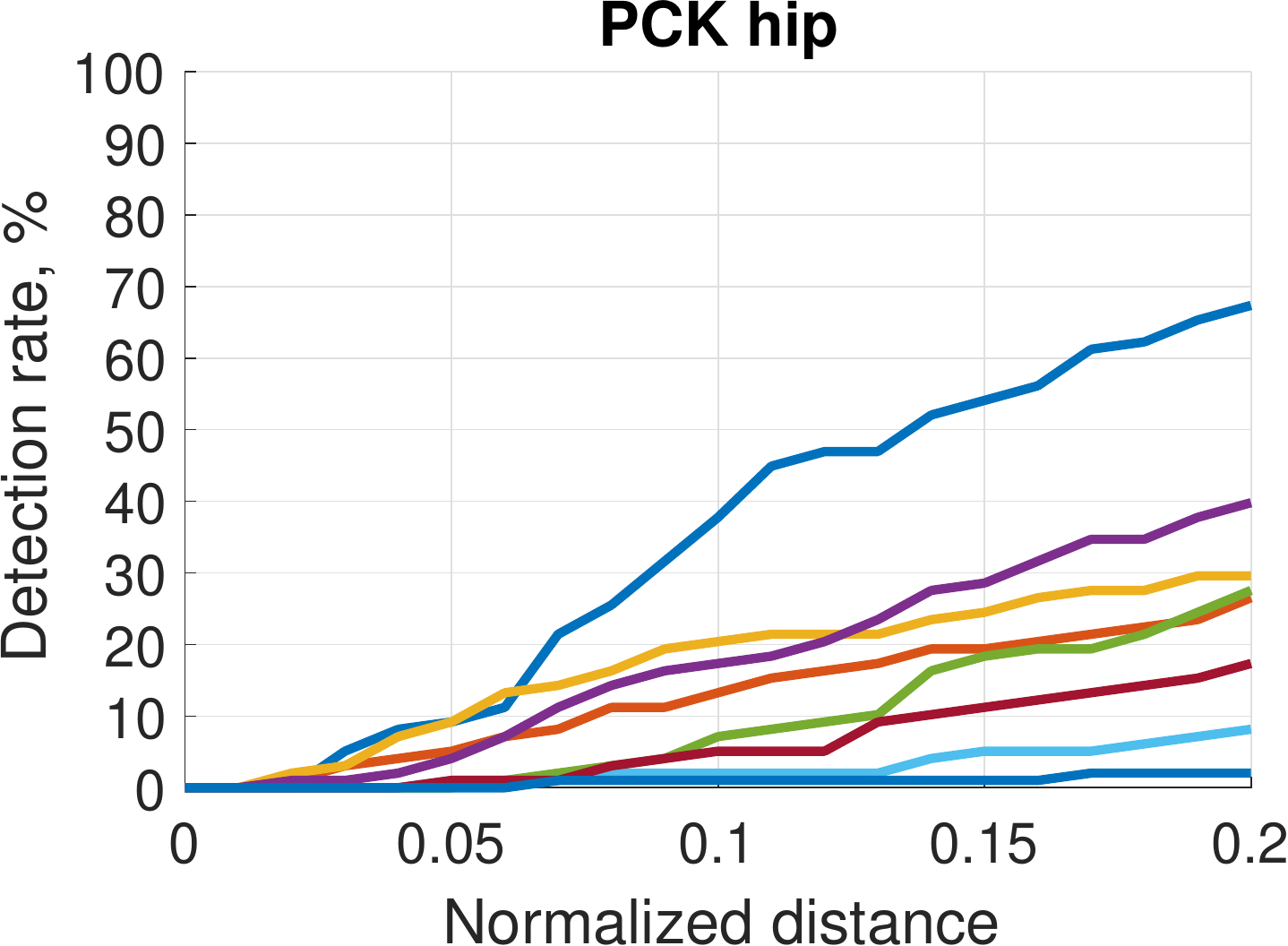}}
   \subfloat[]{\label{fig:pckKneeOrt}\includegraphics[width=0.24\linewidth,{trim=0in 0in 0in 0in,
  clip=true}]{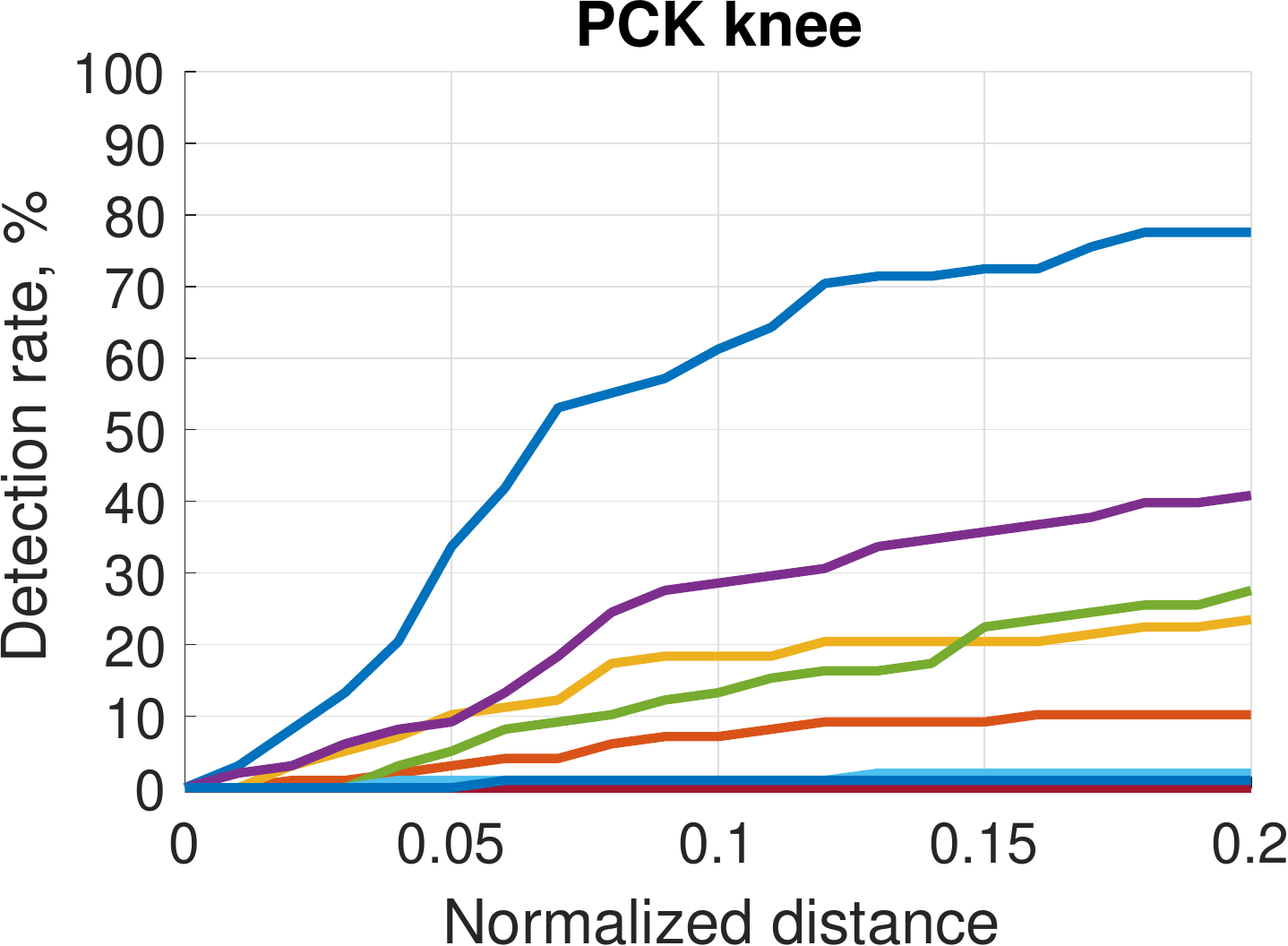}}
   \subfloat[]{\label{fig:pckAnkleOrt}\includegraphics[width=0.24\linewidth,{trim=0in 0in 0in 0in,
  clip=true}]{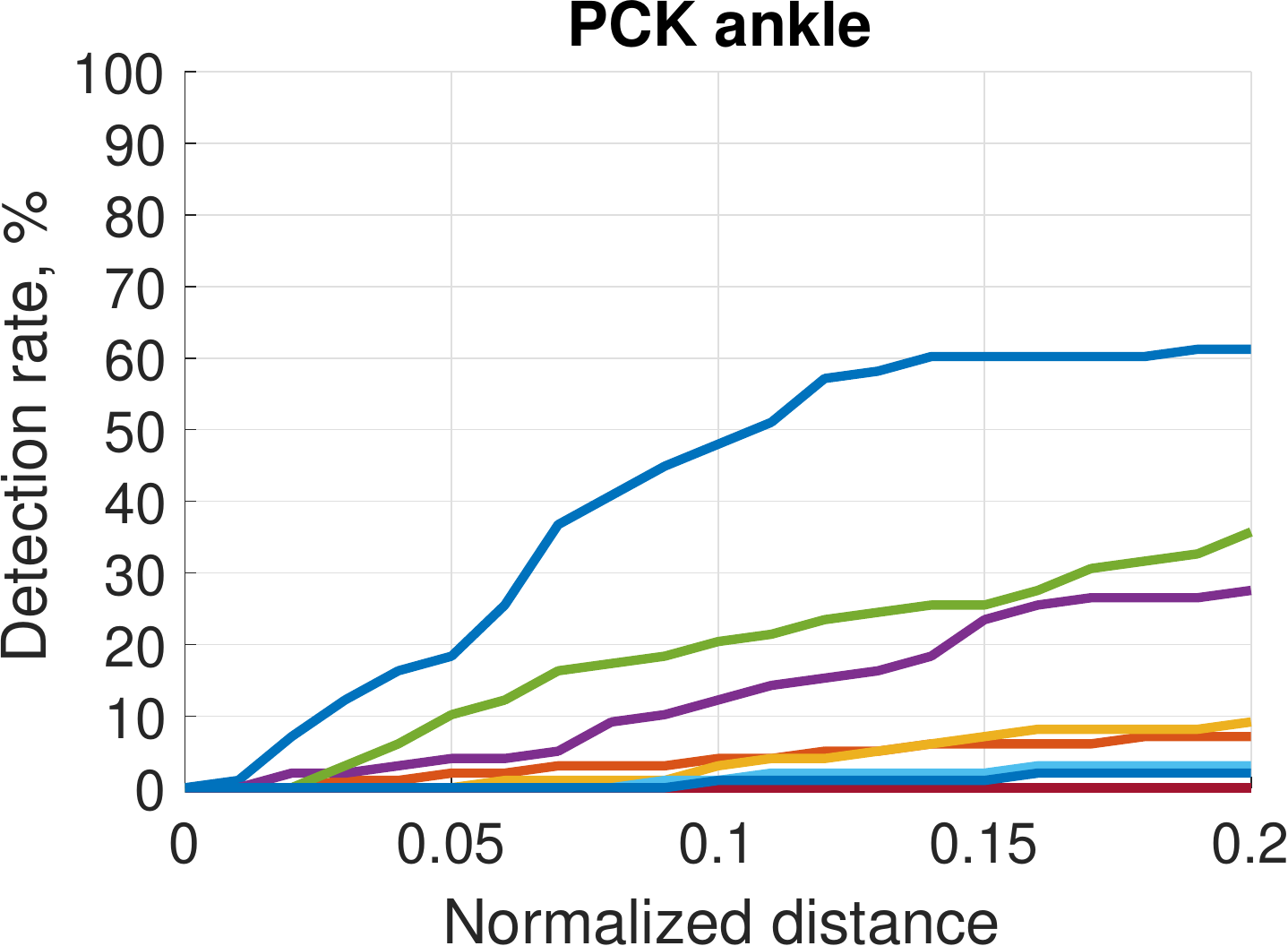}} 
  \\
   \subfloat[]{\label{fig:pckHeadOrt}\includegraphics[width=0.24\linewidth,{trim=0in 0in 0in 0in,
  clip=true}]{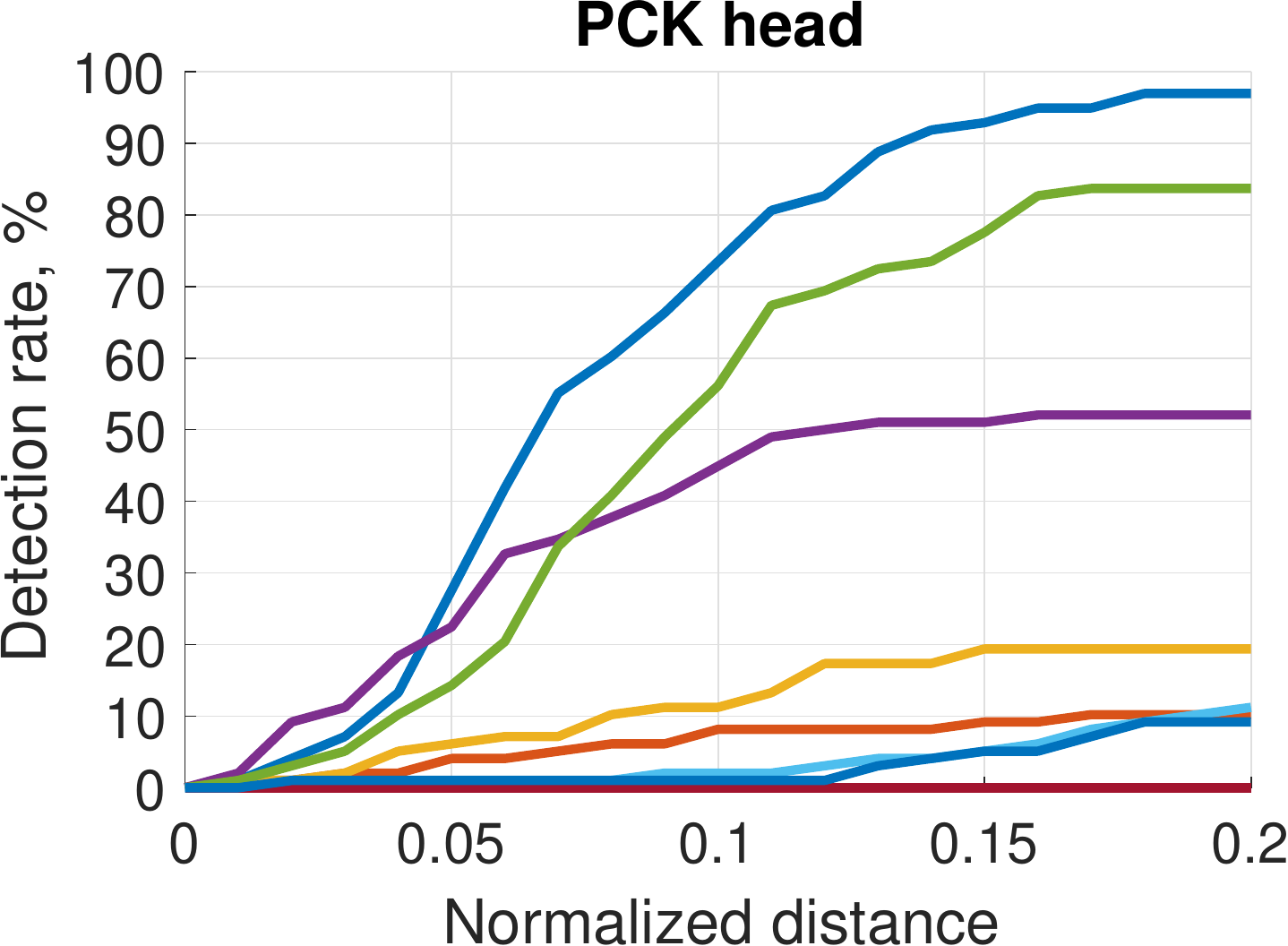}}
   \subfloat[]{\label{fig:pckShoulderOrt}\includegraphics[width=0.24\linewidth,{trim=0in 0in 0in 0in,
  clip=true}]{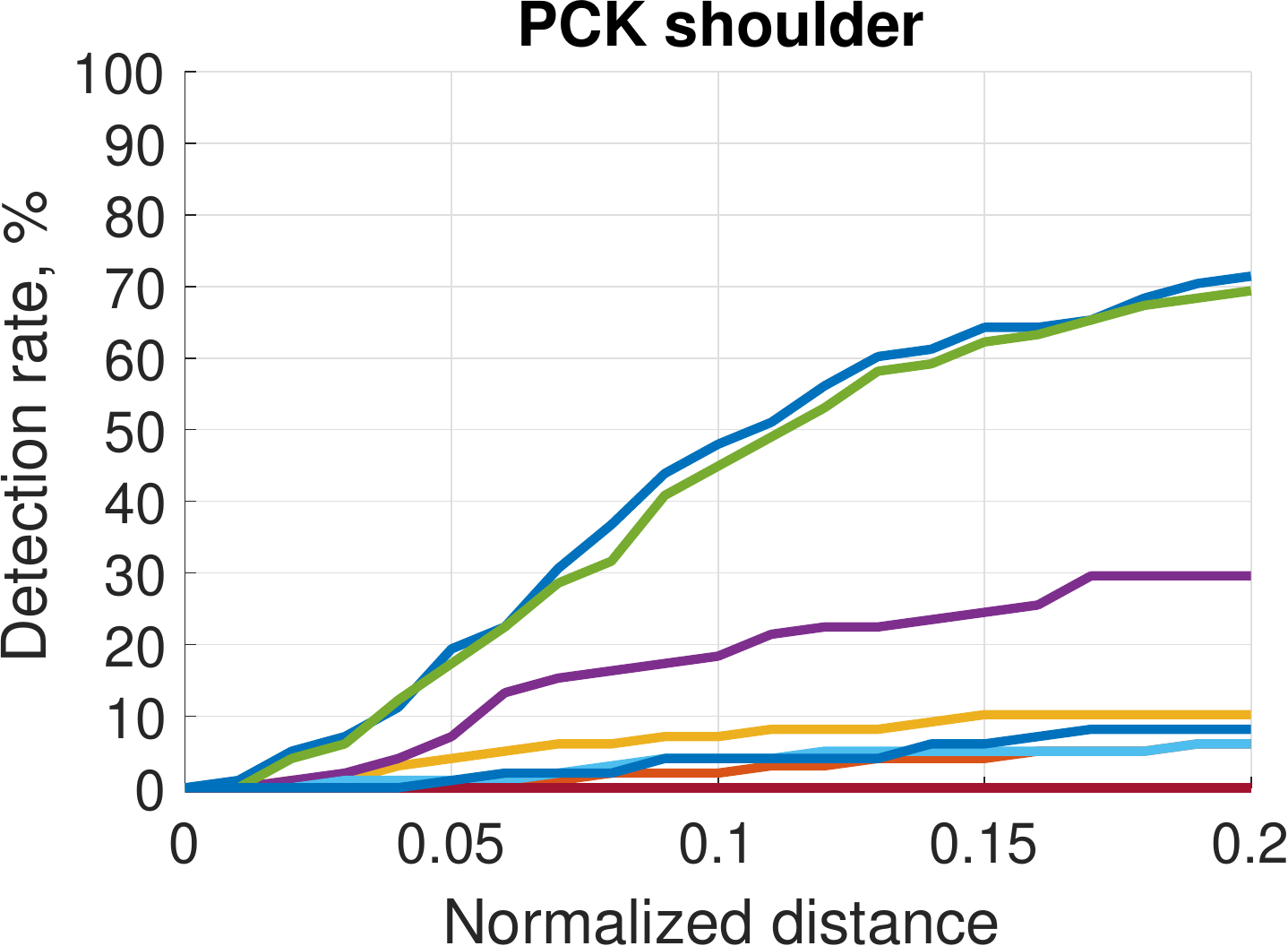}}
   \subfloat[]{\label{fig:pckElbowOrt}\includegraphics[width=0.24\linewidth,{trim=0in 0in 0in 0in,
  clip=true}]{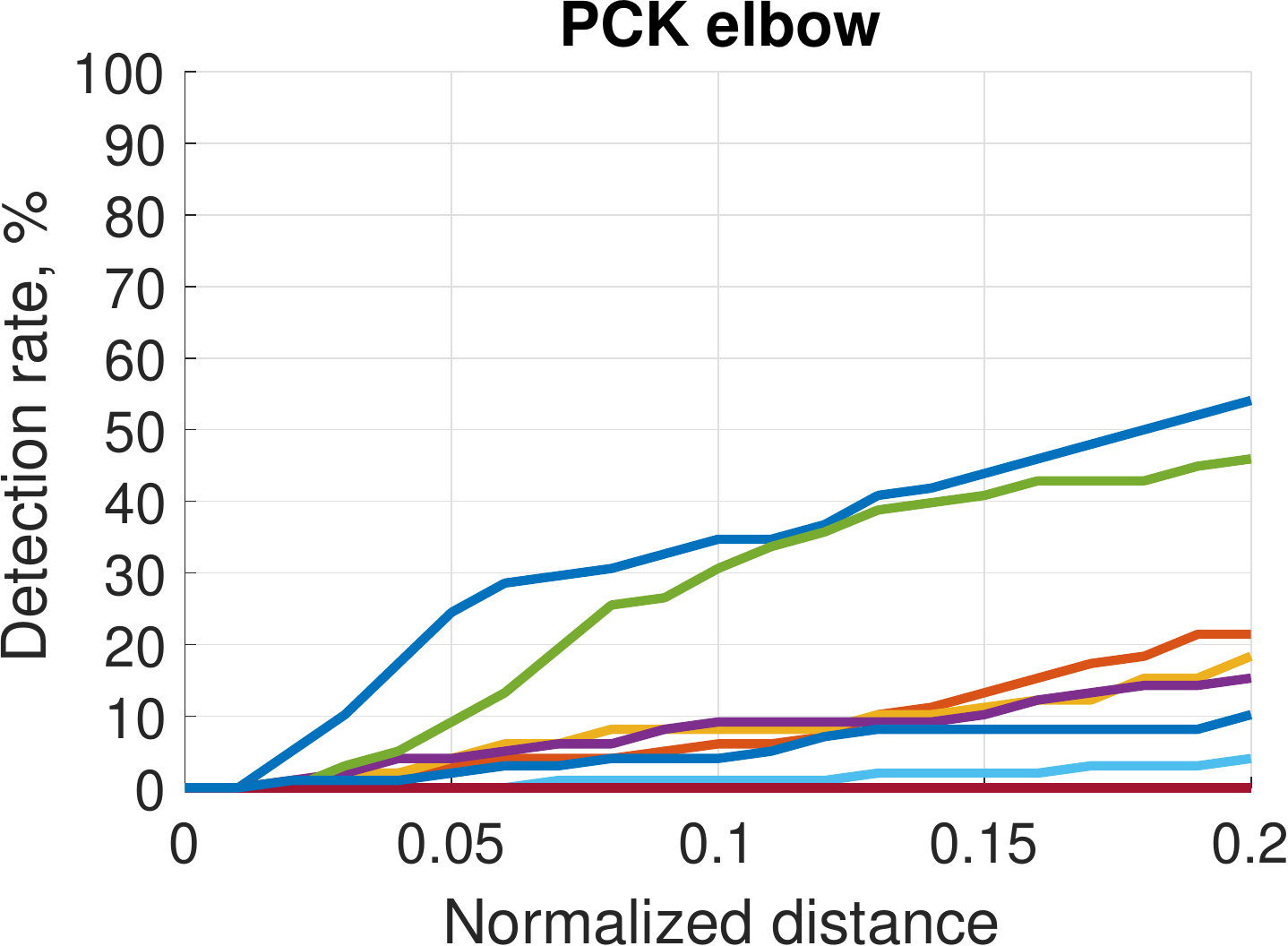}}
   \subfloat[]{\label{fig:pckWristOrt}\includegraphics[width=0.24\linewidth,{trim=0in 0in 0in 0in,
  clip=true}]{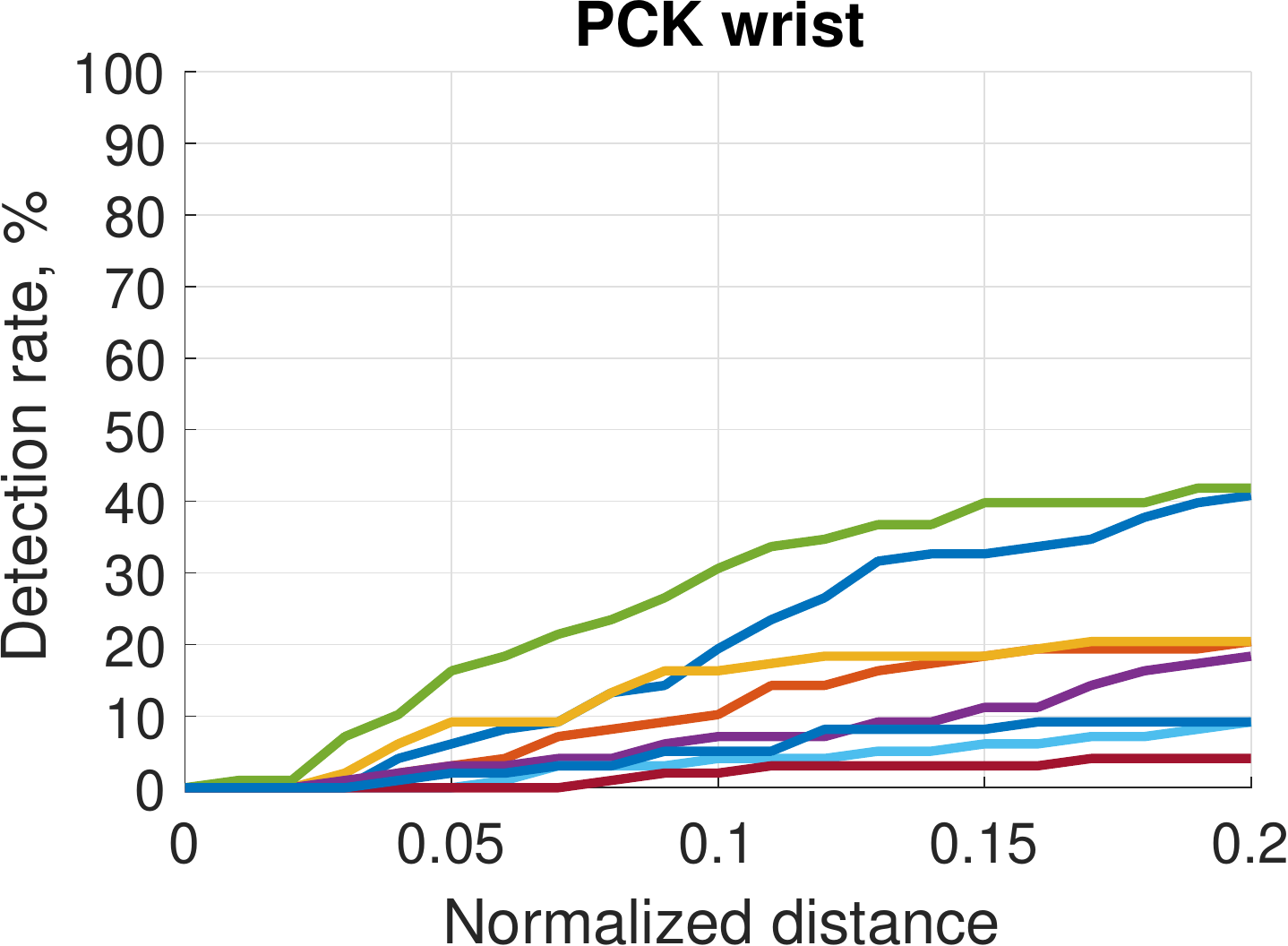}} 
   \caption{Quantitative posture estimation result via MPII-LSP pre-trained CPM model on different in-bed orientation images as well as their 2-end HOG rectified version.}
\label{fig:multOrt}
\vspace{-.2in}
\end{figure*}
}

\newcommand{\figMultMdl}{
\begin{figure*}[t]
 \centering
 \subfloat[]{\label{fig:pckTotalMdl}\includegraphics[width=0.24\linewidth,{trim=0in 0in 0in 0in,
  clip=true}]{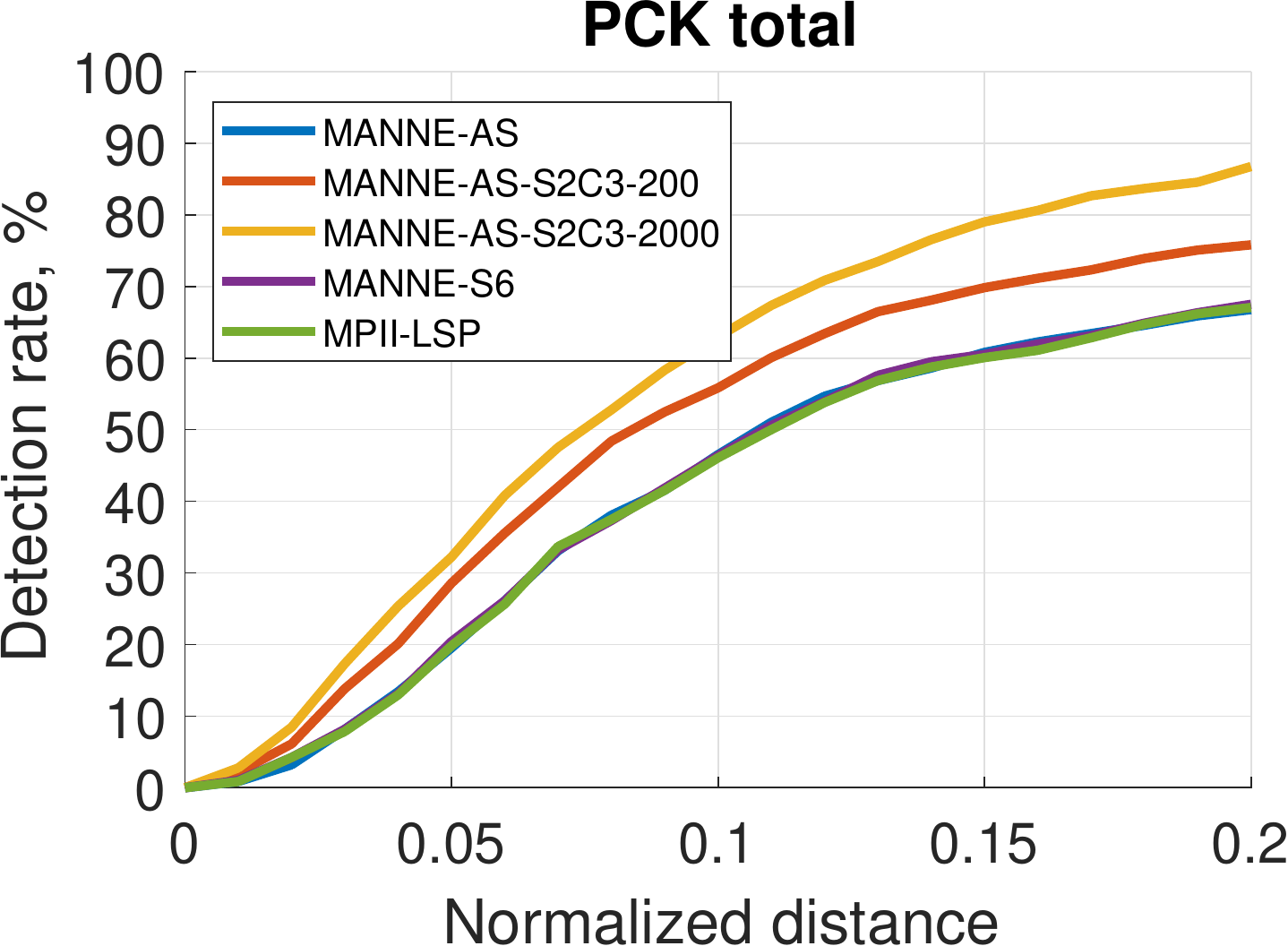}}
  \subfloat[]{\label{fig:pckHipMdl}\includegraphics[width=0.24\linewidth,{trim=0in 0in 0in 0in,
  clip=true}]{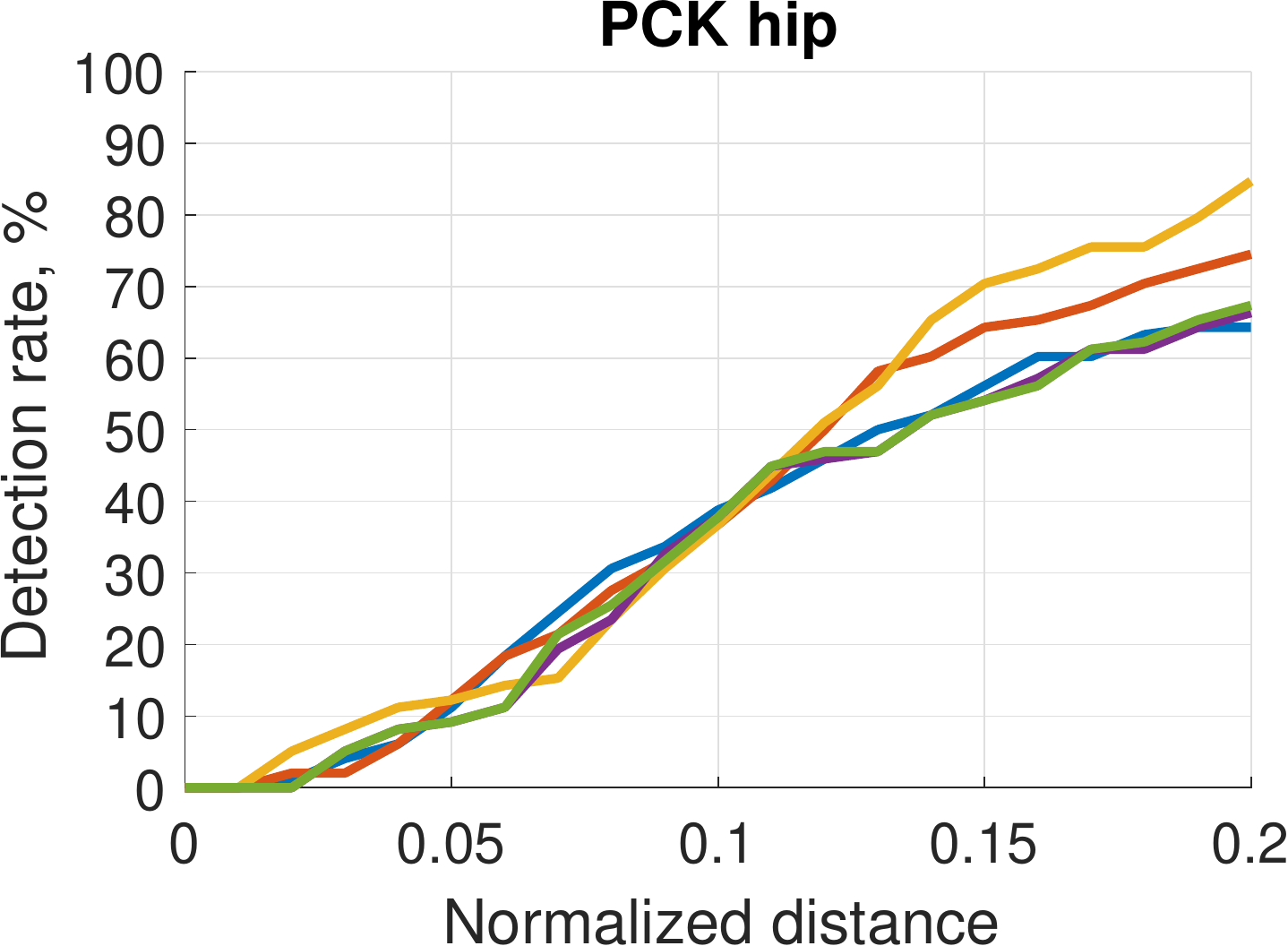}}
  \subfloat[]{\label{fig:pckKneeMdl}\includegraphics[width=0.24\linewidth,{trim=0in 0in 0in 0in,
  clip=true}]{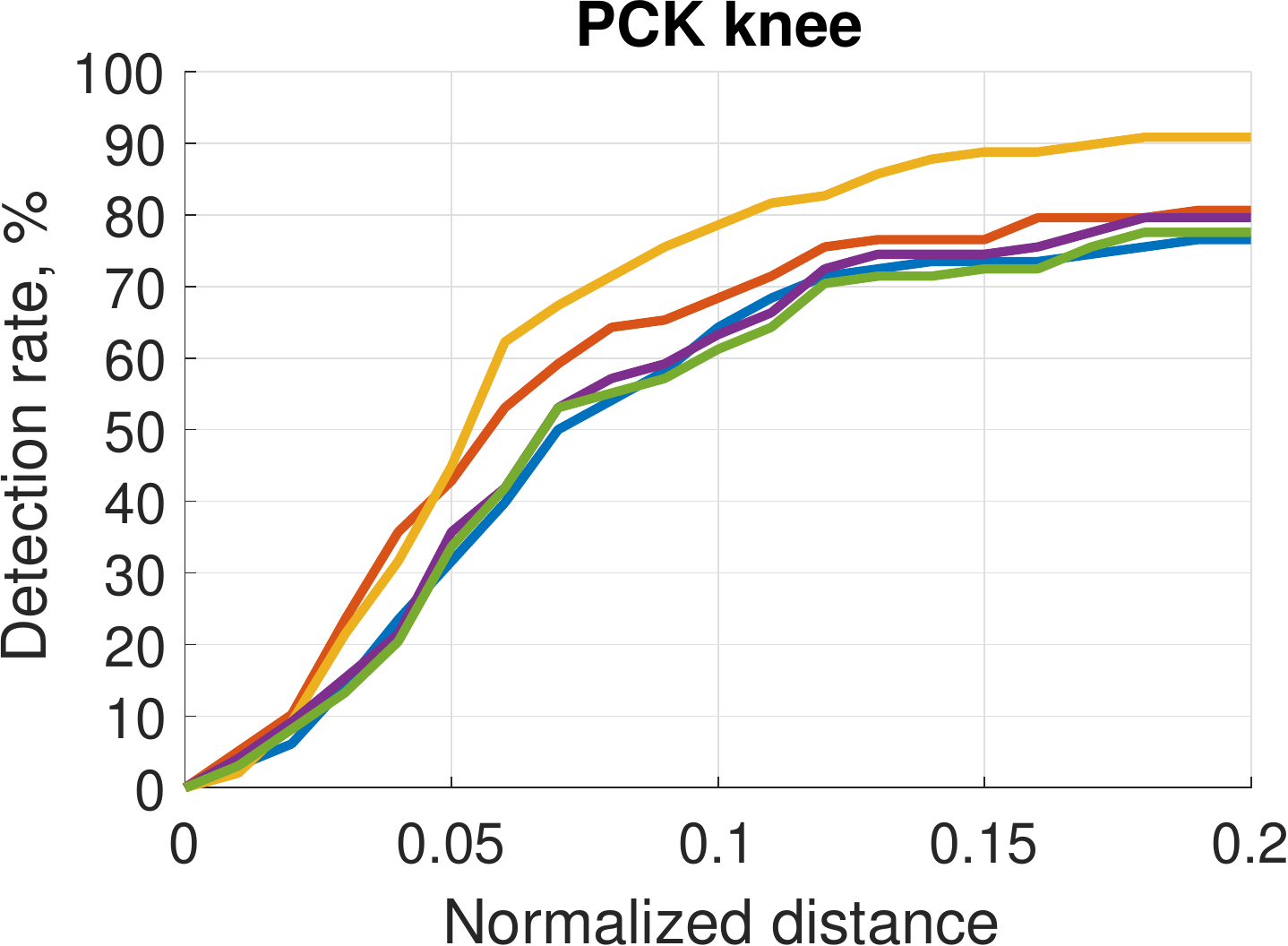}}
  \subfloat[]{\label{fig:pckAnkleMdl}\includegraphics[width=0.24\linewidth,{trim=0in 0in 0in 0in,
  clip=true}]{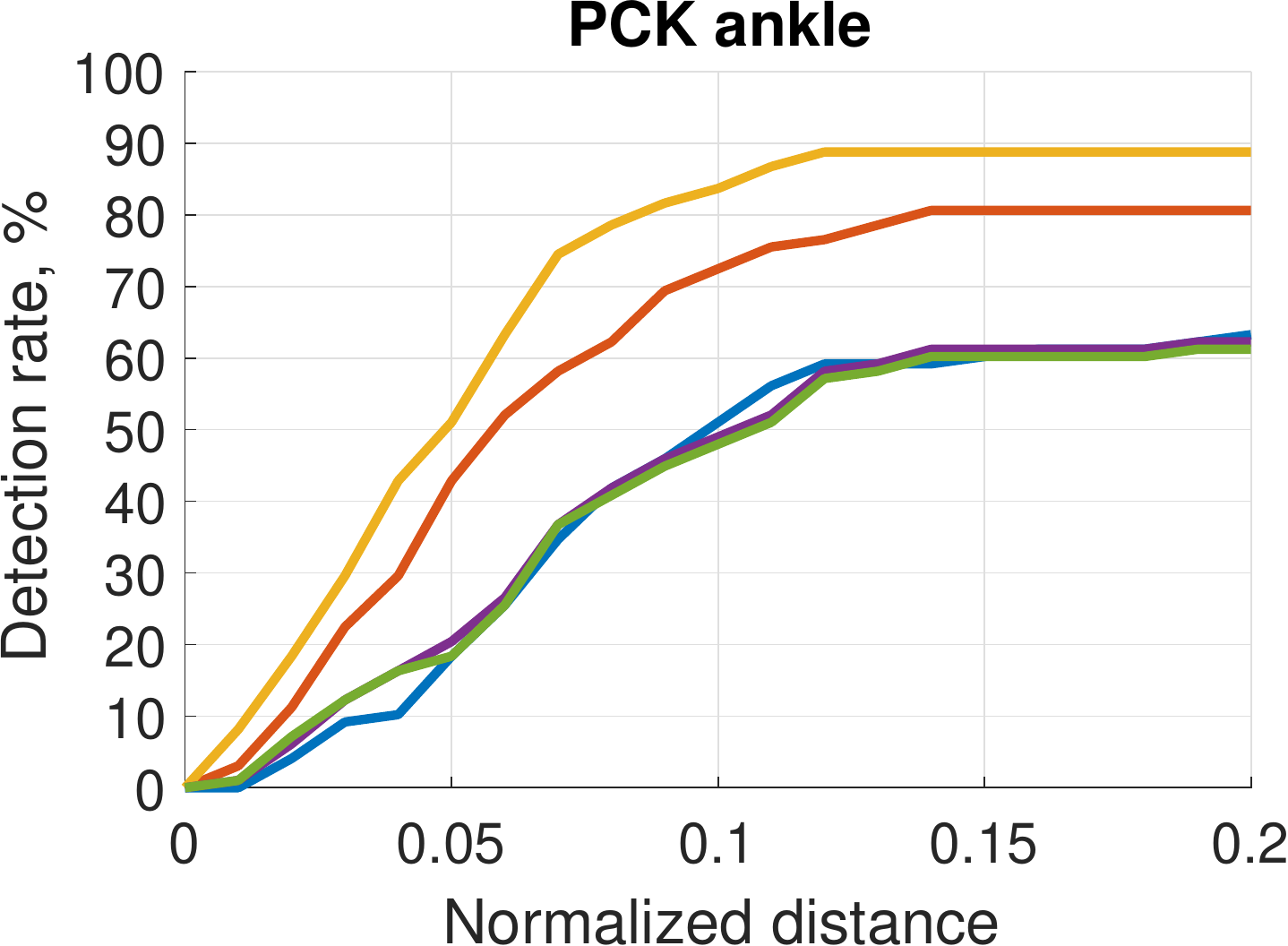}} 
  \\
  \subfloat[]{\label{fig:pckHeadMdl}\includegraphics[width=0.24\linewidth,{trim=0in 0in 0in 0in,
  clip=true}]{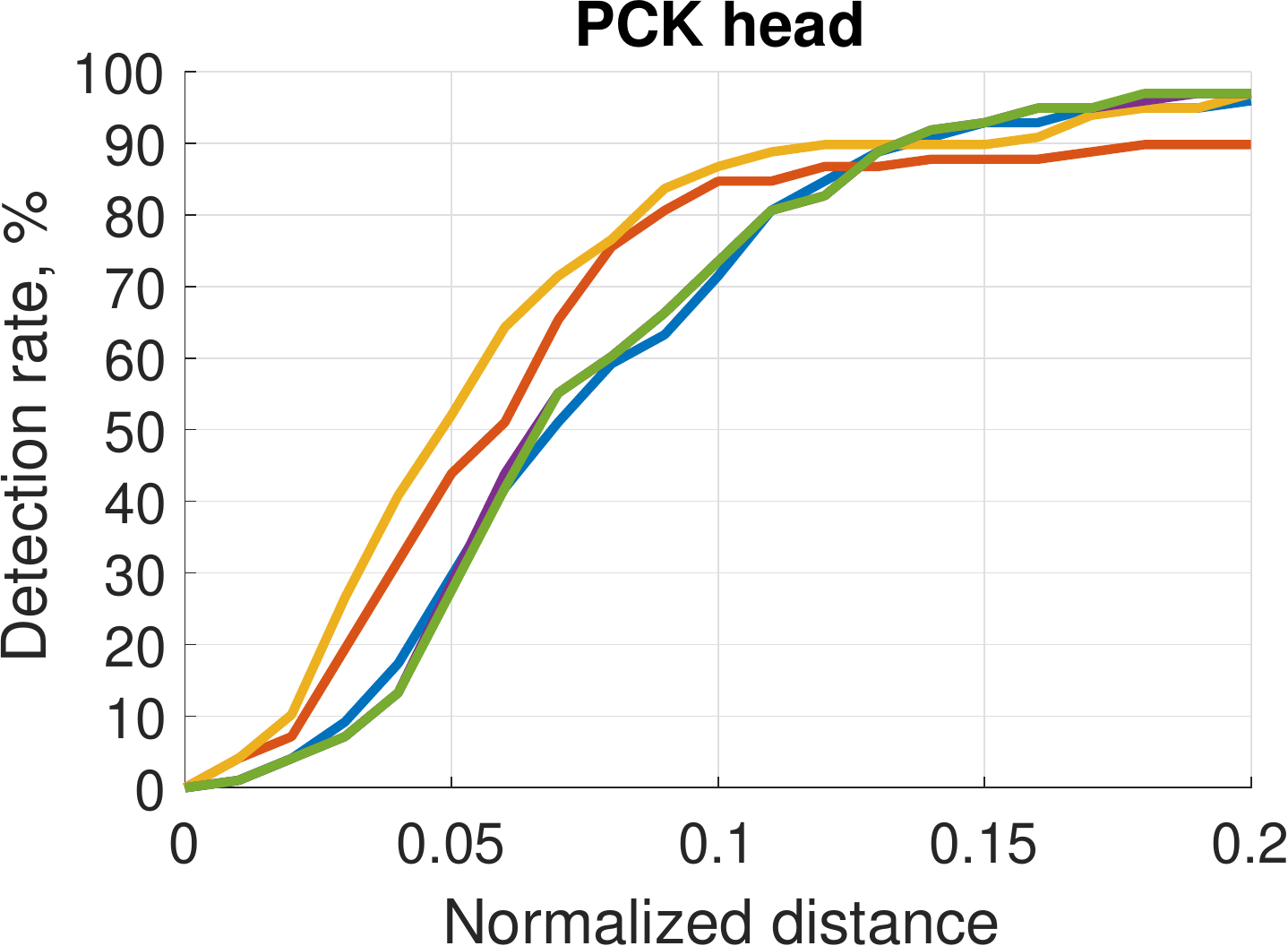}}
  \subfloat[]{\label{fig:pckShoulderMdl}\includegraphics[width=0.24\linewidth,{trim=0in 0in 0in 0in,
  clip=true}]{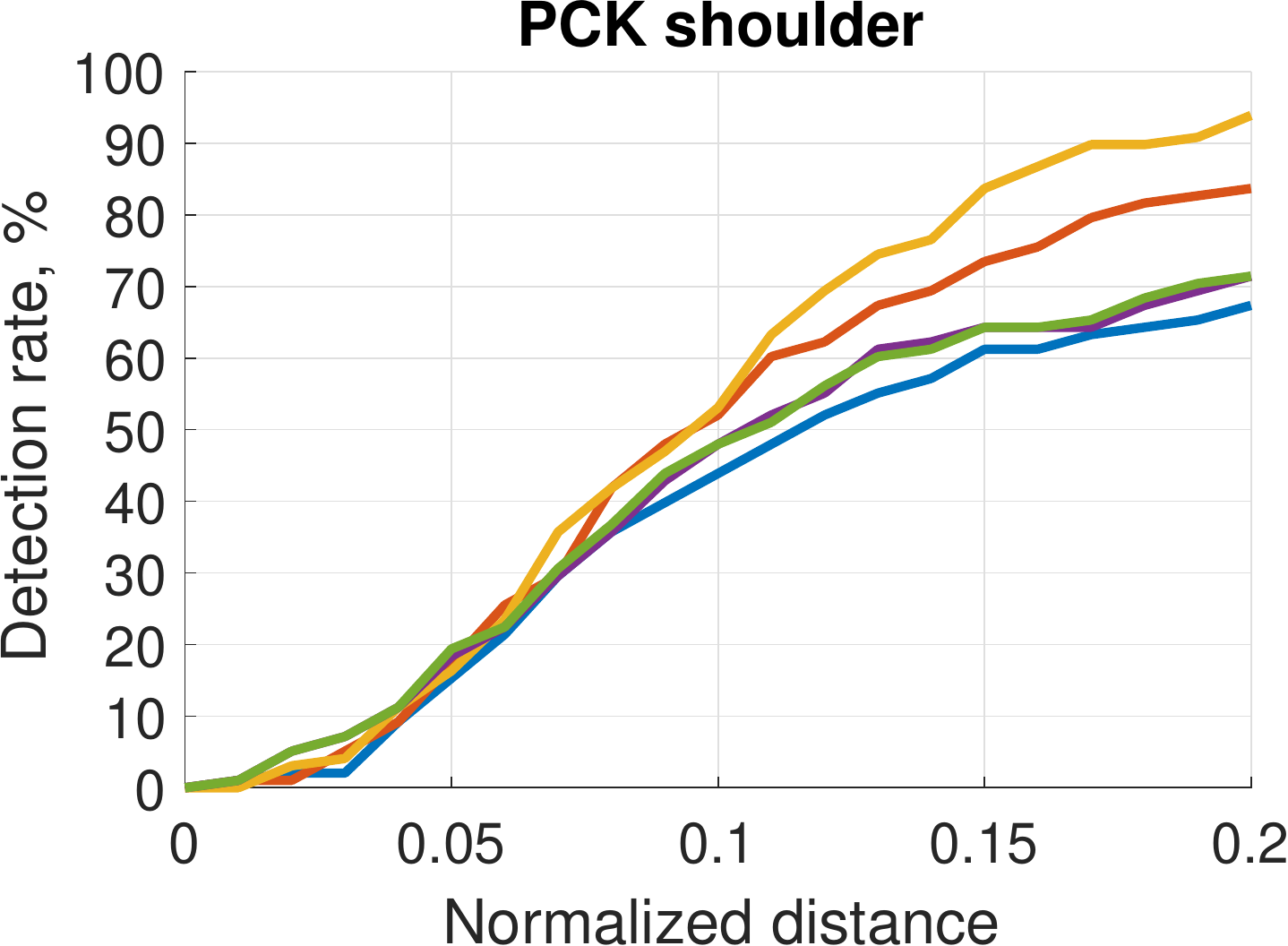}}
  \subfloat[]{\label{fig:pckElbowMdl}\includegraphics[width=0.24\linewidth,{trim=0in 0in 0in 0in,
  clip=true}]{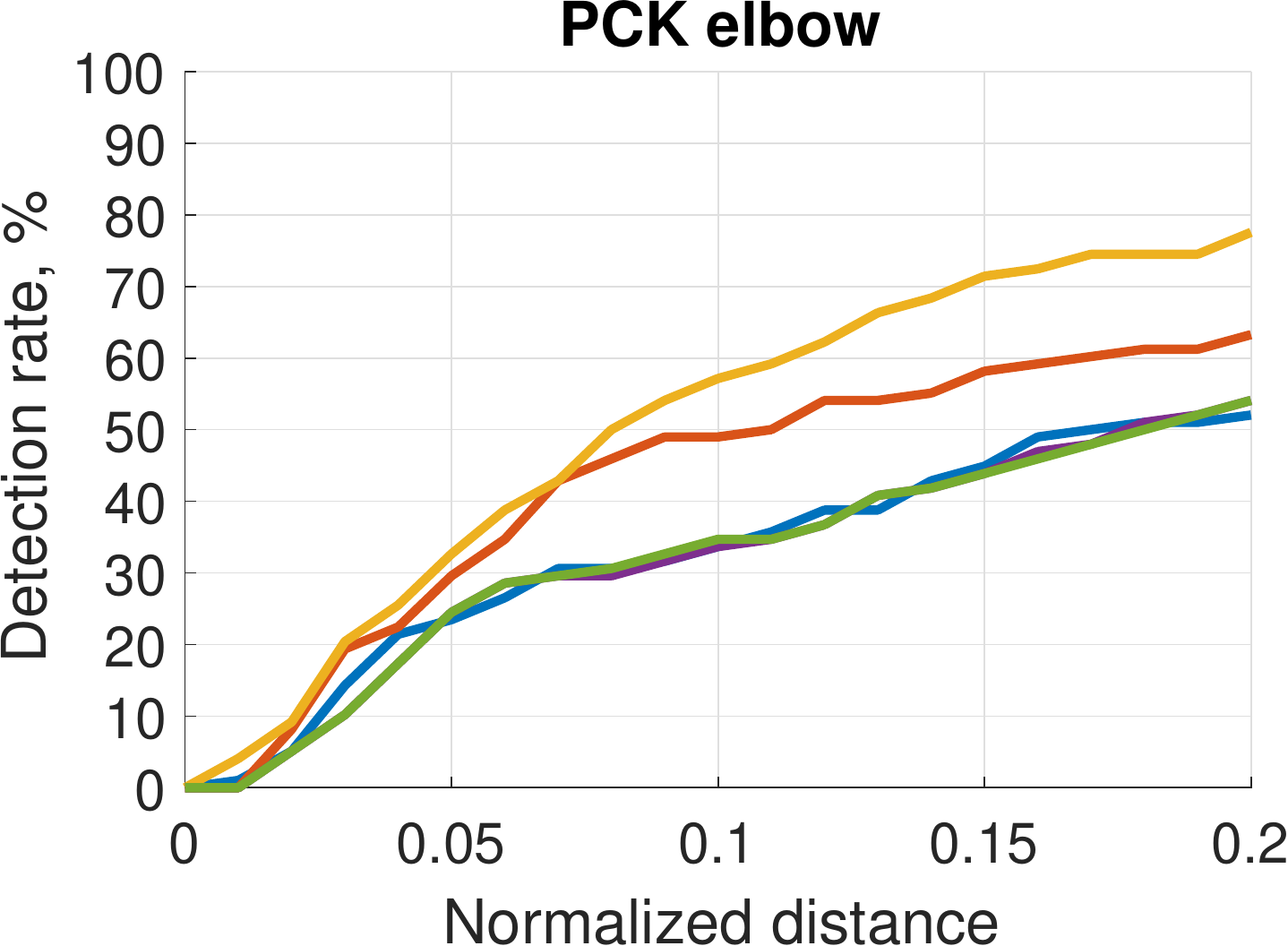}}
  \subfloat[]{\label{fig:pckWristMdl}\includegraphics[width=0.24\linewidth,{trim=0in 0in 0in 0in,
  clip=true}]{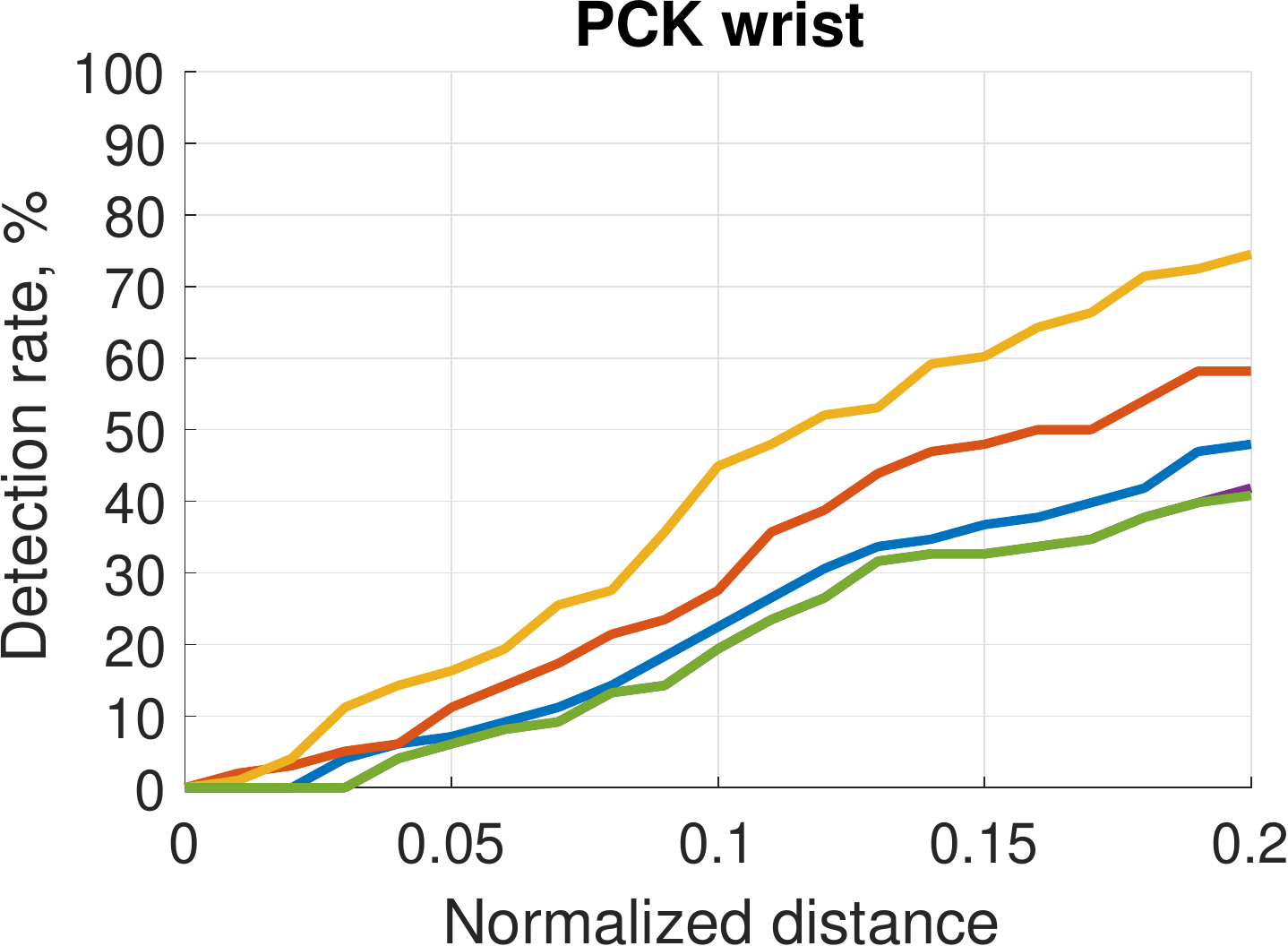}} 
  \caption{Quantitative posture estimation result with different fine-tuning strategies as shown in \figref{tuneMdl}. MPII-LSP stands for the original pre-trained CPM model from MPII and LSP dataset. MANNE-S6 stands for model only fine turned on the last layer of final stage. MANNE-AS stands for the model fine turned with last layers of all stages. MANNE-AS-S2C3-200 and MANNE-AS-S2C3-2000 stand for the model fine turned with last layers of all stages and the 3rd convolutional layer in stage 2 after 200 and 2000 iterations, respectively.}
\label{fig:multMdl}
\vspace{-.2in}
\end{figure*}
}

\newcommand{\figTuneMdl}{
\begin{figure*}[t]
 \centering
 \subfloat[]{\label{fig:tuneMdl1}\includegraphics[width=0.9\linewidth,{trim=0in 0in 0in 0in,
  clip=true}]{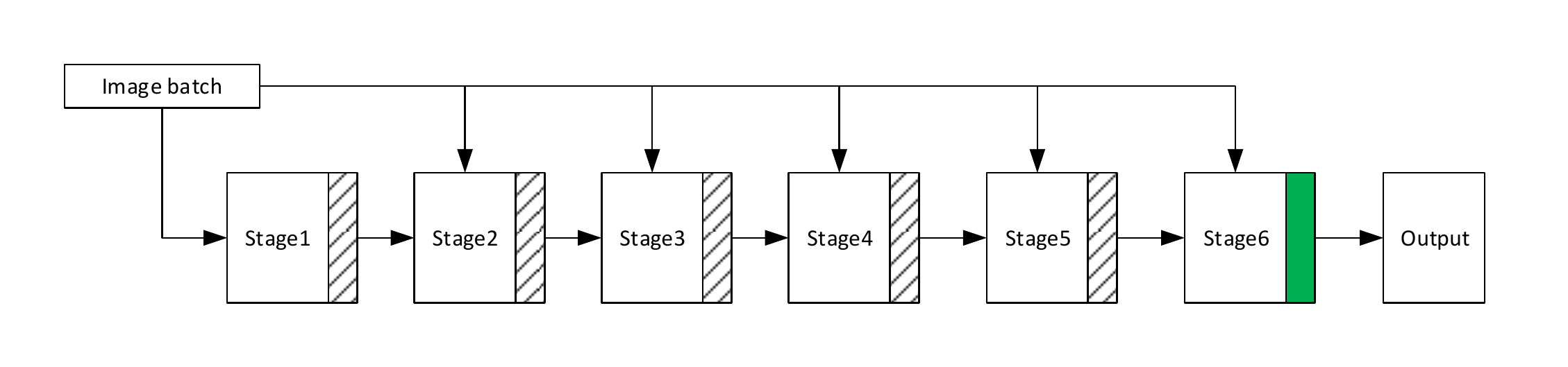}}
  \\
 \subfloat[]{\label{fig:tuneMdl2}\includegraphics[width=0.9\linewidth,{trim=0in 0in 0in 0in,
  clip=true}]{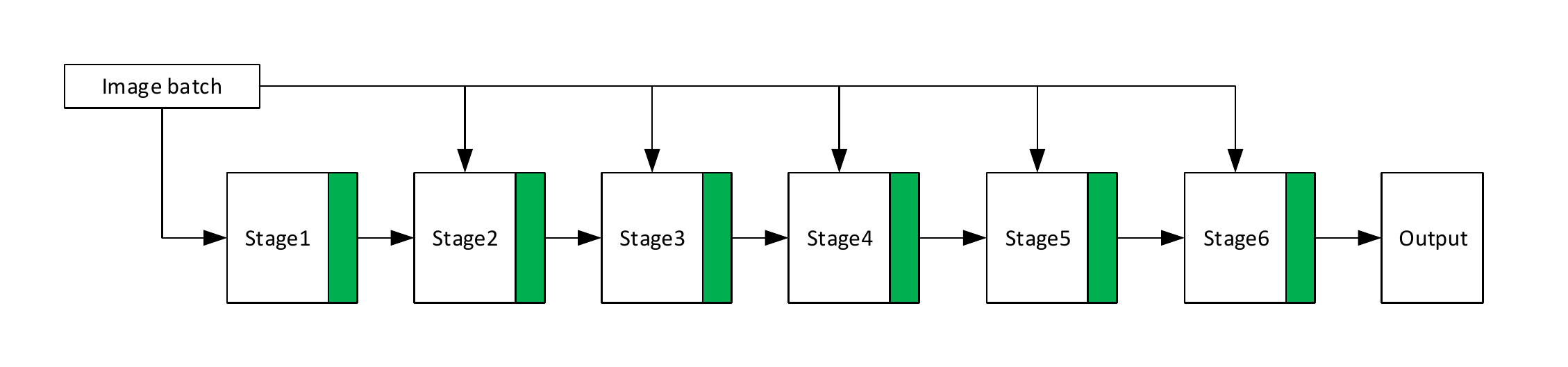}}
  \\
 \subfloat[]{\label{fig:tuneMdl3}\includegraphics[width=0.9\linewidth,{trim=0in 0in 0in 0in,
  clip=true}]{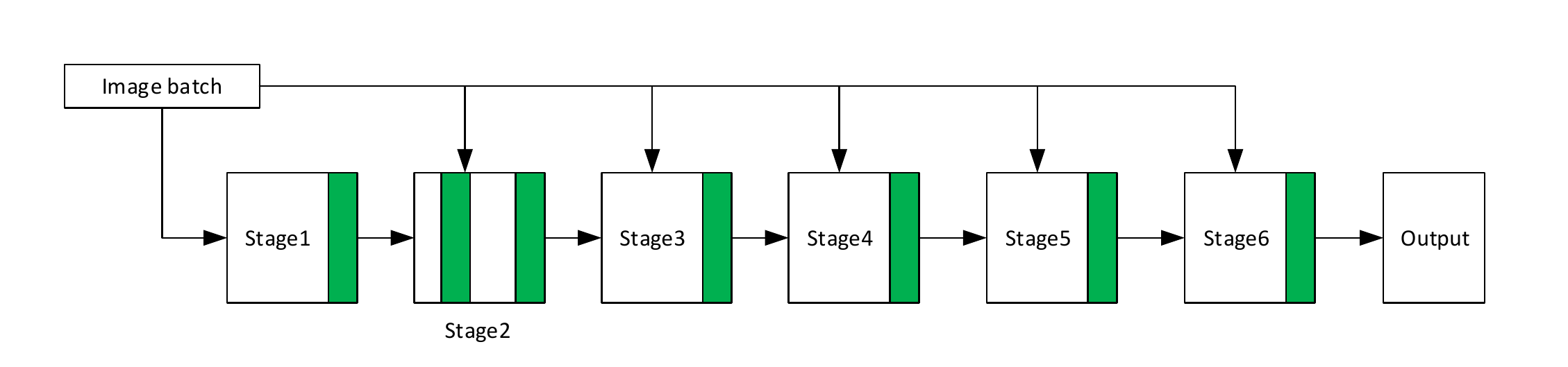}}
   \caption{Fine-tuning configurations: green block indicates the layers for training (a) MANNE-S6: fine-tuning only the last layer before output, (b) MANNE-AS: fine-tuning last layers of all stages, (c) MANNE-AS-S2C3-\#: fine-tuning last layers of all stages as well as the 3rd convolutional layer of stage 2 with \# number of iterations.}
\label{fig:tuneMdl}
\vspace{-.2in}
\end{figure*}
}

\newcommand{\figBlfMap}{
\begin{figure*}[t]
 \centering
 \subfloat[]{\label{fig:000001N1}\includegraphics[width=0.12\linewidth,{trim=0in 0in 0in 0in,
  clip=true}]{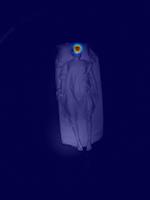}}
   \subfloat[]{\label{fig:000001N2}\includegraphics[width=0.12\linewidth,{trim=0in 0in 0in 0in,
  clip=true}]{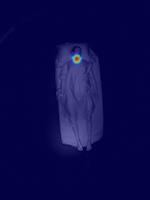}}
   \subfloat[]{\label{fig:000001N3}\includegraphics[width=0.12\linewidth,{trim=0in 0in 0in 0in,
  clip=true}]{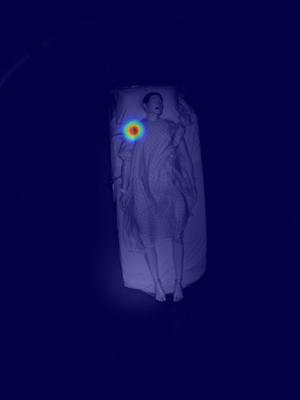}}
   \subfloat[]{\label{fig:000001N4}\includegraphics[width=0.12\linewidth,{trim=0in 0in 0in 0in,
  clip=true}]{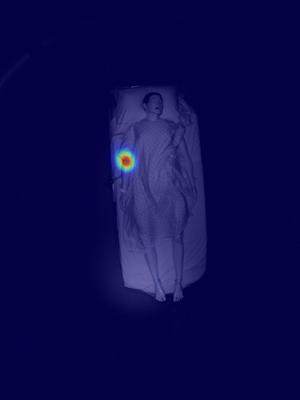}}
   \subfloat[]{\label{fig:000001N5}\includegraphics[width=0.12\linewidth,{trim=0in 0in 0in 0in,
  clip=true}]{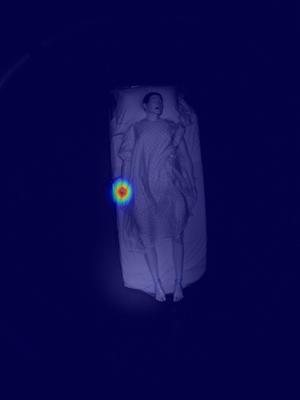}}
   \subfloat[]{\label{fig:000001N6}\includegraphics[width=0.12\linewidth,{trim=0in 0in 0in 0in,
  clip=true}]{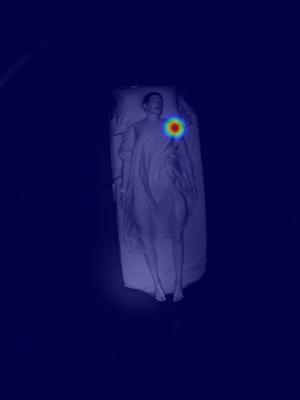}}
   \subfloat[]{\label{fig:000001N7}\includegraphics[width=0.12\linewidth,{trim=0in 0in 0in 0in,
  clip=true}]{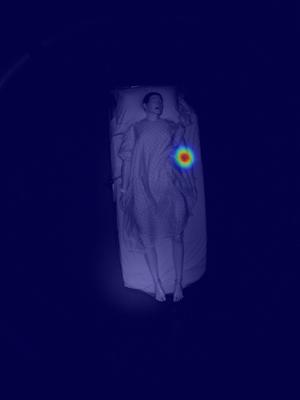}}
   \subfloat[]{\label{fig:000001N8}\includegraphics[width=0.12\linewidth,{trim=0in 0in 0in 0in,
  clip=true}]{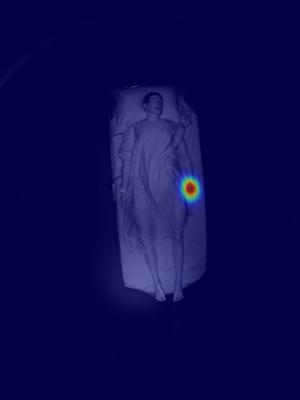}}
  \\
  \subfloat[]{\label{fig:000001N9}\includegraphics[width=0.12\linewidth,{trim=0in 0in 0in 0in,
  clip=true}]{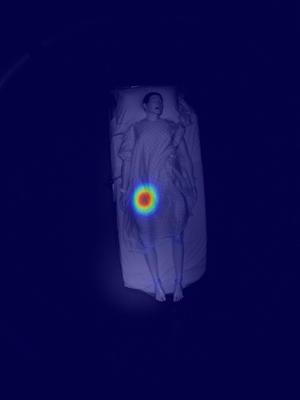}}
   \subfloat[]{\label{fig:000001N10}\includegraphics[width=0.12\linewidth,{trim=0in 0in 0in 0in,
  clip=true}]{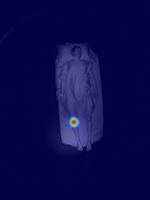}}
   \subfloat[]{\label{fig:000001N11}\includegraphics[width=0.12\linewidth,{trim=0in 0in 0in 0in,
  clip=true}]{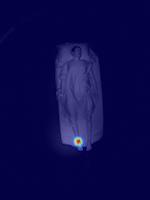}}
   \subfloat[]{\label{fig:000001N12}\includegraphics[width=0.12\linewidth,{trim=0in 0in 0in 0in,
  clip=true}]{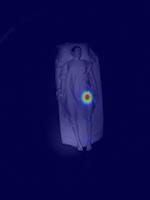}}
   \subfloat[]{\label{fig:000001N13}\includegraphics[width=0.12\linewidth,{trim=0in 0in 0in 0in,
  clip=true}]{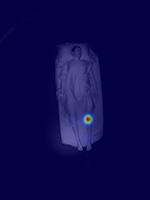}}
   \subfloat[]{\label{fig:000001N14}\includegraphics[width=0.12\linewidth,{trim=0in 0in 0in 0in,
  clip=true}]{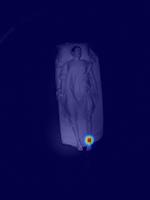}}
   \subfloat[]{\label{fig:000001N15}\includegraphics[width=0.12\linewidth,{trim=0in 0in 0in 0in,
  clip=true}]{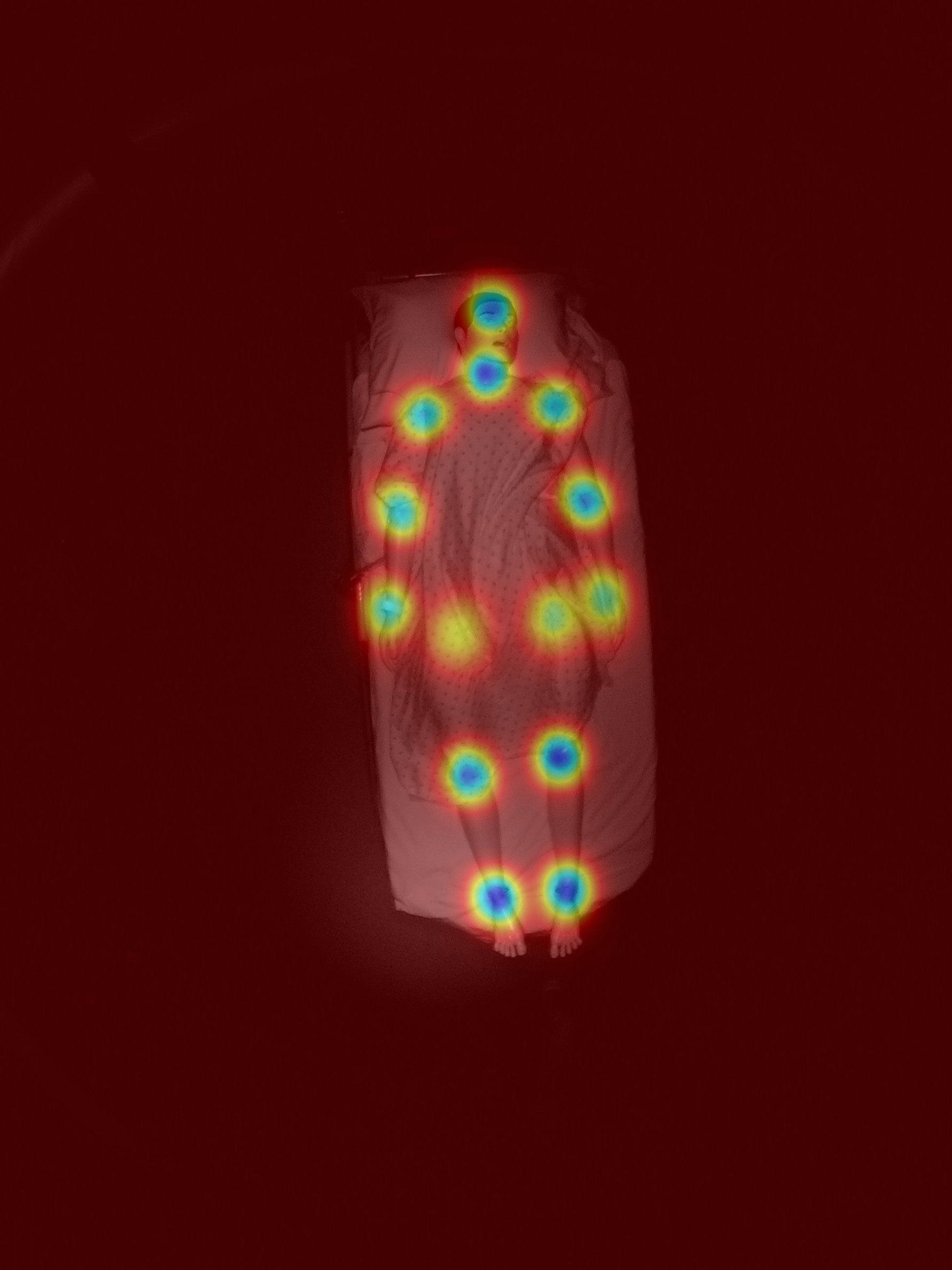}}
   \subfloat[]{\label{fig:000001Nlimb}\includegraphics[width=0.12\linewidth,{trim=0in 0in 0in 0in,
  clip=true}]{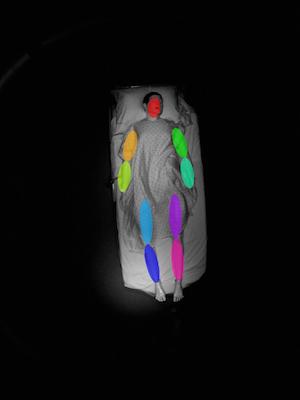}}
  \caption{Human posture estimation result with our MANNE-AS-S2C3-2000 model. (a)-(n) estimated belief map of head, neck, right shoulder, right elbow, right wrist, left shoulder, left elbow, left wrist, right hip, right knee, right ankle, left hip, left knee, and left ankle, (o) background belief map, and (p) pose visualization.}
\label{fig:blfMap}
\vspace{-.2in}
\end{figure*}
}

\newcommand{\figMultPre}{
\begin{figure*}[t]
 \centering
 \subfloat[]{\label{fig:pckTotalPre}\includegraphics[width=0.24\linewidth,{trim=0in 0in 0in 0in,
  clip=true}]{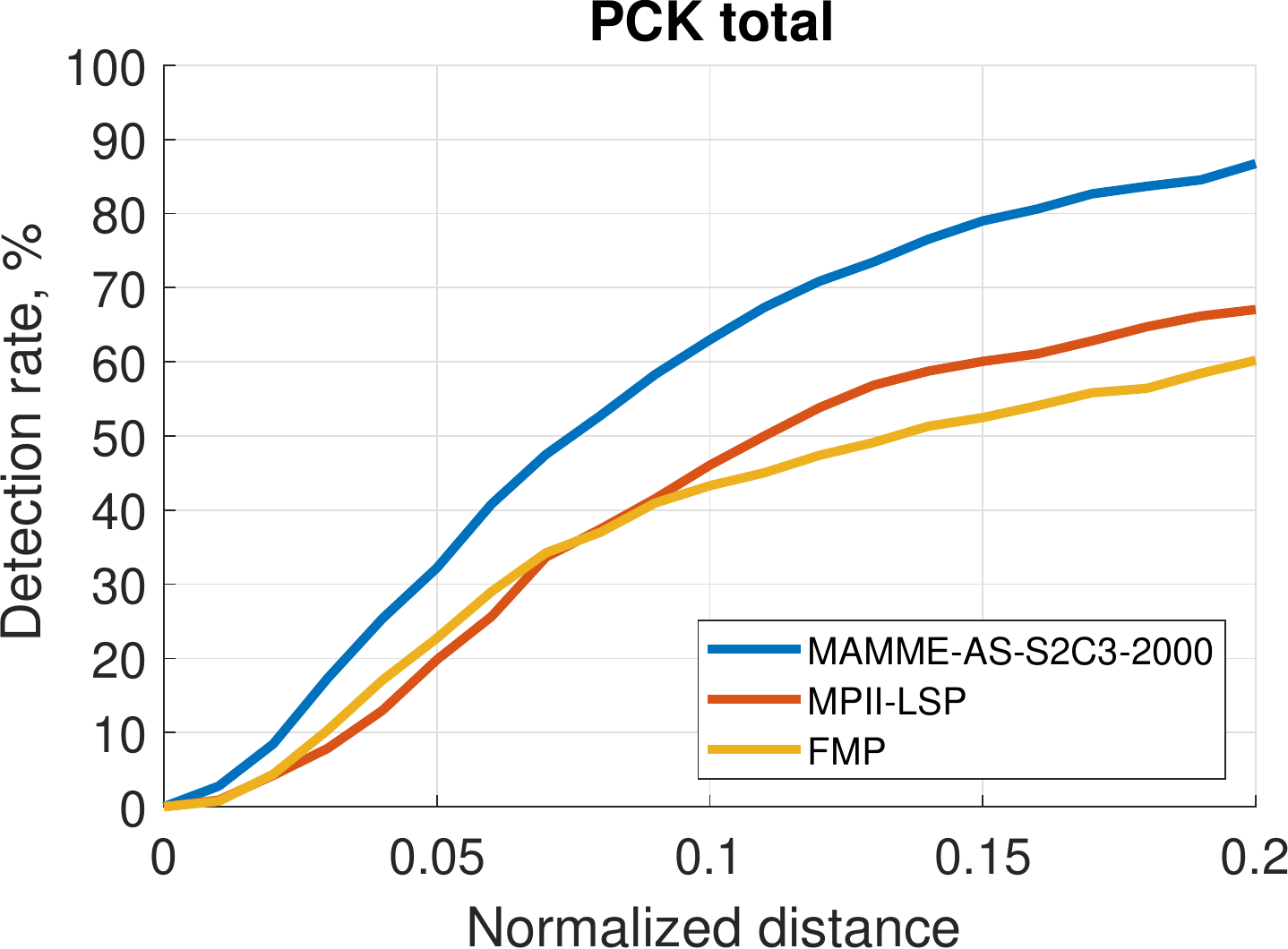}}
   \subfloat[]{\label{fig:pckHipPre}\includegraphics[width=0.24\linewidth,{trim=0in 0in 0in 0in,
  clip=true}]{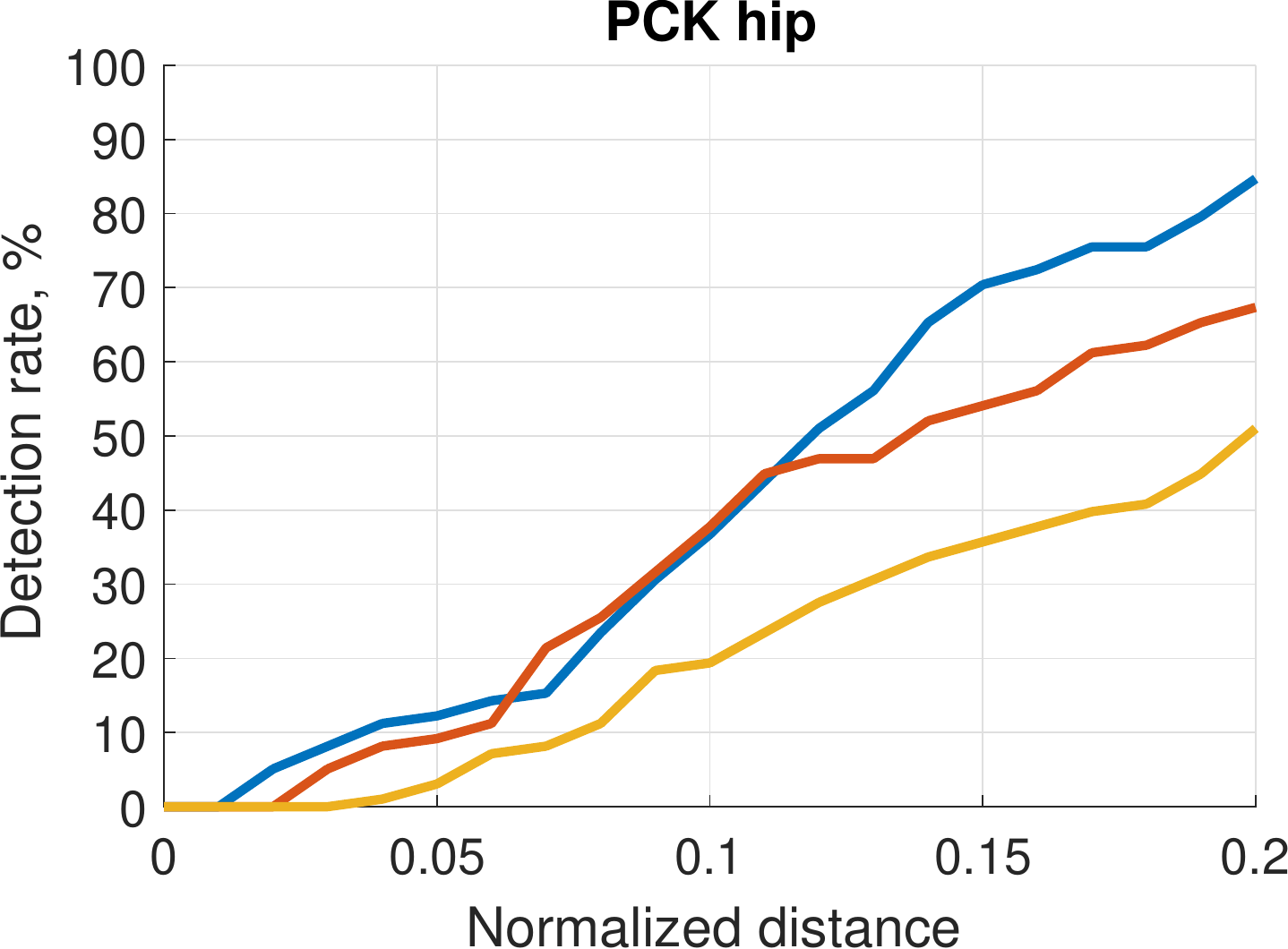}}
   \subfloat[]{\label{fig:pckKneePre}\includegraphics[width=0.24\linewidth,{trim=0in 0in 0in 0in,
  clip=true}]{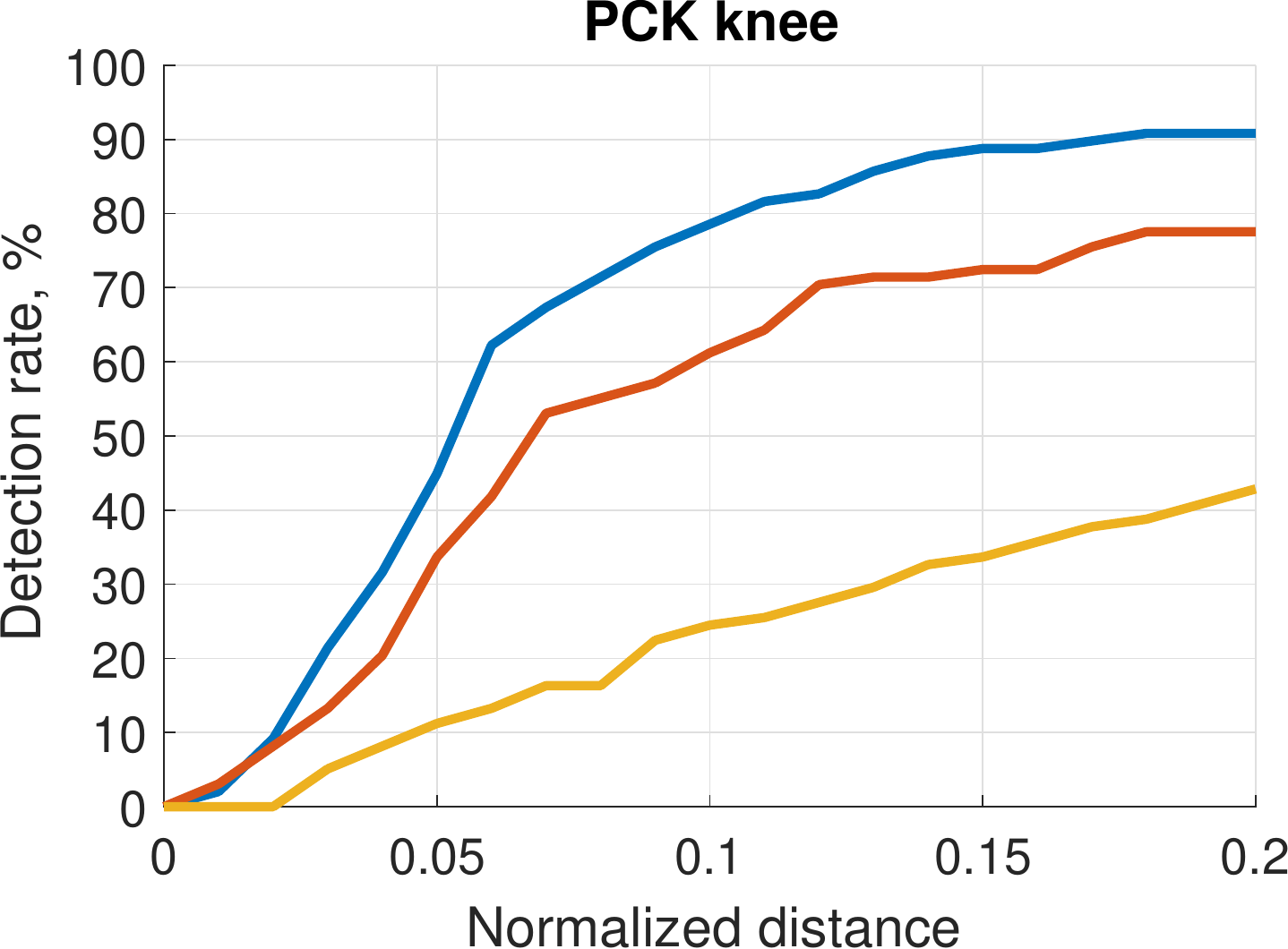}}
   \subfloat[]{\label{fig:pckAnklePre}\includegraphics[width=0.24\linewidth,{trim=0in 0in 0in 0in,
  clip=true}]{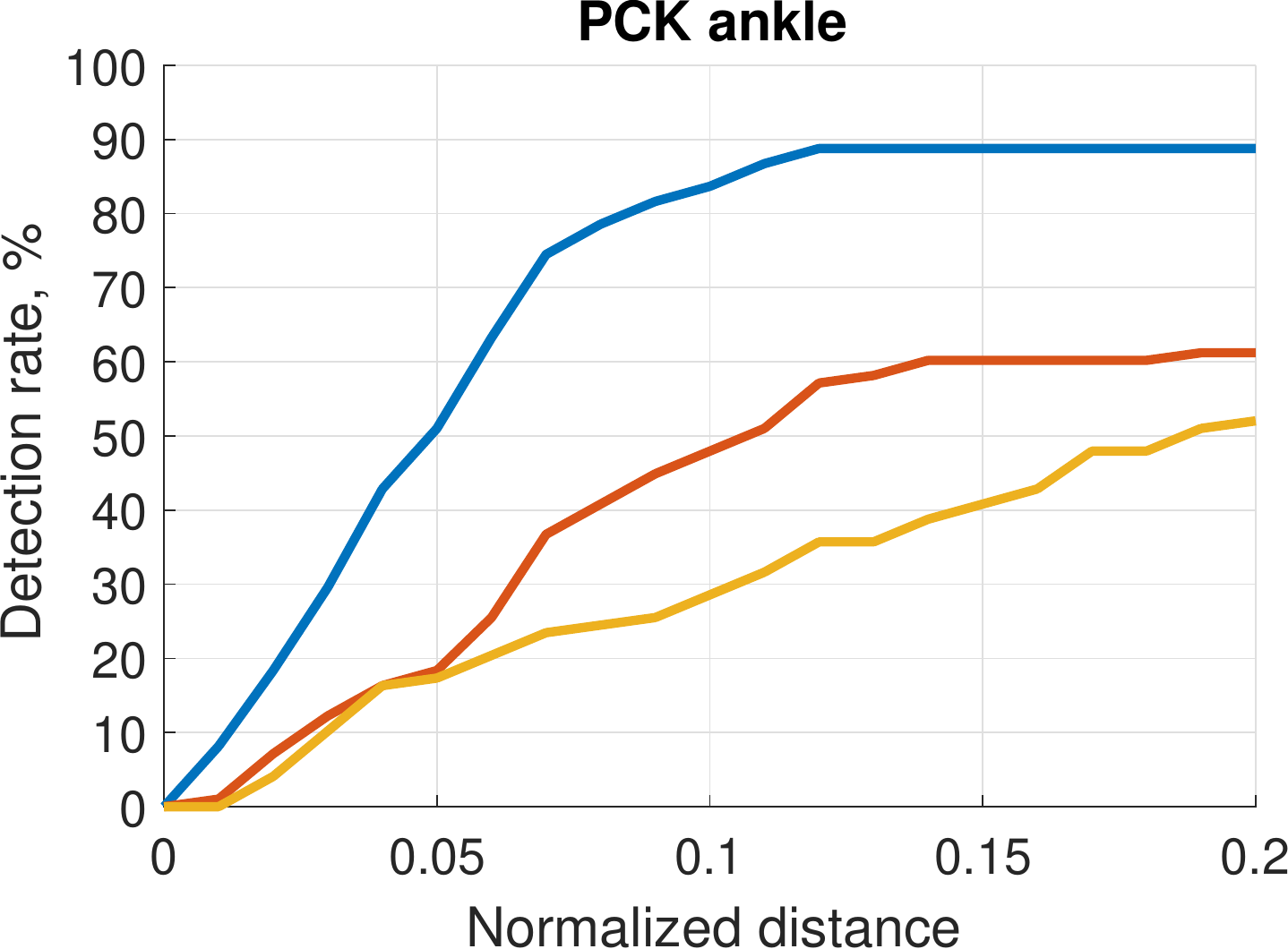}} 
  \\
   \subfloat[]{\label{fig:pckHeadPre}\includegraphics[width=0.24\linewidth,{trim=0in 0in 0in 0in,
  clip=true}]{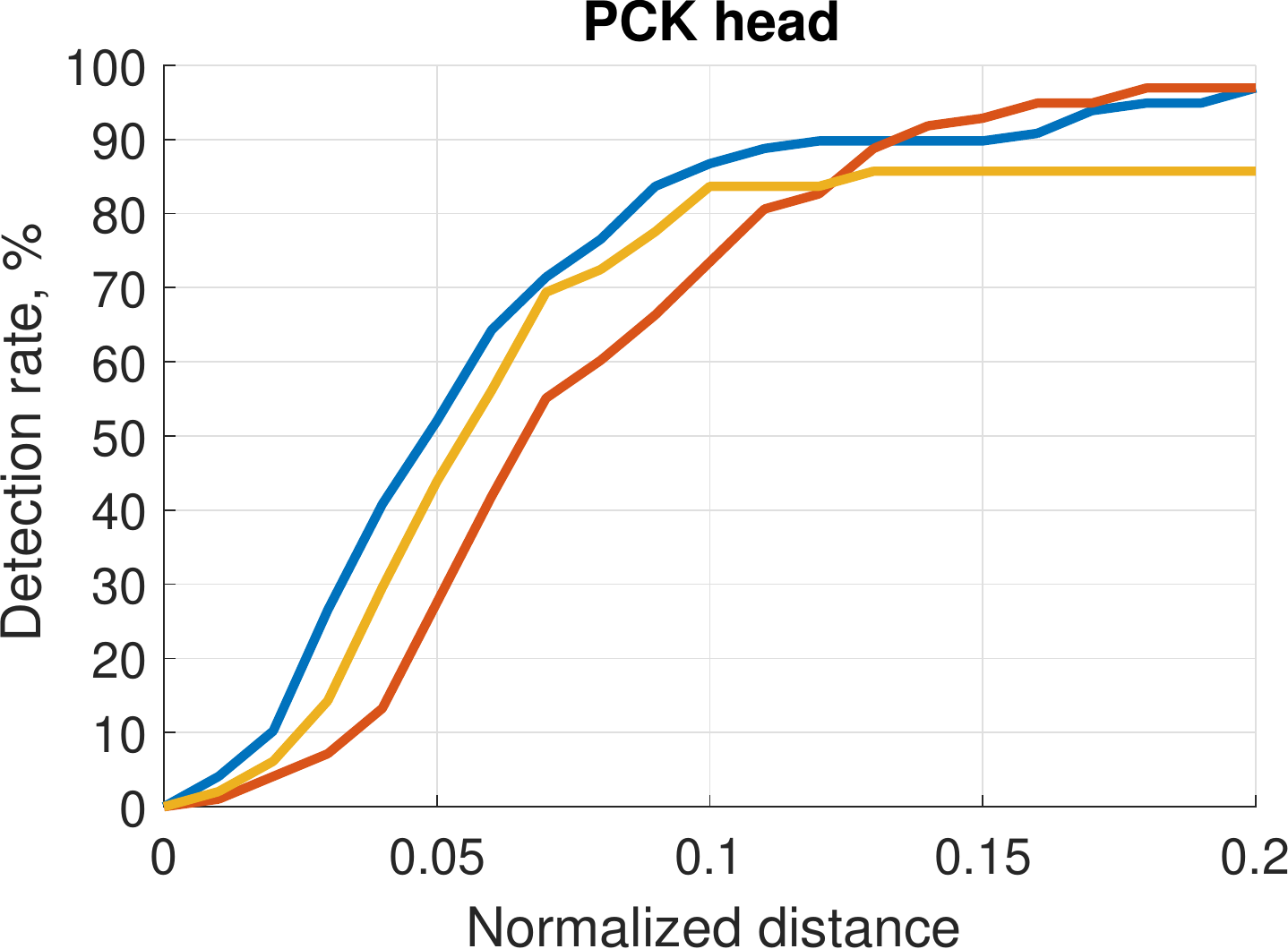}}
   \subfloat[]{\label{fig:pckShoulderPre}\includegraphics[width=0.24\linewidth,{trim=0in 0in 0in 0in,
  clip=true}]{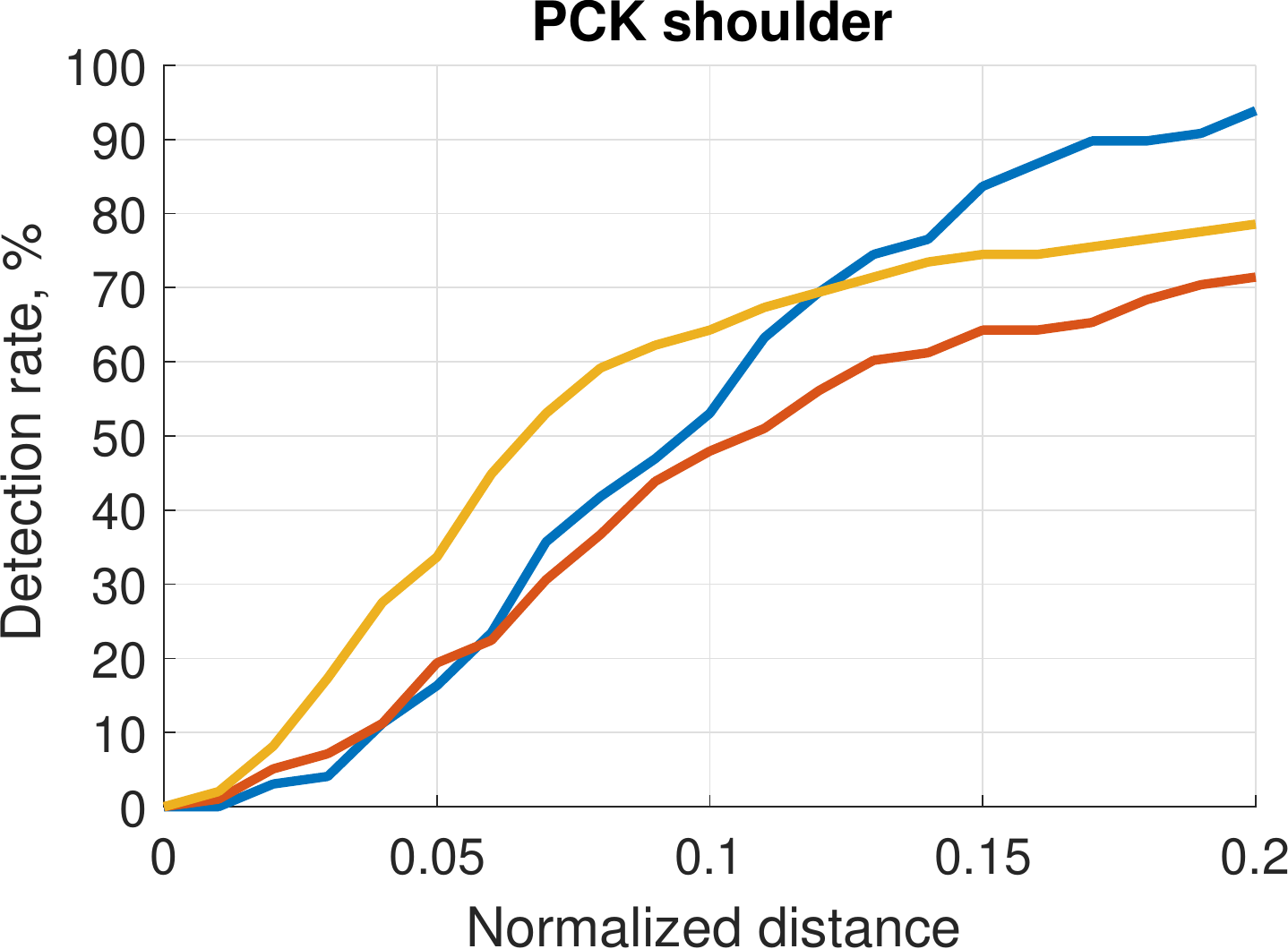}}
   \subfloat[]{\label{fig:pckElbowPre}\includegraphics[width=0.24\linewidth,{trim=0in 0in 0in 0in,
  clip=true}]{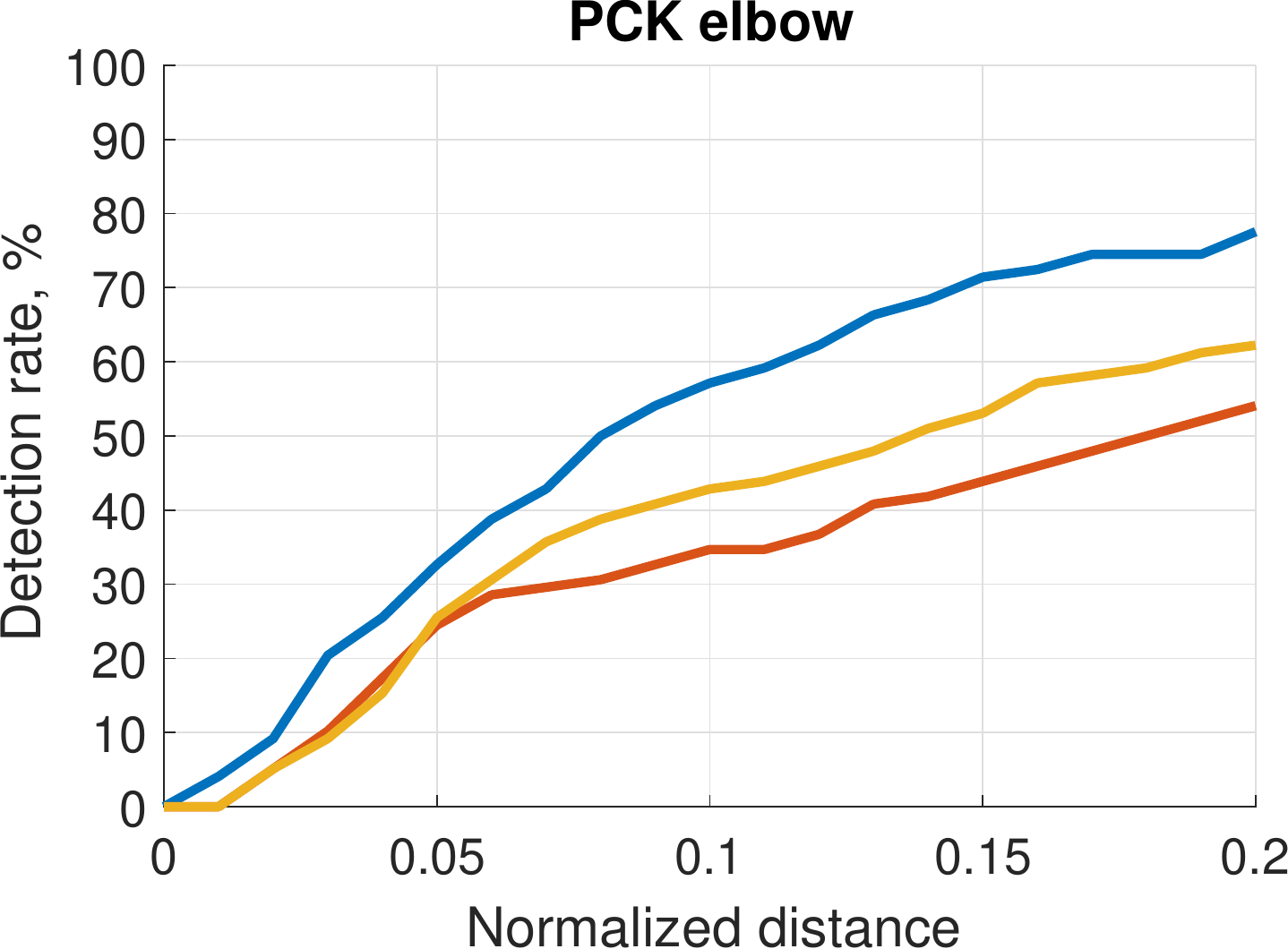}}
   \subfloat[]{\label{fig:pckWristPre}\includegraphics[width=0.24\linewidth,{trim=0in 0in 0in 0in,
  clip=true}]{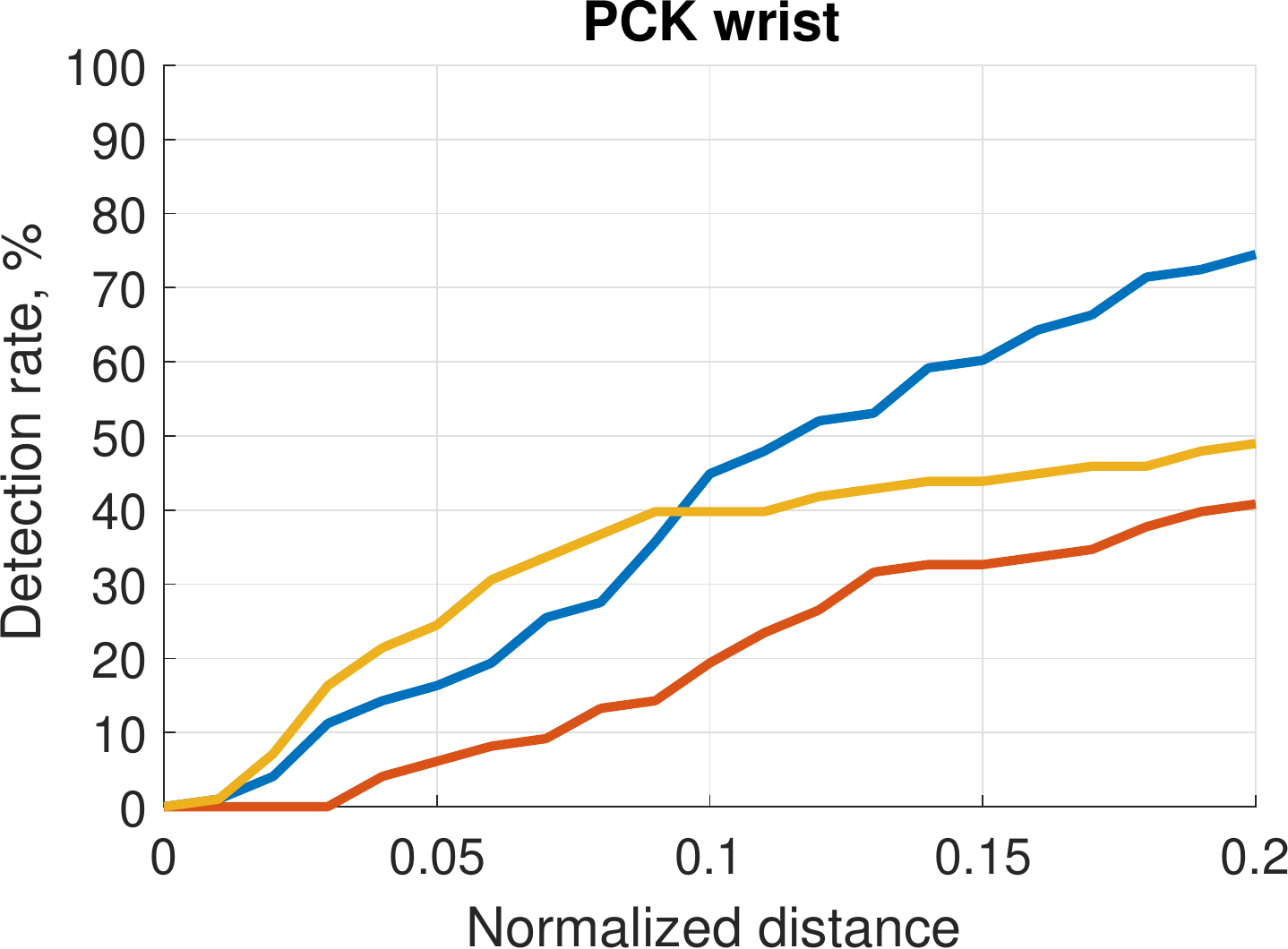}} 
   \caption{Quantitative posture estimation result with different pre-trained general purpose pose detection models against our fine-tuned model on the rectified IRS in-bed pose images. }
\label{fig:multPre}
\vspace{-.2in}
\end{figure*}
}

\newcommand{\figOnDemTrigger}{
\begin{figure}[h]
 \centering
 \includegraphics[width=0.84\linewidth,{trim=0in 0in 0in 0in,
  clip=true}]{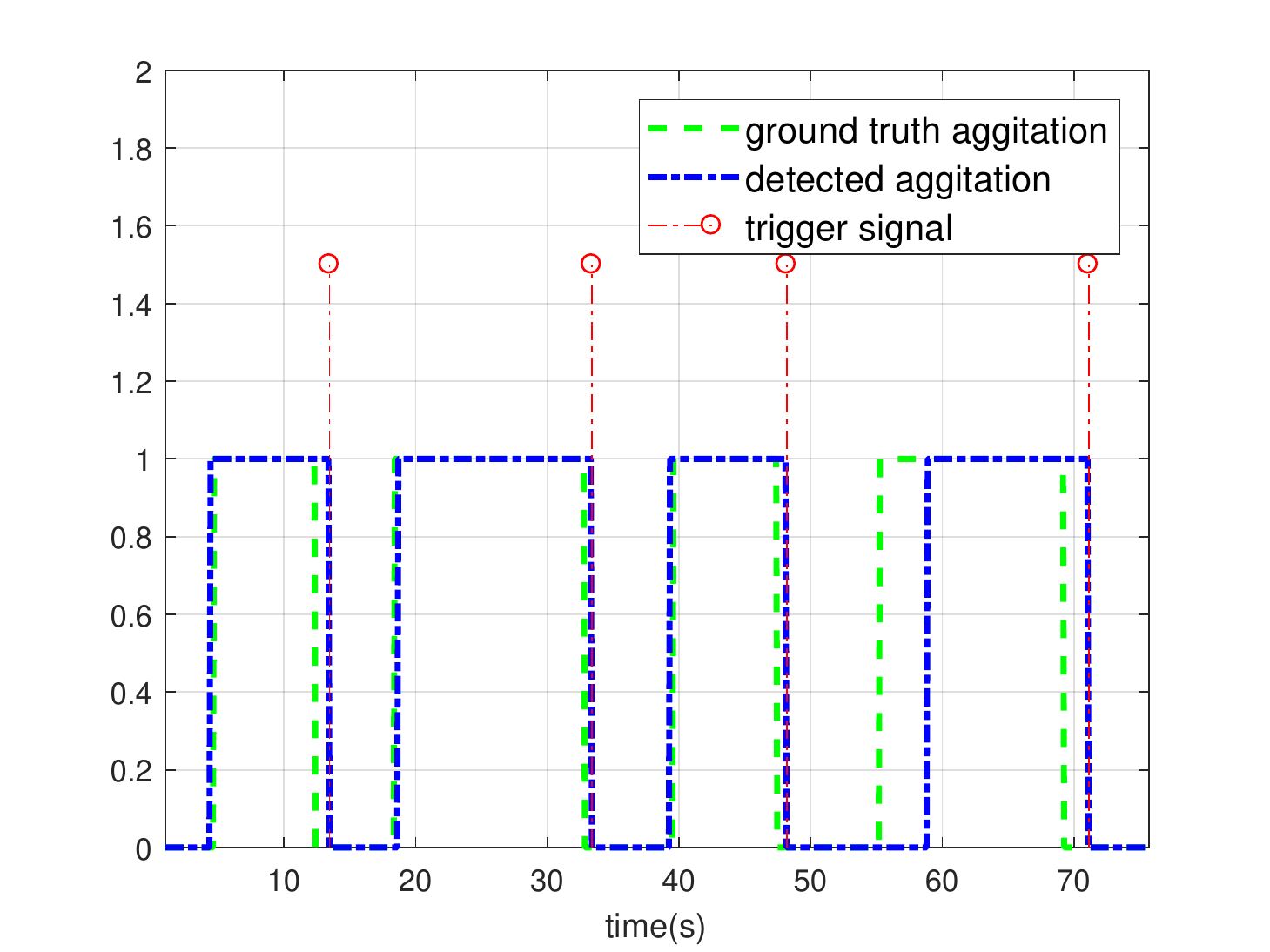}
 \caption{On-demand state estimation trigger result. Our on-demand trigger detection pipeline is applied on a short relocation video of a mannequin on the bed and the estimation result is plotted against the manually-achieved ground truth states.}
\label{fig:onDemTrigger}
\vspace{-.2in}
\end{figure}
}

\newcommand{\figAnnoTool}{
\begin{figure}[b]
 \centering
 \includegraphics[width=0.84\linewidth,{trim=0in 0in 0in 0in,
  clip=true}]{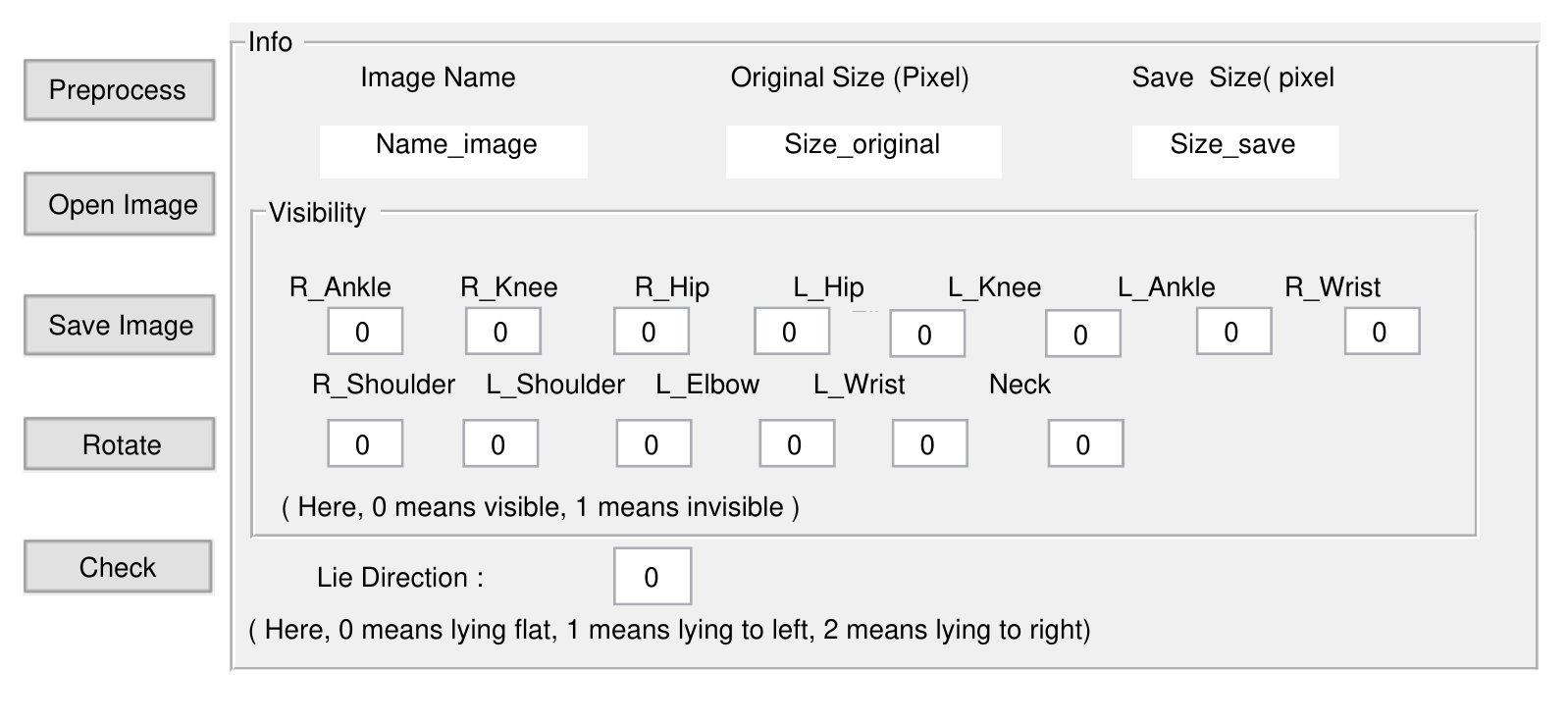}
 \caption{Semi-automatic tool for human pose annotation. Joint position is annotated in a pre-set order, sequentially. Visibility is annotated as 1 or 0 for visible or not. 'Save Image' function is provided to save annotated image for later review, where target size can be assigned by the system user.}
\label{fig:annoTool}
\end{figure}
}

\newcommand{\tabOriTable}{
\begin{table}[h]
\caption{In-bed orientation encoding}
\begin{center}
 \begin{tabular}{c | c  c  c c } 
  categories &  $E$ & $W$ & $N$ & $S$ \\ 
 \hline
 $bit_H$ & 1 & 1& 0 & 0 \\ 
 \hline
 $bit_N$ & 1 & 0 & 1 & 0 \\
\end{tabular}
\label{tbl:oriTable}
\end{center}
\end{table}
}

\newcommand{\tabMdlPCK}{
\begin{table*}
\caption{Pose estimation accuracy in PCK0.2 standard using FPM, pre-trained CPM, and our fine-tuned CPM model}
\begin{center}
 \begin{tabular}{c | c  c  c c c c c c} 
  Models & Total & Ankle & Knee & Hip & wrist & Elbow & Shoulder & Head \\
  \hline
  FPM            & 60.2 & 52.0 & 42.9 & 51.0 & 49.0 & 62.2 & 78.6 & 85.7\\ 
 \hline
 MPII-LSP CPM   & 67.1 & 61.2 & 77.6 & 67.3 & 40.8 & 54.1 & 71.4 & \textbf{96.9} \\
 \hline
 
 MANNE-AS-S2C3-2000   & \textbf{86.7} & \textbf{88.8} & \textbf{90.8} & \textbf{84.7} & \textbf{74.5} & \textbf{77.6} & \textbf{93.9} & \textbf{96.9}\\
\end{tabular}
\label{tbl:mdlPCK}
\end{center}
\end{table*}
}

\begin{document}

%%%%%%%%% TITLE
\title{In-Bed Pose Estimation: \\Deep Learning with Shallow Dataset}

% another suggestion: In-Bed Pose Estimation: Repurposing Pre-trained Convolutional Neural Networks with Limited Data

\author{Shuangjun~Liu, Yu~Yin,
        and~Sarah~Ostadabbas% <-this % stops a space
\thanks{S. Liu, Y. Yin and S. Ostadabbas are with the Augmented Cognition Lab (ACLab) at the Department
of Electrical and Computer Engineering, Northeastern University, Boston,
MA 02115, USA, e-mails: liu.shu@husky.neu.edu, yin.yu1@husky.neu.edu, ostadabbas@ece.neu.edu.}% <-this % stops a space
}

\markboth{JOURNAL OF ,~Vol.~, No.~, Month~Year}%
{Shell \MakeLowercase{\textit{et al.}}: Bare Demo of IEEEtran.cls for IEEE Journals}

\maketitle
%\thispagestyle{empty}

%%%%%%%%% ABSTRACT
\begin{abstract}
This paper presents a robust human posture and body parts detection method under a specific application scenario, in-bed pose estimation. Although human pose estimation for various computer vision (CV) applications has been studied extensively in the last few decades, yet in-bed pose estimation using camera-based vision methods has been ignored by the CV community because it is assumed to be identical to the general purpose pose estimation methods. However, in-bed pose estimation has its own specialized aspects and comes with specific challenges including the notable differences in lighting conditions throughout a day and also having different pose distribution from the common human surveillance viewpoint. In this paper, we demonstrate that these challenges significantly lessen the effectiveness of existing general purpose pose estimation models. In order to address the lighting variation challenge, infrared selective (IRS) image acquisition technique is proposed to provide uniform quality data under various lighting conditions. In addition, to deal with unconventional pose perspective, a 2- end histogram of oriented gradient (HOG) rectification method is presented. Deep learning framework proves to be the most effective model in human pose estimation, however the lack of large public dataset for in-bed poses prevents us from using a large network from scratch. In this work, we explored the idea of employing a pre-trained convolutional neural network (CNN) model trained on large public datasets of general human poses and fine-tuning the model using our own shallow (limited in size and different in perspective and color) in-bed IRS dataset. We developed an IRS imaging system and collected IRS image data from several realistic life-size mannequins in a simulated hospital room environment. A pre-trained CNN called convolutional pose machine (CPM) was repurposed for in-bed pose estimation by fine-tuning its specific intermediate layers. Using the HOG rectification method, the pose estimation performance of CPM significantly improved by 26.4\% in PCK0.1 (probability of correct keypoint) criteria compared to the model without such rectification. Even testing with only well aligned in-bed pose images, our fine-tuned model still surpassed the traditionally-tuned CNN by another 16.6\% increase in pose estimation accuracy. 
\end{abstract}


\IEEEpeerreviewmaketitle

%%%%%%%%% BODY TEXT
\section{Introduction}
%\com{The paper needs a careful revise. The figures captions needs to be revised as well. This paper is now under revision for the Elsevier journal on Image and Vision Computing.}

Human  in-bed  pose and posture  are  important  health-related metrics  with  potential values in many medical applications such as sleep monitoring. It is shown that sleeping pose affects the symptoms of many diseases such as sleep apnea \cite{lee2015changes}, pressure ulcers \cite{black2007national}, and even carpal tunnel syndrome \cite{mccabe2011preferred,mccabe2010evaluation}. Moreover, patients are usually required to maintain specific poses after certain surgeries to get a better recovery result. Therefor, long-term monitoring and automatically detecting in-bed poses are of critical interest in healthcare \cite{abouzari2007role}. 

\figSystem 
Currently, besides self-reporting by patients and visual inspection by the caregivers, in-bed pose estimation methods mainly rely on the use of pressure mapping systems. Authors in \cite{pouyan2013continuous} extracted binary signatures from pressure images obtained from a commercial pressure mat and used a binary pattern matching technique for pose classification. The same group also introduced a Gaussian mixture model (GMM)-based clustering approach for concurrent pose classification and limb identification using pressure data \cite{Ostadabbas2014}. Pictorial structure model of the body based on both appearance and spatial information was employed to localize the body parts within pressure images in \cite{liu2014bodypart}. The authors considered each part of the human body as a vertex in a tree and found how well the appearance of each body part matches its template as well as how far the body parts deviate from their expected respective locations. Finally, the best configuration of body parts was selected by minimizing the total cost. Although pressure mapping based methods are effective at localizing areas of increased pressure and even automatically classifying overall postures \cite{Ostadabbas2014}, the pressure sensing mats are expensive ($>$\$10K) and require frequent maintenance. These obstacles have prevented pressure-based pose monitoring solutions from achieving large-scale popularity.

By contrast, camera-based vision methods for human pose estimation show great advantages including their low cost and ease of maintenance. General purpose human pose estimation has become an active area in computer vision and surveillance research. The methods and algorithms for pose estimation can be categorized into five categories: (i) The classical articulated pose estimation model, which is a pictorial structures model \cite{andriluka2009pictorial,andriluka2010monocular,felzenszwalb2005pictorial,johnson2010clustered,pishchulin2013poselet,pishchuli2013strong}. It employs a tree-structured graphical model to constrain the kinematic relationship between body parts. However, it requires the person to be visible and is prone to errors such as double counting evidence. Some recent works have augmented this structure by embedding flexible mixture of parts (FMP) into the model \cite{Antol2014,Ferrari08}. (ii) Hierarchical models, which represent the body part in different scale in a hierarchical tree structure, where parts in larger scale can help to localize small body parts \cite{sun2011articulated,tian2012exploring}. (iii) Non-tree models, which augment the tree structure with additional edges to capture the potential long range relationship between body parts \cite{dantone2013human,karlinsky2012using,wang2008multiple}. (iv) Sequential prediction frameworks, which learn the implicit spatial model directly from training process \cite{munoz2010stacked,pinheiro2014recurrent}. (v) Deep neural network based method usually in a convolutional neural network (CNN) configuration \cite{pishchul2016deepcut, tompson2014joint, tompson2015efficient}. A recent CNN-based work, called convolutional pose machine (CPM) employed multi-stage CNN structures to estimate various human poses \cite{wei2016convolutional}. The CPM was tested on several well-recognized public datasets and promising results were obtained in estimating general purpose poses. 

Although our work focuses on in-bed pose estimation, due to the use of camera for imaging instead of pressure mat, this line of research is categorized under camera-based human pose estimation. It is sensible to assume that pre-trained models on existing datasets of various human poses should be able to address in-bed pose estimation as well. However, it turned out that when it comes to pose monitoring and estimation from individual in sleeping postures, there are significant distinctions between two problems. Since in-bed pose estimation is often based on a long-term monitoring scenario, there will be notable differences in lighting conditions throughout a day (with no light during sleep time), which makes it challenging to keep uniform image quality via classical methods. Moreover, if night vision technology is employed to address this challenge, the color information will be lost. Another difference is on the imaging angle, which for in-bed applications is overview (bird's-eye view) and subject overall orientation will have a different distribution from a common human surveillance viewpoint. For instance, it is possible that human appears upside-down in an overview image, but it is quite rare to see an upside-down human from a side viewpoint. In addition, the similarity between the background (bed sheets) and foreground (human clothing) is magnified in in-bed applications. To the extent of our knowledge, there is no existing work that has addressed these issues. In addition, no specific in-bed human pose dataset has been released to demonstrate and compare the possibilities of employing existing models to serve for in-bed pose estimation. 

In this paper, we address the aforementioned challenges and make the following contributions: (i) Developing an infrared selective (IRS) image acquisition method to provide stable quality images under significant illumination variations between day and night. (ii) Improving the pose estimation performance of a pre-trained CPM model from side viewpoint dataset by adding a 2-end histogram of oriented gradient (HOG) orientation rectification method, which improved performance of the existing model over 26.4\% on our dataset. (iii) Proposing a fine-tuning strategy for intermediate layers of CPM, which has surpassed the classical model accuracy by 16.6\% in detecting in-bed human poses. (iv) Considering practical cases and embedded implementation requirements (e.g. to preserve privacy), an on-demand trigger estimation framework is proposed to reduce computational cost. (v) Building an in-bed human pose dataset with annotation from several realistic life-size mannequins with clothing differing in color and texture in a simulated hospital room via proposed IRS system. The dataset also includes a semi-automated body part annotation tool.

%3. We find bounding box and lying orientation will apparently affect the estimation result of existing CNN model. 

\section{Methods}
Most human pose estimation works exclusively address the pose estimation when a human-contained bounding box is given. Instead, our work presents a system level automatic pipeline, which extracts information directly from raw video sequence inputs, while contains all the related pre-processing parts. An overview of our system is presented in \figref{system}. In \secref{IRS}, we first introduce the IRS acquisition method to address the lighting condition variation issue during day and night. Then in \secref{ortDet}, we suggest the n-end HOG rectification method to handle the unusual pose distribution from overview angle. \secref{onDemTrigger} describes on-demand trigger mechanism, which provides on-demand pose estimation. Finally in \secref{poseEst}, an example of general purpose pose estimation models based on deep neural networks is repurposed for in-bed pose estimation.

In particular, we used convolutional pose machine (CPM) as a pre-trained CNN \cite{wei2016convolutional}. We also employed a high performance pictorial structure oriented method, called flexible mixture of parts (FMP) during experimental analysis for estimation accuracy comparison. The rational behind using CPM and FMP is that these two algorithms represent two typical frameworks for pose estimation, one is based on deep learning, and the other is based on the pictorial structure, one of the  most classical pose estimation models. 
In the case of CPM, to deal with the high-volume data requirement issue, a fine-tuning strategy is also suggested, which is based on training only a few specific layers rather than retraining the whole network. Therefore, we were able to evaluate the pose estimation accuracy of both models using our "shallow" in-bed dataset. We chose the term "shallow" to indicate the differences between our IRS in-bed pose data and publicly available general purpose pose data. These differences include limited size of dataset, lack of color information, and irregular orientation and poses that one may take while being in bed.

\figIRS
\subsection{Infrared Selective (IRS) Image Acquisition System} \label{sec:IRS}
Available datasets for pose estimation are collected under well illuminated environment and the subjects are visible enough to be captured by regular cameras. However, in-bed pose estimation requires to be conducted not only during daytime but also during night time, which means to be functional under a totally dark environment. Night vision cameras are commercially available, however the resultant images are significantly different than images from regular cameras, which raises great challenges to the pose estimation methods. 

\subsubsection{IRS imaging system implementation}
To address this issue, we developed an IRS image acquisition method, which provides stable quality image under huge illumination variation between day and night. The IRS imaging benefits from the difference between human vision and charge coupled device (CCD) cameras, which show different sensitivity to the same spectrum. CCD cameras capture larger range of spectrum beyond human capability, which makes the visualization possible under dark environment to human \cite{CCDSpectrum}. Our system avoids the visible light spectrum, which ranges from 400nm to 700nm, and selects the infrared spectrum ranging from 700nm to 1mm. Different from traditional night vision cameras, which only employ the IR light to enhance the lighting condition during night, we filter out the whole visible light spectrum in order to make the image quality invariant to lighting conditions, thus making robust performance estimation possible. The IRS imaging process and the hardware implementation are shown in \figref{IRSProc} and \figref{IRSdiagram}, respectively.

\figref{IRSsampling} shows the images captured by IRS system and a comparing pair from a normal webcam. It clearly demonstrates that IRS system  provides stable image quality under huge illumination variations. This makes the night monitoring possible without disturbing subjects during sleep. Another advantage of using IRS imaging is it produces high contrast foreground and background, which makes the segmentation easier. In terms of the safety of our IRS imaging system, it is proved that IR light is a non-ionizing radiation, which has insufficient energy to produce any type of damage to human tissue. Most common effect generated by IR is heating \cite{zamanian2005electromagnetic}. In our case, the visualization radiation is far below the dangerous level due to its low power density.

\figIRSsampling

\subsubsection{New challenges from IRS}
IRS provides a way for stable image acquisition for day long monitoring, however the use of the IRS setup results in new challenges. \figref{webLightOn} and \figref{webLightOff} show the images captured from regular cameras with light on and off, respectively. \figref{IRSlightOn} and \figref{IRSlightOff} show the images captured by our IRS system the under same conditions. 
As you can see, the color information is totally lost from this process and the purple color in the image \figref{IRSlightOff} is resulted form filtering process. To employ existing pose estimation models, we assumed this false color as gray intensity information and replicated this to three channels, what is the standard input format for most pose estimation models. It is shown that the color information is not trivial in pose estimation and its effect on pose estimation accuracy is given in \secref{expColor}.

Moreover, in-bed pose distribution under overview angle will be  different from most public dataset collected from regular side viewpoint. Subjects can be commonly upside-down in an overview image because of their in-bed orientation, which is a rare case from a side viewpoint. This difference is also not trivial during estimation process which is shown in both models under our test (\secref{extOrt}). One example is shown in \figref{CPMorient} where we employed a pre-trained CPM model to test the pose estimation accuracy of same image in our dataset but with different orientations \cite{wei2016convolutional}. The result showed notable differences between the image with portrait orientation and the inverse one.

\subsection{In-bed Pose Orientation Detection}\label{sec:ortDet}
Classically, subject orientation problem during pose estimation is handled by data augmentation technique \cite{krizhevsky2012imagenet}, which artificially enlarge the pose dataset using label-preserving transformations \cite{ciregan2012multi}. However, this technique often results in an extensive re-training computational time. Assuming that the chosen model is capable of capturing pose information from side view, to utilize the model trained on a large dataset, we present an orientation rectification method to re-align the image to a similar position to training set. In order to employ the pre-trained pose estimation models directly in our application, here we present an n-end HOG rectification method to minimize the image misalignment. 
 
\figCPMorient

\subsubsection{Bounding box detection} 
We assume under usual home/hospital settings, beds are aligned with one of the walls of the room. In the case of cuboid rooms, this will result in four general categories of in-bed orientations. Suppose the camera is correctly setup to capture images with major axes approximately parallel to the wall orientations. We define these four general in-bed orientations as north, east, south, and west $\{N,E,S,W\}$. 
The first step to find the general in-bed orientation, is 
locating the human-contained bounding box in the image. This could be a computationally intensive process over multi-scale extensive search for a common vision task. However in our case, due to IRS imaging, foreground appears with high contrast from the background, which makes the segmentation a straightforward threshold-based algorithm. We further noticed that under IRS, the foreground shows visible edges, in which the bounding box can also be extracted from a classical edge detection algorithm using the 'Sobel' operator. The results of applying these two methods are shown in \figref{BB}. When there is no disturbance at surroundings, the edge based bounding box extraction will be more accurate to locate the boundaries. However, threshold based method will be more robust to the noise and a multiple scale search can be employed to improve the results. Information associated with a bounding box are $\mathcal{B} = \{\mathcal{B}_{x_c},\mathcal{B}_{y_c}, \mathcal{B}_w, \mathcal{B}_h\}$, where $\mathcal{B}_{x_c},\mathcal{B}_{y_c}$ represent the coordinate of the up-left corner, and $\mathcal{B}_w,\mathcal{B}_h$ represent the width and height of the bounding box, respectively. 
From the bounding box width and height ratio, in-bed orientations are first categorized into horizontal and vertical ones. To further rectify the orientation, we apply an n-end HOG rectification method as descrined below. 

\figBB

\subsubsection{N-end HOG pose rectification method} \label{sec:NendHog}

HOG features were first employed for pedestrian detection \cite{dalal2005histograms}, which captured the localized portion features by estimating the local gradient orientation statistics. 
These features show the benefit of invariant to the geometric and photometric transformations. Since all the horizontally-orientated images can be detected based on the $\mathcal{B}_w/\mathcal{B}_h$ and rotated back into vertical ones, here the classification is between upside-down images vs. portrait ones, all in vertical cases. As upper and lower body parts show clear differences in their overall geometry, we captured information from large scale patches instead of small grids. Therefore, unlike extracting HOG features on dense grid, we only extracted HOG features on sparse locations. To form HOG features in this way, two information is needed. One is HOG descriptor parameters and the other is \emph{interest points'} locations, where HOG operator to be applied at.

For HOG parameters, we employed a $2 \times 2$ cell structure for each block to capture overall information. The block size is determined by the size of the estimated bounding box as follows:
\begin{equation}
    l_{block} = \text{min} (\bar{\mathcal{B}_w}, \bar{\mathcal{B}_h})
\end{equation}
where $\bar{\mathcal{B}_w}$ and $\bar{\mathcal{B}_h}$ represent the average of width and height of the bounding boxes in images from our IRS dataset, respectively. In practice, the average bounding box information can be achieved by a short period of initial monitoring. Consequently, the cell size is $l_{cell}=l_{block}/2$. For long-term in-bed monitoring applications, once set up, the scale information would stay the same during the monitoring time.

For interesting points' locations, we assumed $\mathcal{B}_w < \mathcal{B}_h$ and the coordination of the first and last interesting points are given as: 
\begin{equation}
\begin{aligned}
  & C_{hog}(1) = ( \mathcal{B}_{x_c}+l_{block}/2, & \mathcal{B}_{y_c}+ l_{block}/2) \\
   & C_{hog}(n) = (\mathcal{B}_{x_c}+l_{block},& \mathcal{B}_{y_c}+\mathcal{B}_h-l_{block}/2)
\end{aligned}    
\end{equation}
where $n$ is an integer stands for the total number of interesting points, and $C_{hog}(n)$ is the center of the n-th HOG descriptors. Once the  two end interesting point coordination are achieved, other interesting point can be extracted from linear interpolation from them. In our case, we chose $n=2$ and 2-end HOG features are generated as shown in \figref{HogNend}. Extracted HOG features from the interest points are cascaded in top to bottom order into HOG feature vector, $f_{2e}$. A support vector machine (SVM) model is then employed as binary classifier on extracted HOG features to give prediction result from orientation categories of $\{N,\neq N\}$. We assign this result to an indicator $bit_N$. Another indicator $bit_H$ comes from the bounding box to show if the subject is horizontal or vertical. The final orientation is decoded from the encoding table shown in \tblref{oriTable}. This process automatically forms a two-layer decision tree as shown in \algref{alg1}. 

\figHogNend

\begin{algorithm}
 \KwIn{Image $I$}
 \KwResult{General in-bed orientation from $\{N,E,S,W\}$, reclined portrait image}
 Initialization;
 
 Edge detection on $I$, output $I_{bw}$;

 Bounding box extraction from $I_{bw}$;
 
 Calculate $\mathcal{B}_w/\mathcal{B}_h$;
 
 \eIf{
 $\mathcal{B}_w/\mathcal{B}_h>1$ 
 }{
Subject has a horizontal in-bed orientation, $bit_H=1$;
Rotate original image to vertical in-bed orientation $I$ = Rotate($I$, -90$^o$);
 }{
Subject has a vertical in-bed orientation, $bit_H=0$;
 }
 
 2-end HOG extraction to form vector $f_{2e}$;

 Get orientation from the SVM classification;

\eIf{$N$}{
$bit_N$ = 1;
}{
$bit_N$ = 0;
}
Predict in-bed orientation from encoding table. 

Rectify image $I$ to $N$ category.
\\~         % man made empty line 
\\
 \caption{2-end HOG rectification method.}
 \label{alg:alg1}
\end{algorithm}

\tabOriTable

\subsection{On-Demand Trigger for Pose Estimation} \label{sec:onDemTrigger}
Typical applications of in-bed pose estimation are overnight sleep monitoring and long-term monitoring of bed-bound patients. In these cases, human on the bed is often less physically active or even totally immobile. Therefore, we can reasonably hold the following hypothesis:
"when the scene is stable, the human pose stays the same." This means that we only need to estimate the pose after each variation in the scene rather than continuously process the video, frame by frame. In this scenario, we propose an on-demand estimation trigger scheme to reduce the computational and power cost of our pose estimation algorithm. This power efficiency is crucial for patient's privacy reasons, since it enables us to build an \emph{in-situ} embedded pose processing system rather than sending all raw videos of the patient during his/her sleep to a base-station for further processing. 

Since this process is conducted in an indoor environment, a threshold-based method is used to detect foreground variations. The pose estimation process then is triggered when the scene recover from the variation. Suppose the current state is $\mathcal{S}_{cur} \in \{0,1\} $ and previous state is $\mathcal{S}_{pre}$, where 1 stands for a dynamic scene and 0 stands for a static one. To get the state value, we make a difference operation by adjacent video frames. If this difference is greater than a threshold, it is assumed to be a dynamic frame, otherwise a static one. When in-bed pose changes, it could be caused by the subject herself or the caregiver. Based on the speed of repositioning, the process possibly contains piece-wise static periods. To suppress this false static state, we employed a backward window $\mathcal{W}_{bf}$ of size $N_{bf}$ to filter the raw state result. The filtered state $\hat{\mathcal{S}}_{cur}$ is 0 only when all states in the backward window show static states, otherwise it is 1. This operation as shown in \algref{onDemand} is designed to favor dynamic states and guarantees a gap between static states if short disturbance occurs between them. 
\begin{algorithm}[h] 
 \KwIn{Video stream $I$}
 \KwResult{Trigger pose estimation process}
 Initialization;
 
 \While{new frame}
 {  
    Get difference of adjacent frames\\    
    Update $\mathcal{W}_{bf}$     
    
    \eIf{\text{max} $(\mathcal{W}_{bf})==1$}
    {
        $\hat{\mathcal{S}}_{cur} =1$
    }{
        $\hat{\mathcal{S}}_{cur} = 0$
    }
    \If{$\hat{\mathcal{S}}_{cur}$- $\hat{\mathcal{S}}_{pre}<0$ }
    {        
        Trigger pose estimation process
    }
    $\hat{\mathcal{S}}_{pre} =\hat{\mathcal{S}}_{cur}$
}
 \caption{On-demand pose estimation trigger.}
 \label{alg:onDemand}
\end{algorithm}

\figTuneMdl
\subsection{Fine-Tuning CPM for In-bed Pose Estimation Purpose} \label{sec:poseEst}
Even with larger orientation possibilities and full loss of color information, in-bed human poses still share great similarities with ones taken from side views. We believe a well-trained general purpose pose estimation model is still able to capture body parts' features and kinematic constraints between them. In this work, a recent CNN-based pose estimation approach, called convolutional pose machine (CPM) is employed as a pre-trained pose estimation model \cite{wei2016convolutional}. CPM employs multi-stage structure to estimate human pose, in which each stage is a multi-layer CNN. Each stage takes in not only the image features, but also previous stage's belief map results as input. The final stage outputs the final estimation results, which are the 14 key joints' coordinates in image domain that include left and right (L/R) ankles, L/R knees, L/R hips, L/R wrists, L/R elbows, L/R shoulders, top of the head and neck. In the original work that introduced CPM \cite{wei2016convolutional}, CPM with 6 stages has shown promising estimation results on large scale dataset such as MPII \cite{andriluka20142d}, LSP \cite{johnson2010clustered} and FLIC \cite{sapp2013modec}. However, for a new query image, manual intervention was still required to indicate the exact bounding box of the human in the scene.

Due to the IRS imaging system and 2-end HOG method, our proposed method is able to  accurately locate the human-contained bounding box and efficiently rectify the image orientation, which drastically save the cost of extensive search across multi-scale. These properties provide a more efficient way to directly apply pre-trained CNN model on an in-bed pose dataset.  
Furthermore, in order to adapt to the input layer dimension of the pre-trained model, each input image is amplified into three channels, which share the same intensity value. 

When available dataset is limited in size, such as our IRS in-bed pose data, it is a golden rule to fine-tune the deep neural network model with only fully connected layers or the last layer \cite{yosinski2014transferable}. However, based on the multi-stage configuration of the CPM, other fine-tuning approaches can also be applied. In this work, three fine-tuning strategies are proposed, which are illustrated in \figref{tuneMdl}.
First strategy, called MANNE-S6 takes the convention to train the very last layer before output or fully connected layer \cite{yosinski2014transferable}. Due to the CPM's special configuration with multiple stages, in second configuration, we train the last layer of each stage, which is called MANNE-AS. We also notice that there is a shared layer in CPM structure, which is the 3rd convolutional layer located in stage 2. Therefore, in third strategy, we further put this layer under training. This strategy is called MANNE-AS-S2C3-200, when it is trained with 200 iterations, and is called MANNE-AS-S2C3-2000, when it is trained with 2000 iterations. More iterations will enhance the probability of capturing more training samples' patterns and tuning the model weights to more representative values. Without any fine-tuning, the pre-trained CPM model using MPII and LSP dataset is called MPII-LSP. 

To compare the effectiveness of the deep learning against other non-deep models when our IRS in-bed pose dataset is used, we employed a recently proposed pictorial structure oriented model with flexible mixtures of parts (FMP) \cite{yang2013articulated}, which has also shown great general purpose pose estimation performance on small scale human pose datasets such as PARSE \cite{Antol2014} and BUFFY \cite{Ferrari08}. 

\section{Experimental Setup and Analysis}
\figIRSex
\subsection{Building an In-Bed Pose Dataset}
Although there are several public human pose datasets available such as MPII \cite{sapp2013modec}, LSP \cite{andriluka20142d}, FLIC \cite{johnson2010clustered}, Buffy \cite{ramanan2006learning}, they are all mainly from scenes such as sports, TV shows, and other daily activities. None of them provides any specific in-bed poses. To fill this gap, we crafted an image acquisition system based on an IRS configuration (\figref{IRSdevice}) and collected IRS data from one male and one female realistic life-size mannequins in a simulated hospital room (\figref{exSetup}). 

\figSampImg

\figSampImgColor
Using mannequins gave us the option to collect images from different in-bed postures (supine, left side lying, and right side lying) by changing their poses with high granularity. Limited by the number of available mannequins, we collected data from the mannequins with different clothes, mainly different color/texture hospital gowns. We totally collected 419 poses, some of which are shown in \figref{smpImg}. For comparison purpose, a color edition in-bed pose dataset is also established under the same setting but with a overview normal webcam. Some samples of the colored in-bed pose dataset is shown in \figref{sampleImgColor}. A semi-automated tool for human pose annotation is designed in MATLAB, in which the joint indices follow the LSP convention \cite{andriluka20142d}, as shown in \figref{smpImg}(g) to (l). The GUI of this tool (shown in \figref{annoTool}),  provides the convenience to label join locations and visibility in a semi-automated way. 

\figAnnoTool

\figTestBC

\subsection{Pose Estimation Performance Measure}
Throughout the result section, probability of correct keypoint (PCK) criteria is employed for pose estimation performance evaluation, which is the measure of joint localization accuracy \cite{yang2013articulated}. 
The distance between the estimated joint position and the ground-truth position is compared against a threshed defined as fraction of the person's torso length, where torso length is defined as the distance between person's left shoulder and right hip \cite{andriluka14cvpr}. For instance, PCK0.1 metric means the estimation is correct when the distance between the estimated joint position and the ground-truth position is less than 10\% of the person's torso length.
This is usually considered a high precision regime. For the experimental analysis, we illustrate the pose estimation results of different models for the body part categories of total (all body parts), hip, knee, ankle, head, shoulder, elbow, and wrist by combining the estimation results of left and right corresponding limbs.

\figMultOrt

\subsection{Does Color Information Matter?} \label{sec:expColor}
One obvious difference between the IRS in-bed dataset and the publicly available general purpose datasets is the loss of color information. To investigate the influence of color loss on pre-trained models, we employed a pre-trained CPM model (trained on MPII and LSP dataset) to estimate poses of our mannequin dataset collected using IRS imaging and a normal webcam, respectively. To exclude the influence of unusual orientation, we only compare these two datasets from portrait image angle. To show the general effect of color loss on other pre-trained models, an FMP model is also evaluated under the same setting.

% The correct detection is based on how far the predicted joint position bias from the its true position in percentage of body length. We deem 10\% as a high regime and 20\% as a low regime. The percentage of the correct detection out of the all predicted joints is deemed as detection ratio. 

As shown in \figref{testBC}, both pre-trained models show better result on colored dataset than its black and white (BW) counterpart. Improvements from color information bring much more improvement in CPM than FMP. It shows color information is important in both models and is more helpful in CNN framework. In overall performance, the CPM gives better result. Even its performance on BW edition surpasses the FMP color edition. These results once more clarify our rationale for choosing a CNN based framework as the main pre-trained model for in-bed pose estimation.

\figMultMdl
\figBlfMap
\subsection{Unusual Orientation Handling} \label{sec:extOrt}
To handle the unusual orientation resulted from overview camera angle, a 2-end HOG rectification method was employed. We evaluated the effectiveness of the process in two phases. In the first phase, we tested the accuracy of 2-end HOG orientation detection and rectification method. We augmented our IRS in-bed pose dataset by synthesizing and adding several in-bed orientations in $\{N,E,S,W\}$ general categories for each image. $bit_H$ and $bit_N$ were obtained for a given image in dataset based on  $\mathcal{B}_w/\mathcal{B}_h$ and the results of the SVM classifier as explained in \algref{alg1}. Using a 10-fold cross validation scheme, 99\% accuracy in the general orientation detection was achieved.

In the second phase, to further evaluate the pose estimation performance on unusually oriented images (belonging to $\{E,S,W\}$ categories)  vs. rectified images (all re-aligned to $\{N\}$ position), we employed a pre-trained CPM model from MPII \cite{andriluka14cvpr} and LSP \cite{johnson2010clustered} dataset and also the flexible of parts (FMP) model \cite{yang2013articulated} as our pose estimation models. We then divided our IRS in-bed pose dataset into two subsets: 370 images for training and 49 for test and used PCK metric for performance evaluation, as suggested in \cite{johnson2010clustered}.
The estimation performance on images belonging to  $\{E,S,W\}$ in-bed orientations categories is compared to the portrait images after 2-end HOG rectification and the results are shown in \figref{multOrt}. These results demonstrate that in-bed orientation significantly affects the pose estimation accuracy and our proposed 2-end HOG rectification method boosts the estimation performance by a large margin for both CNN based and pictorial structure based models. Our method shows promising to act as a generic purpose tool to enhance the performance of pre-trained models for in-bed case.

\subsection{Fine-Tuning of a Deep Model}
To further improve the performance of our chosen neural network model, the MPII-LSP pre-trained CPM, we performed fine-tuning with different configurations as shown in \figref{tuneMdl}. We trained all three proposed configurations with small iteration (=200) with batch size of 16.
The performance of CPM model after different fine-tuning strategies are shown in \figref{multMdl}, compared to the original MPII-LSP pre-trained CPM \cite{wei2016convolutional}. Interestingly, our third fine-tuning configuration, MANNE-AS-S2C3-200, showed the highest estimation performance when compared to the traditional fine-tuning approach. Then we increased the iteration number to 2000 for the third configuration, which further improved the estimation results.

In 200 iteration training test, MANNE-S6 does not show improvement over original model, however our proposed strategy, MANNE-AS-S2C3-200 shows clear improvement after 200 iteration in all body parts except the head part. MANNE-AS-S2C3-200 model shows improvement at PCK0.1, however falls behind at PCK0.2. It means the model either gives accurate answer for the head location or drifts far away from the correct location. This may come from the fact that the head part depends more on local image features. This drawback however is resolved after more iterations. Our final fine-tuned model MANNE-AS-S2C3-2000 surpassed the original pre-trained CPM MPII-LSP and also the traditional fine-tuned model MANNE-S6 by nearly 20\% at PCK0.2 criterion. One sample of an estimation belief map is shown in \figref{blfMap}. We hypothesize that the success of MANNE-AS-S2C3-2000 is due to the fact that its first 3 layers are reused in all the following stages, which means it has larger influence on the final output, and the outcome performances validate this hypothesis.

\figMultPre

\tabMdlPCK  % different place 

Here, we also present and compare the results of pose estimation using a classical framework against the deep neural network model. We employed a recent augmented pictorial structure based method with flexible mixture of parts (FMP), which showed best pose estimation performance on PARSE dataset \cite{Antol2014} at that time and comparable performance to the state-of-the-art non-deep leaning methods \cite{johnson2011learning,yang2013articulated}. Test result on our IRS dataset with our fine-tuned CPM model, pre-trained CPM with MPII-LSP dataset and FMP model are shown in \figref{multPre}. Our fine-tuned model shows advantages in total accuracy across all PCK standards. However surprisingly, trained only on a small dataset, FMP surpasses the CPM performance in all upper body parts' estimation in a high precision regime (PCK0.1) and slightly inferior in the low precision regime (PCK0.2) for head detection. This result once more emphasizes the importance of color information in the CPM model. Instead, FMP essentially employs the HOG features, which highly depend on image gradients. This is the reason that FMP surpasses the pre-trained CPM in several body parts' estimation. For example, the head, shoulder, and elbow show obvious shape features compared to other body parts, which is more easily captured by the HOG descriptors than the color information. The quantitative result of PCK0.2 is shown in \tblref{mdlPCK} and our fine-tuned model surpasses the second best model by 19.6\%.

\subsection{On-Demand Estimation Trigger}
To validate the effectiveness of our on-demand trigger pipeline, we video monitored a mannequin on bed via using IRS system. In this video, we mimicked the practical scenario, where hospital bed is moved around and kept stable for a while after each relocation. In this process, mannequin was located in different in-bed orientations as part of the $\{N,E,S,W\}$ general categories, defined in \secref{ortDet}. We simulated 4 times relocation in the video and each stable period in between lasted approximately 6 to 8 seconds, which is enough for our algorithm to distinguish the static states ($\mathcal{S} = 0$) from the dynamic states ($\mathcal{S} = 1$). The pose estimation algorithm is triggered only at each falling edge, when $\mathcal{S}$ transits from 1 to 0, and not frame by frame. 

To generate the the ground-truth label for the video, we replayed the video and annotated the start and end points of the dynamic states manually by recording their frame index. Our on-demand trigger method is also applied on this video with backward window $\mathcal{W}_{bf}$ of size $N_{bf} = 30$. As the test video has a frame rate of 11.28 frame/s, this window is approximate 2.66s. \figref{onDemTrigger} shows our state estimation results against the ground-truth and the trigger signal to initiate the estimation pipeline. It shows that our algorithm is successful in triggering the estimation after each dynamic to static state transition. There is a slight lag between our trigger and ground-truth label, which is due to the use of backward window of size 2.66s. In practice, caregivers in nursing homes and hospitals usually perform posture repositioning for pressure re-distribution on a regular basis to prevent bed born complications such as pressure ulcers \cite{ostadabbas2011posture}. Considering a recommended 2-hour interval between repositioning, even 10 seconds lag can only result in 0.12\% information loss and the loss caused by our lag is much smaller. 

\figOnDemTrigger

\section{Discussion and Future Work} 

In this work, we have presented a comprehensive system to estimate in-bed human poses and address the challenges associated with this specific pose estimation problem. The issue of huge lighting variations for this application is addressed by our proposed IRS imaging system. The image differences between the overview angle used for human in-bed monitoring and the side angle often used in available human pose datasets is handled by our proposed 2-end HOG rectification method, which effectively improve the performance of existing pose estimation models for irregular poses. In CV applications, this issue is usually handled by extensively augmenting the dataset to cover all possible orientations. However, our rectification method avoids the time/memory expense of retraining the whole network by the proposed preprocessing steps.

Without a large dataset, retraining a deep neural network from scratch is not feasible. In this paper, we explored the idea of using a shallow (limited in size and different in perspective and color) dataset collected from in-bed poses to retrained a CNN, which was pre-trained on general human poses. We showed that classical fine-tuning principle is not always effective and the network architecture matters. For the specific CNN, the CPM model, our proposed fine-tuning model demonstrated clear improvement over the classical one. 

The problem of in-bed pose estimation still has other challenges that remain. The main one is the high probability of being covered by a sheet or blanket while on bed. In fact, vision-based methods would no longer be functional in this case. Other sensing modalities may provide other forms of indication for pose inference, however it is less likely to be able to retrieve color information from those modalities. In this respect, this work is also a pilot study for pose estimation under information loss. In future work, we plan to address this issue by employing other sensing modalities to complement vision information. Test on real human data is also anticipated in our next step. 

\balance
\bibliographystyle{IEEEtran}
\small
\bibliography{paper}

\begin{thebibliography}{10}
\providecommand{\url}[1]{#1}
\csname url@samestyle\endcsname
\providecommand{\newblock}{\relax}
\providecommand{\bibinfo}[2]{#2}
\providecommand{\BIBentrySTDinterwordspacing}{\spaceskip=0pt\relax}
\providecommand{\BIBentryALTinterwordstretchfactor}{4}
\providecommand{\BIBentryALTinterwordspacing}{\spaceskip=\fontdimen2\font plus
\BIBentryALTinterwordstretchfactor\fontdimen3\font minus
  \fontdimen4\font\relax}
\providecommand{\BIBforeignlanguage}[2]{{%
\expandafter\ifx\csname l@#1\endcsname\relax
\typeout{** WARNING: IEEEtran.bst: No hyphenation pattern has been}%
\typeout{** loaded for the language `#1'. Using the pattern for}%
\typeout{** the default language instead.}%
\else
\language=\csname l@#1\endcsname
\fi
#2}}
\providecommand{\BIBdecl}{\relax}
\BIBdecl

\bibitem{lee2015changes}
C.~H. Lee, D.~K. Kim, S.~Y. Kim, C.-S. Rhee, and T.-B. Won, ``Changes in site
  of obstruction in obstructive sleep apnea patients according to sleep
  position: a dise study,'' \emph{The Laryngoscope}, vol. 125, no.~1, pp.
  248--254, 2015.

\bibitem{black2007national}
J.~Black, M.~M. Baharestani, J.~Cuddigan, B.~Dorner, L.~Edsberg, D.~Langemo,
  M.~E. Posthauer, C.~Ratliff, G.~Taler \emph{et~al.}, ``National pressure
  ulcer advisory panel's updated pressure ulcer staging system,''
  \emph{Advances in skin \& wound care}, vol.~20, no.~5, pp. 269--274, 2007.

\bibitem{mccabe2011preferred}
S.~J. McCabe, A.~Gupta, D.~E. Tate, and J.~Myers, ``Preferred sleep position on
  the side is associated with carpal tunnel syndrome,'' \emph{Hand}, vol.~6,
  no.~2, pp. 132--137, 2011.

\bibitem{mccabe2010evaluation}
S.~J. McCabe and Y.~Xue, ``Evaluation of sleep position as a potential cause of
  carpal tunnel syndrome: preferred sleep position on the side is associated
  with age and gender,'' \emph{Hand}, vol.~5, no.~4, pp. 361--363, 2010.

\bibitem{abouzari2007role}
M.~Abouzari, A.~Rashidi, J.~Rezaii, K.~Esfandiari, M.~Asadollahi, H.~Aleali,
  and M.~Abdollahzadeh, ``The role of postoperative patient posture in the
  recurrence of traumatic chronic subdural hematoma after burr-hole surgery,''
  \emph{Neurosurgery}, vol.~61, no.~4, pp. 794--797, 2007.

\bibitem{pouyan2013continuous}
M.~B. Pouyan, S.~Ostadabbas, M.~Farshbaf, R.~Yousefi, M.~Nourani, and
  M.~Pompeo, ``Continuous eight-posture classification for bed-bound
  patients,'' \emph{2013 6th International Conference on Biomedical Engineering
  and Informatics}, pp. 121--126, 2013.

\bibitem{Ostadabbas2014}
S.~Ostadabbas, M.~Pouyan, M.~Nourani, and N.~Kehtarnavaz, ``{In-bed posture
  classification and limb identification},'' \emph{2014 IEEE Biomedical
  Circuits and Systems Conference (BioCAS) Proceedings}, pp. 133--136, 2014.

\bibitem{liu2014bodypart}
J.~J. Liu, M.-C. Huang, W.~Xu, and M.~Sarrafzadeh, ``Bodypart localization for
  pressure ulcer prevention,'' \emph{2014 36th Annual International Conference
  of the IEEE Engineering in Medicine and Biology Society}, pp. 766--769, 2014.

\bibitem{andriluka2009pictorial}
M.~Andriluka, S.~Roth, and B.~Schiele, ``Pictorial structures revisited: People
  detection and articulated pose estimation,'' \emph{Computer Vision and
  Pattern Recognition, 2009. CVPR 2009. IEEE Conference on}, pp. 1014--1021,
  2009.

\bibitem{andriluka2010monocular}
M.~Andriluka and S.~Roth, ``Monocular 3d pose estimation and tracking by
  detection,'' \emph{Computer Vision and Pattern Recognition (CVPR), 2010 IEEE
  Conference on}, pp. 623--630, 2010.

\bibitem{felzenszwalb2005pictorial}
P.~F. Felzenszwalb and D.~P. Huttenlocher, ``Pictorial structures for object
  recognition,'' \emph{International journal of computer vision}, vol.~61,
  no.~1, pp. 55--79, 2005.

\bibitem{johnson2010clustered}
S.~Johnson and M.~Everingham, ``Clustered pose and nonlinear appearance models
  for human pose estimation.'' \emph{BMVC}, vol.~2, p.~5, 2010.

\bibitem{pishchulin2013poselet}
L.~Pishchulin, M.~Andriluka, P.~Gehler, and B.~Schiele, ``Poselet conditioned
  pictorial structures,'' \emph{Proceedings of the IEEE Conference on Computer
  Vision and Pattern Recognition}, pp. 588--595, 2013.

\bibitem{pishchuli2013strong}
L.~Pishchulin, M.~Andriluka, and P.~Gehler, ``Strong appearance and expressive
  spatial models for human pose estimation,'' \emph{Proceedings of the IEEE
  International Conference on Computer Vision}, pp. 3487--3494, 2013.

\bibitem{Antol2014}
S.~Antol, C.~L. Zitnick, and D.~Parikh, ``{Zero-Shot Learning via Visual
  Abstraction},'' \emph{ECCV}, 2014.

\bibitem{Ferrari08}
V.~Ferrari, M.~Marin-Jimenez, and A.~Zisserman, ``Progressive search space
  reduction for human pose estimation,'' \emph{Proceedings of the IEEE
  Conference on Computer Vision and Pattern Recognition}, Jun. 2008.

\bibitem{sun2011articulated}
M.~Sun and S.~Savarese, ``Articulated part-based model for joint object
  detection and pose estimation,'' \emph{Computer Vision (ICCV), 2011 IEEE
  International Conference on}, pp. 723--730, 2011.

\bibitem{tian2012exploring}
Y.~Tian, C.~L. Zitnick, and S.~G. Narasimhan, ``Exploring the spatial hierarchy
  of mixture models for human pose estimation,'' \emph{European Conference on
  Computer Vision}, pp. 256--269, 2012.

\bibitem{dantone2013human}
M.~Dantone, J.~Gall, C.~Leistner, and L.~Van~Gool, ``Human pose estimation
  using body parts dependent joint regressors,'' \emph{Proceedings of the IEEE
  Conference on Computer Vision and Pattern Recognition}, pp. 3041--3048, 2013.

\bibitem{karlinsky2012using}
L.~Karlinsky and S.~Ullman, ``Using linking features in learning non-parametric
  part models,'' \emph{European Conference on Computer Vision}, pp. 326--339,
  2012.

\bibitem{wang2008multiple}
Y.~Wang and G.~Mori, ``Multiple tree models for occlusion and spatial
  constraints in human pose estimation,'' \emph{European Conference on Computer
  Vision}, pp. 710--724, 2008.

\bibitem{munoz2010stacked}
D.~Munoz, J.~Bagnell, and M.~Hebert, ``Stacked hierarchical labeling,''
  \emph{Computer Vision--ECCV 2010}, pp. 57--70, 2010.

\bibitem{pinheiro2014recurrent}
P.~Pinheiro and R.~Collobert, ``Recurrent convolutional neural networks for
  scene labeling,'' \emph{International Conference on Machine Learning}, pp.
  82--90, 2014.

\bibitem{pishchul2016deepcut}
L.~Pishchulin, E.~Insafutdinov, S.~Tang, B.~Andres, M.~Andriluka, P.~V. Gehler,
  and B.~Schiele, ``Deepcut: Joint subset partition and labeling for multi
  person pose estimation,'' \emph{Proceedings of the IEEE Conference on
  Computer Vision and Pattern Recognition}, pp. 4929--4937, 2016.

\bibitem{tompson2014joint}
J.~J. Tompson, A.~Jain, Y.~LeCun, and C.~Bregler, ``Joint training of a
  convolutional network and a graphical model for human pose estimation,''
  \emph{Advances in neural information processing systems}, pp. 1799--1807,
  2014.

\bibitem{tompson2015efficient}
J.~Tompson, R.~Goroshin, A.~Jain, Y.~LeCun, and C.~Bregler, ``Efficient object
  localization using convolutional networks,'' \emph{Proceedings of the IEEE
  Conference on Computer Vision and Pattern Recognition}, pp. 648--656, 2015.

\bibitem{wei2016convolutional}
S.-E. Wei, V.~Ramakrishna, T.~Kanade, and Y.~Sheikh, ``Convolutional pose
  machines,'' \emph{Proceedings of the IEEE Conference on Computer Vision and
  Pattern Recognition}, pp. 4724--4732, 2016.

\bibitem{CCDSpectrum}
A.~Treiman, ``{Life at the Limits: Earth, Mars, and Beyond},''
  \url{http://www.lpi.usra.edu/education/fieldtrips/2005/activities/ir_spectrum/},
  [Online; accessed 19-July-2008].

\bibitem{zamanian2005electromagnetic}
A.~Zamanian and C.~Hardiman, ``Electromagnetic radiation and human health: A
  review of sources and effects,'' \emph{High Frequency Electronics}, vol.~4,
  no.~3, pp. 16--26, 2005.

\bibitem{krizhevsky2012imagenet}
A.~Krizhevsky, I.~Sutskever, and G.~E. Hinton, ``Imagenet classification with
  deep convolutional neural networks,'' \emph{Advances in neural information
  processing systems}, pp. 1097--1105, 2012.

\bibitem{ciregan2012multi}
D.~Ciregan, U.~Meier, and J.~Schmidhuber, ``Multi-column deep neural networks
  for image classification,'' \emph{Computer Vision and Pattern Recognition
  (CVPR), 2012 IEEE Conference on}, pp. 3642--3649, 2012.

\bibitem{dalal2005histograms}
N.~Dalal and B.~Triggs, ``Histograms of oriented gradients for human
  detection,'' \emph{2005 IEEE Computer Society Conference on Computer Vision
  and Pattern Recognition (CVPR'05)}, vol.~1, pp. 886--893, 2005.

\bibitem{andriluka20142d}
M.~Andriluka, L.~Pishchulin, P.~Gehler, and B.~Schiele, ``2d human pose
  estimation: New benchmark and state of the art analysis,'' \emph{Proceedings
  of the IEEE Conference on Computer Vision and Pattern Recognition}, pp.
  3686--3693, 2014.

\bibitem{sapp2013modec}
B.~Sapp and B.~Taskar, ``Modec: Multimodal decomposable models for human pose
  estimation,'' \emph{Proceedings of the IEEE Conference on Computer Vision and
  Pattern Recognition}, pp. 3674--3681, 2013.

\bibitem{yosinski2014transferable}
J.~Yosinski, J.~Clune, Y.~Bengio, and H.~Lipson, ``How transferable are
  features in deep neural networks?'' \emph{Advances in neural information
  processing systems}, pp. 3320--3328, 2014.

\bibitem{yang2013articulated}
Y.~Yang and D.~Ramanan, ``Articulated human detection with flexible mixtures of
  parts,'' \emph{IEEE Transactions on Pattern Analysis and Machine
  Intelligence}, vol.~35, no.~12, pp. 2878--2890, 2013.

\bibitem{ramanan2006learning}
D.~Ramanan, ``Learning to parse images of articulated bodies,'' \emph{NIPS},
  vol.~1, p.~7, 2006.

\bibitem{andriluka14cvpr}
M.~Andriluka, L.~Pishchulin, P.~Gehler, and B.~Schiele, ``2d human pose
  estimation: New benchmark and state of the art analysis,'' \emph{IEEE
  Conference on Computer Vision and Pattern Recognition (CVPR)}, June 2014.

\bibitem{johnson2011learning}
S.~Johnson and M.~Everingham, ``Learning effective human pose estimation from
  inaccurate annotation,'' \emph{Computer Vision and Pattern Recognition
  (CVPR), 2011 IEEE Conference on}, pp. 1465--1472, 2011.

\bibitem{ostadabbas2011posture}
S.~Ostadabbas, R.~Yousefi, M.~Nourani, M.~Faezipour, L.~Tamil, and M.~Pompeo,
  ``A posture scheduling algorithm using constrained shortest path to prevent
  pressure ulcers,'' \emph{Bioinformatics and Biomedicine (BIBM), 2011 IEEE
  International Conference on}, pp. 327--332, 2011.

\end{thebibliography}

\end{document}